\documentclass[11pt, a4paper]{article}

\usepackage[T1]{fontenc}
\usepackage{microtype}
\usepackage[utf8]{inputenc}

\usepackage[a4paper, margin=1in]{geometry}
\usepackage{setspace}
\usepackage{titlesec}
\titleformat{\section}{\Large\bfseries\sffamily}{\thesection}{1em}{}
\titleformat{\subsection}{\large\bfseries\sffamily}{\thesubsection}{1em}{}
\titleformat{\subsubsection}{\normalsize\bfseries\itshape}{\thesubsubsection}{1em}{}
\titlespacing*{\section}{0pt}{1.5em}{0.75em}
\titlespacing*{\subsection}{0pt}{1.2em}{0.5em}

\usepackage{graphicx}
\graphicspath{{figures/}{supp_figures/}}
\usepackage{float}
\usepackage[section]{placeins}
\usepackage{booktabs}

\usepackage[font=small, labelfont=bf, labelsep=period]{caption}
\usepackage{subcaption}

\usepackage{amsmath, amssymb}

\usepackage[numbers, sort&compress]{natbib}
\usepackage[dvipsnames]{xcolor}
\definecolor{linkblue}{HTML}{2166ac}
\usepackage[
    colorlinks=true,
    linkcolor=linkblue,
    citecolor=linkblue,
    urlcolor=linkblue
]{hyperref}

\usepackage{enumitem}
\setlist{nosep, leftmargin=*}
\usepackage{siunitx}
\usepackage[ruled, lined, linesnumbered]{algorithm2e}

\newcommand{\sgc}{SGC}
\newcommand{\af}{AlphaFold2}
\newcommand{\plddt}{pLDDT}
\newcommand{\ptm}{pTM}
\newcommand{\sig}{\sigma}
\newcommand{\lam}{\lambda}

\usepackage{fancyhdr}
\fancypagestyle{plain}{
    \fancyhf{}
    \fancyfoot[C]{\thepage}
    
}

\usepackage{multirow}
\usepackage{tikz}
\usetikzlibrary{positioning,arrows.meta,calc}

\begin{document}
\thispagestyle{plain}

\begin{center}
    {\fontsize{18}{22}\selectfont\bfseries
    Neural spectroscopy of AlphaFold2 reveals\\[4pt]
    encoded protein conformational landscapes}

    \vspace{1.0em}

    {\normalsize Kaustav Mehta}\\[0.15em]
    {\small Independent researcher, India}\\[0.15em]
    {\small Correspondence: \href{mailto:kaustav.mehta@protonmail.com}{\textcolor{black}{\ttfamily kaustav.mehta@protonmail.com}}}\\[0.5em]
    {\small July 17, 2026}
\end{center}

\vspace{1.0em}

\begin{abstract}
\noindent
AlphaFold2's 93 million parameters, shaped by the evolutionary record of protein structure encoded in the Protein Data Bank and in sequence alignments, are conventionally treated only as machinery for converting sequence to structure. We propose they are also a scientific object that can be analyzed directly: a learned encoding of protein conformational organization that can be probed and characterized. By smoothing the Evoformer's weight tensors with a Gaussian convolution and scaling the result, we show that the trained model produces physically structured conformational landscapes. Under perturbation, ubiquitin's native contacts break in the order established by decades of folding experiments. For KaiB, five independently trained models agree that the alternative fold is not recovered under perturbation. For $\alpha$-synuclein, five models produce five different but coherent landscapes, mapping where the training signal has determined the representation and where it has not. Matched-power noise controls confirm that random corruption of equal magnitude produces debris, not conformations. The model learned to predict static structures; the conformational organization visible under perturbation was not an explicit training target, suggesting it emerged as a byproduct of that objective. AlphaFold2's weights appear to encode structural constraints, shaped by evolutionary and structural training data, that extend beyond what unperturbed inference reveals. We call the approach of reading them neural spectroscopy, and Scaled Gaussian Convolution one such protocol.
\end{abstract}

\vspace{0.5em}

\newpage

\section{Introduction}
\label{sec:intro}

\subsection{What has AlphaFold2 learned?}
\label{sec:intro-learned}

AlphaFold2 predicts protein structures with remarkable accuracy, but the source of that accuracy remains largely opaque. The model achieves near-experimental agreement across a broad range of protein families, including proteins with no close structural homologues in the Protein Data Bank. That accuracy has held up across years of community evaluation on held-out targets, novel folds, and proteome-scale prediction~\cite{jumper2021,varadi2022,kryshtafovych2023}. The model is not infallible: fold-switching proteins~\cite{chakravarty2022,chakravarty2024}, intrinsically disordered regions~\cite{ruff2021}, and many protein complexes and other multi-chain systems remain difficult to predict reliably~\cite{yin2022}. What it does get right, however, places a sharp constraint on what the model must have learned during training. A system that merely memorized a catalog of known folds could not generalize to sequences with no close homologue. A system that learned the \emph{underlying patterns of folding}---the physical and evolutionary patterns that govern how polypeptide chains arrange themselves in three dimensions---could.

The distinction matters. If AlphaFold2's 93 million learned parameters encode something closer to a transferable understanding of protein folding than to a lookup table of solved structures, then those parameters are not just a means to an end. They are a scientific object in their own right: a compressed, data-driven representation of protein folding that no human has written down. The natural question is not only what the model predicts, but what it \emph{knows}.

We argue that these parameters contain more information than standard inference expresses. We probe what the Evoformer's learned weights encode about protein conformational space by systematically perturbing them. We smooth their fine-grained features and attenuate their magnitudes. We call this perturbation scheme \emph{scaled Gaussian convolution} (\sgc{}). The result is a structured readout of what the weights have internalized beyond the single structure they normally converge to.

The encoding correlates with known physical properties of the tested proteins. The conformations that \sgc{} elicits reproduce the topology of the known ubiquitin conformational landscape. They show non-trivial agreement with per-residue flexibility patterns from microsecond-scale equilibrium molecular dynamics simulations of ubiquitin. When the perturbation is deepened progressively, they dismantle native contacts in an order that mirrors the experimentally established stability ordering.

\subsection{Recycling, convergence, and a clue about conformational heterogeneity}
\label{sec:intro-heterogeneity}

AlphaFold2 refines its predictions through a recycling mechanism that feeds each predicted structure back into the model as input for the next iteration. For well-folded proteins, three iterations are typically sufficient. The model approaches an asymptotic confidence limit, beyond which further recycling yields diminishing improvement. Two metrics track this convergence. The predicted local distance difference test (\plddt{}) estimates per-residue lDDT-C$\alpha$ accuracy; the predicted template modeling score (\ptm{}) estimates global TM-score and can indicate confidence in domain packing. Both are the model's self-assessment of how well it has folded the protein. This rapid convergence is efficient, but it compresses the full sequence of structural updates into a handful of steps that are difficult to study.

In our unperturbed runs, a suggestive observation emerges from running many more recycling iterations than are typically used. In proteins with disordered regions, the residues that carry low \plddt{} values are the same ones whose predicted coordinates continue to shift across successive recycles. In the unperturbed regime, these residues also tend to have high crystallographic B-factors and low NMR order parameters. The correlation is not perfect, and \plddt{} is a confidence estimate rather than a physical measurement. Still, the pattern is consistent. Residues that the model struggles to pin down correspond to residues that are genuinely mobile.

This suggested that the model's learned representations encode information about conformational heterogeneity beyond the single structure they converge to. Chang and Perez~\cite{chang2026} recently showed that running AlphaFold2 without MSAs or templates produces iterative folding trajectories. The resulting intermediates and transition-state-ensemble ordering are consistent with experimental $\psi$-value analysis and prior molecular dynamics, indicating that the model has internalized aspects of the folding process itself. If folding knowledge is present in the weights, it is natural to ask what else they encode about protein conformational space. We also want to know whether that encoding can be systematically probed, and how far it extends.

The Evoformer's learned weights shape the internal representations that the structure module ultimately reads to produce a 3D prediction. Modulating those weights and observing how the predicted structure responds is therefore a direct probe of what the weights encode. This probe requires a regime where weight perturbation has a measurable structural effect. Specifically, one where the learned weights, rather than the input alignment alone, shape the prediction (Section~\ref{sec:methods-impl}). \sgc{} does this through three continuous axes applied to the Evoformer's weight tensors. A scaling factor $\lam$ uniformly attenuates weight magnitudes, a Gaussian width $\sig$ selectively smooths fine-grained weight features, and perturbation depth controls how many consecutive Evoformer blocks are modified. The perturbation is deterministic. The same parameters always produce the same output, tracing different paths through the model's learned representational manifold rather than injecting randomness. As we show, these three axes alter both the character and extent of the conformational space that the weights reveal.

\begin{figure}[!htbp]
    \centering
    \includegraphics[width=\textwidth]{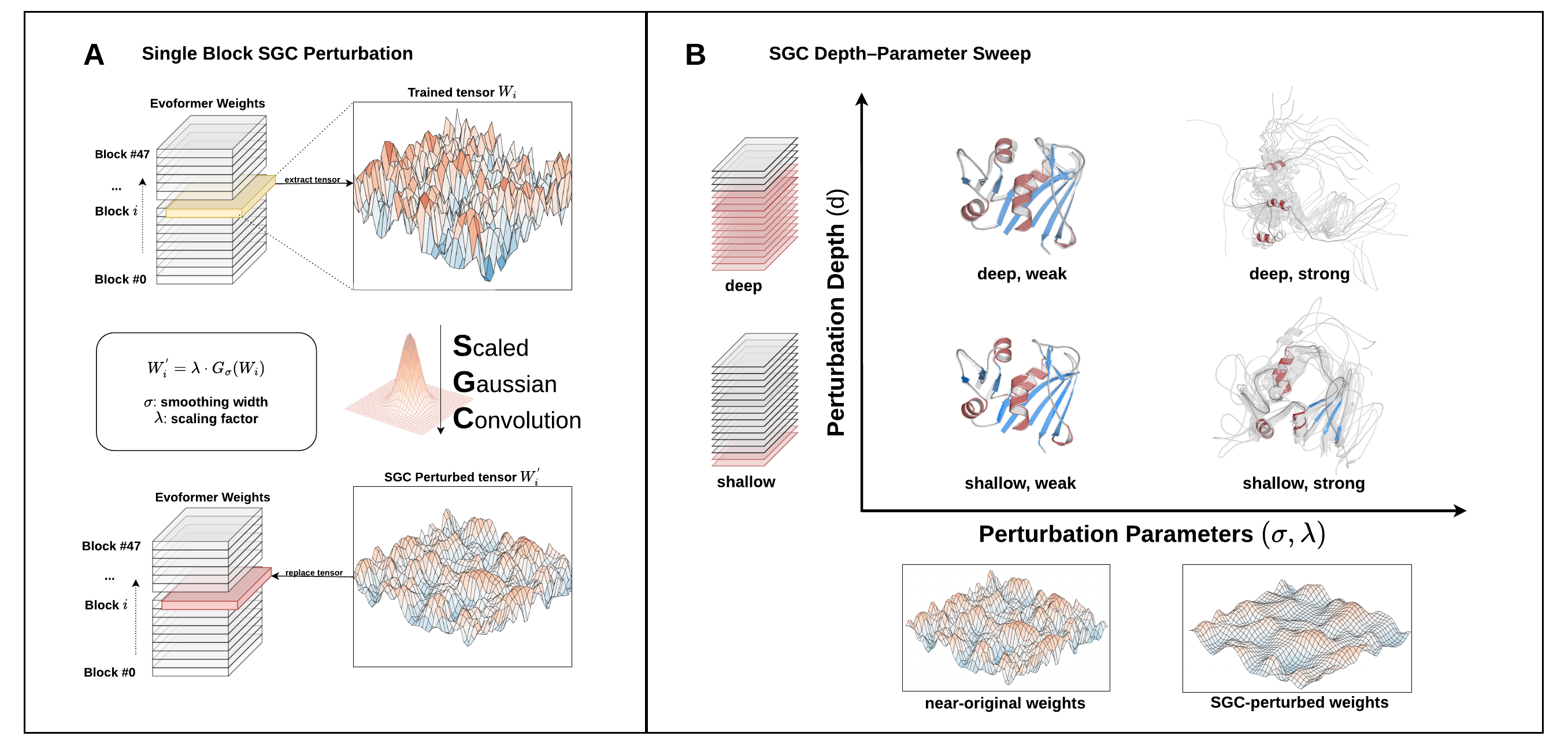}
    \caption{\textbf{Scaled Gaussian convolution (SGC) perturbs Evoformer weights to probe encoded conformational structure.}
    \textbf{(A)} SGC shown for one representative Evoformer block $i$. A trained weight tensor $W_i$ is extracted, smoothed by Gaussian convolution with width $\sig$, scaled by $\lam$, and replaced as $W'_i = \lam G_\sig(W_i)$ before inference. In depth-$d$ experiments, the same operation is applied cumulatively to blocks $0,\ldots,d{-}1$.
    \textbf{(B)} The perturbation sweep varies depth $d$ and parameters $(\sig,\lam)$, mapping how structural predictions change as progressively more Evoformer blocks are modified and the weight tensors are moved farther from their trained values.}
    \label{fig:method}
\end{figure}

Several other approaches have explored perturbation to induce conformational diversity in AlphaFold2. Input-side MSA perturbation methods such as MSA subsampling~\cite{delalamo2022}, AFsample2~\cite{kalakoti2025} (column masking), and Wayment-Steele et al.~\cite{waymentsteele2024} (sequence clustering) modify the model's input to increase prediction variance. A separate line of work fine-tunes the model with a generative objective to sample ensembles directly, as in AlphaFlow~\cite{jing2024}. AlphaFold-RandomWalk~\cite{taneja2026} adds structured stochastic noise to AlphaFold's Evoformer weights. It produces conformations that it both analyzes directly and clusters to seed unbiased molecular dynamics for free-energy landscape estimation. \sgc{} is deterministic, its parameters are empirically interpretable, and its conformations are analyzed directly rather than being used to seed a downstream MD workflow. Matched noise controls establish that these distinctions are substantive. White noise at the same magnitude produces atomized debris with broken polypeptide chains; spectral noise produces bimodal, seed-dependent outcomes. Neither reproduces \sgc{}'s graded, seed-robust structural progression. The conformational response therefore depends on structure already present in the learned weights. \sgc{} provides a deterministic readout of that response.

We apply \sgc{} to three systems, each chosen to probe a different regime of what AlphaFold2 has learned. Ubiquitin is a well-characterized globular protein where the encoding shows strong physical correspondence. KaiB is a fold-switching protein where the encoding appears incomplete. $\alpha$-synuclein is an intrinsically disordered protein where the SGC readout has not converged. The responses map a spectrum of learned representational quality that is itself informative about the boundaries of AlphaFold2's learned representation.

\section{Methods}
\label{sec:methods}

A note on terminology: throughout this paper, ``conformational ensemble'' refers to the collection of structures produced by \sgc{} at a given parameter setting. Each structure is an independent prediction, not a time point in a trajectory. We make no claims about kinetic ordering, transition rates, or Boltzmann weighting.

Two further terminological distinctions recur. ``Experimentally established stability ordering'' refers to the ordering of ubiquitin contact groups (G1 last, G2 first under chemical denaturation~\cite{went2005,sosnick2004}). This is a property of the protein, not a claim about \sgc{}'s outputs. ``Conformational landscape'' refers to the topological structure of the accessible space without implying Boltzmann weighting, free-energy quantification, or kinetic ordering.

\subsection{Scaled Gaussian convolution}
\label{sec:methods-sgc}

AlphaFold2's Evoformer consists of 48 sequential blocks, each containing attention mechanisms and feedforward networks that iteratively refine two internal representations: one encoding relationships between sequences in the multiple sequence alignment, and another encoding pairwise relationships between residues. Each block contributes to these representations through learned weight tensors. These matrices govern how information is routed, combined, and transformed at each step. \sgc{} perturbs these tensors directly.

For each selected block $i$, every weight tensor $W_i$ is replaced with:
\begin{equation}
    W'_i = \lam \cdot G_\sig(W_i)
\end{equation}
where $G_\sig$ denotes convolution with a Gaussian kernel of width $\sig$ and $\lam$ is a uniform scaling factor. The Gaussian smoothing replaces each weight with a local average of its neighbors in the weight tensor, preserving coarse, global patterns while dissolving fine-grained local structure. At $\sig = 0.30$, this smoothing is mild. The kernel has a center weight of 0.992 and extends only to nearest neighbors (kernel width~3). The scaling factor uniformly attenuates all weight magnitudes, reducing each block's influence on the representations it updates. Together, the two operations modulate both the \emph{strength} (how much each block contributes) and the \emph{character} (the local structure of what it computes) of each block's processing.

\sgc{} is parameterized by three continuous axes:
\begin{itemize}
    \item \textbf{Scaling factor} $\lam \in (0, 1]$: Controls overall weight magnitude. At $\lam = 1$, weights retain their original scale; as $\lam$ decreases, each block's output is progressively attenuated. Scaling dominates the total perturbation. Decomposing the perturbation into a scaling component and a smoothing residual,
    \begin{align}
        \Delta W &= \underbrace{(\lam - 1)\,W}_{\Delta_{\text{scale}}} \;+\; \underbrace{\lam\bigl(G_\sig(W) - W\bigr)}_{\Delta_{\text{blur}}} \label{eq:decomp}
    \end{align}
    the scaling component accounts for $>$99\% of $\|\Delta W\|^2$ at $\sig = 0.30$, $\lam = 0.75$ (per-tensor median 99.8\%, range 99.5--100\%, computed across all 93 weight tensors per block and 48 blocks; Supplementary Section~\ref{sec:supp-power-decomp}). Although $\Delta_{\text{blur}}$ contributes $<$1\% of perturbation power and is itself largely parallel to $\Delta_{\text{scale}}$, the component orthogonal to $\Delta_{\text{scale}}$ is geometrically distinct from uniform scaling and, as we show in Section~\ref{sec:results}, has outsized effects at the structural transition boundary.

    \item \textbf{Gaussian width} $\sig \geq 0$: Controls the degree of local weight smoothing. At $\sig = 0$, no smoothing occurs and \sgc{} reduces to pure uniform scaling. As $\sig$ increases, the smoothing kernel broadens and progressively dissolves fine-grained weight structure. Although small in magnitude at our primary operating point ($\sig = 0.30$), this smoothing component has a part orthogonal to uniform scaling that is geometrically distinct from it. Where $\lam$ preserves the relative structure within each weight matrix at lower amplitude, smoothing alters the geometry of the computation itself, changing local weight correlations rather than merely attenuating them uniformly. At $\sig = 0.30$ this effect is small in magnitude but has disproportionate structural consequences at the transition boundary (Section~\ref{sec:results-sigma-lambda}; Supplementary Section~\ref{sec:supp-power-decomp}; Supplementary Fig.~S18).

    \item \textbf{Perturbation depth} $d \in \{1, \ldots, 48\}$: The number of consecutive Evoformer blocks modified, counting from block~0. At depth~$d$, blocks $\{0, 1, \ldots, d{-}1\}$ are perturbed and blocks $\{d, \ldots, 47\}$ retain their trained weights. This is the primary controllability axis: depth governs how much of the Evoformer's sequential processing is affected.
\end{itemize}

The perturbation is applied once, before inference, directly to the model parameters. It is fully deterministic. Given fixed inputs and random seed, the same ($\sig$, $\lam$, $d$) triple always produces identical output.

\begin{algorithm}[H]
\caption{Scaled Gaussian Convolution}\label{alg:sgc}
\KwIn{Model $M$ with Evoformer blocks $B_0, \ldots, B_{47}$;\newline
      parameters $\sig$ (Gaussian width), $\lam$ (scaling factor), $d$ (depth);\newline
      input features (sequence, MSA)}
\KwOut{Predicted structures}
checkpoint $\leftarrow \{B_i.\text{parameters}() \;\text{for}\; i = 0 \ldots 47\}$\;
\For{block $i \in \{0, 1, \ldots, d{-}1\}$}{
    \ForEach{weight tensor $W \in B_i.\text{parameters}()$}{
        $W \leftarrow \lam \cdot \text{GaussianFilter}(W, \sig)$\;
    }
}
structures $\leftarrow M(\text{input features})$\;
Restore weights from checkpoint\;
\Return{structures}
\end{algorithm}

\subsection{Noise controls}
\label{sec:methods-noise}

To establish that \sgc{}'s effects depend on the specific structure of the perturbation rather than its magnitude alone, we designed two matched noise controls. Both deliver the same total perturbation power as \sgc{} on a per-tensor basis, isolating the role of spectral structure and phase coherence.

\textbf{White noise control.} Each weight tensor receives additive Gaussian noise $\eta \sim \mathcal{N}(0, \sigma_n^2)$, where $\sigma_n^2$ is set so that $\|\eta\|^2 = \|\lam \cdot G_\sig(W) - W\|^2$ for each tensor independently. This preserves the total perturbation energy while destroying all structure---both spectral and phase.

\textbf{Spectral noise control.} The perturbation $\Delta W = \lam \cdot G_\sig(W) - W$ is computed, transformed to the frequency domain via FFT, and its magnitude spectrum is preserved while the phases are replaced with uniform random values. The inverse FFT yields a perturbation with identical power distribution across frequencies but no coherent relationship to the original weights. This isolates whether the spectral \emph{shape} of \sgc{}'s perturbation, as opposed to its phase coherence, is responsible for the structured output. Note that the FFT assumes periodic boundaries along each tensor axis; since Evoformer weight tensors carry no natural spatial periodicity, this control tests sensitivity to phase structure specifically, not to general spectral properties.

These three conditions form a clean decomposition:

\begin{table}[h]
\centering\small
\begin{tabular}{llll}
\toprule
Condition & Spectral shape & Phase coherence & Result \\
\midrule
\sgc{} & Smoothing profile & Coherent with weights & Tunable, structured \\
Spectral noise & Smoothing profile & Randomized & Bimodal, seed-dependent \\
White noise & Flat & Randomized & Stochastic collapse \\
\bottomrule
\end{tabular}
\end{table}

Power matching is performed per-tensor rather than globally, because Evoformer blocks contain approximately 100 parameter tensors spanning three orders of magnitude in size. Global matching would systematically over-perturb small tensors. All noise controls use a dedicated random number generator seeded independently of the model's inference state, ensuring reproducibility. Implementation details and pseudocode for the noise control procedures are provided in Supplementary Methods.

\subsection{Implementation}
\label{sec:methods-impl}

We implemented \sgc{} within OpenFold v2.2.0~\cite{ahdritz2024openfold}, an open-source PyTorch reimplementation of AlphaFold2. Perturbation is applied in-place to the live model parameters for each selected Evoformer block. Original parameter values are saved to a checkpoint dictionary before perturbation and restored after each inference run, allowing sequential experiments without reloading the model. Input file integrity is verified via SHA-256 checksums of the parameter weights, FASTA sequence, and MSA alignment files, recorded in per-run provenance manifests. All experiments were run without structural templates (empty template features), isolating the Evoformer's learned weights from template-derived structural information.

All experiments use the \texttt{model\_1\_ptm} parameter set (93M parameters, 48 Evoformer blocks) unless otherwise noted. Model independence is verified by repeating key experiments across all five AlphaFold2 parameter sets (Section~\ref{sec:results-model-indep}). Inference uses 32 recycling iterations per run (yielding 33 structures, including the initial prediction), providing a detailed view of the model's iterative refinement process. For intrinsically disordered proteins, where the model does not converge to a single structure, the extended recycling window captures a broader range of the model's iterative behaviour. Random seeds control MSA subsampling; all multi-seed experiments use seeds 42, 43, and 44 with deterministic CUDA operations enabled.

All primary experiments use MSA depth $64 \times 64$ rather than the default $512 \times 5{,}120$. Across the reduced MSA depths tested here, increasing depth suppresses the structural response, whereas shallower MSAs make the model more sensitive to perturbation by reducing the external coevolutionary constraint. We therefore use reduced MSA depth to place the model in a regime where perturbing the Evoformer weights has a measurable effect. Across MSA depths 32, 64, and 128, the transition regime, contact-loss ordering, and landscape topology are preserved, but sensitivity decreases monotonically with depth (Supplementary Section~\ref{sec:supp-msa-depth}; Supplementary Fig.~\ref{sfig:msa-sensitivity}). We choose $64 \times 64$ because it sharply reduces atomization relative to MSA~32 while remaining appreciably responsive relative to MSA~128. Behaviour at default MSA depth, which may require stronger perturbations, remains an open question. Chang and Perez~\cite{chang2026} showed the limiting case by removing MSAs and templates entirely; related input-side MSA perturbation methods include AFsample2~\cite{kalakoti2025} and Wayment-Steele et al.~\cite{waymentsteele2024}. As a practical benefit, the reduced depth decreases inference time by approximately $4\times$, enabling the scale of parameter exploration reported here.

Inference was performed on two consumer-grade laptop GPUs: an NVIDIA GeForce RTX 5090 Mobile (175W, 10{,}496 CUDA cores) and an NVIDIA GeForce RTX 2060 Mobile (80W, 1{,}920 CUDA cores). At MSA depth $64 \times 64$ for a 76-residue protein (ubiquitin), a single 32-recycle inference takes approximately 10.5~seconds on the RTX~5090 and approximately 46~seconds on the RTX~2060. Perturbation setup across all 48 blocks takes 1.1~seconds on either GPU. Full experiment specifications, YAML configuration files, and orchestration details are provided in Supplementary Methods.

\subsection{Protein systems}
\label{sec:methods-proteins}

We selected three proteins to probe different regimes of AlphaFold2's learned representation:

\textbf{Ubiquitin} (PDB: 1UBQ; UniProt: P0CG48; 76 residues). A small, thermostable globular protein with a five-stranded $\beta$-grasp fold. Ubiquitin has been extensively characterized by NMR spectroscopy, X-ray crystallography, and molecular dynamics simulation, including millisecond-scale folding trajectories and microsecond-scale equilibrium simulations from D.E.\ Shaw Research. Its folding pathway, experimentally established stability ordering, and per-residue flexibility are well established, making it an ideal system for quantitative validation of what the Evoformer has learned.

\textbf{KaiB} (PDB: 2QKE/5JYT; UniProt: Q79PF5; 102 residues). A cyanobacterial circadian clock protein that undergoes a dramatic fold switch between an unusual $\alpha/\beta$ ground-state fold (PDB: 2QKE) and a thioredoxin-like fold-switched state (PDB: 5JYT). AlphaFold2 consistently predicts only the fold-switched state. KaiB tests whether \sgc{} can access conformational states that the model may not have learned, or whether the method is limited to reading out what the weights already encode.

\textbf{$\alpha$-synuclein} (UniProt: P37840; 140 residues). An intrinsically disordered protein (IDP) implicated in Parkinson's disease, with no unique native structure. IDPs are known to be challenging for structure prediction methods because they lack a single PDB-like structural target for AlphaFold2 to produce. Microsecond-scale MD trajectories exist for $\alpha$-synuclein across multiple force fields, but no experimental consensus ensemble is available, making this system a test of what \sgc{} reveals at the boundary of AlphaFold2's training distribution.

Canonical sequences, multiple sequence alignments, and per-protein experimental references are provided in Supplementary Table~S1.

\subsection{Structural analysis}
\label{sec:methods-structural}

All structural metrics are computed from the predicted C$\alpha$ coordinates after Kabsch superposition~\cite{kabsch1976} onto the reference structure, using MDAnalysis~\cite{mdanalysis2011} for alignment and coordinate manipulation.

\textbf{RMSD.} Root mean square deviation of C$\alpha$ positions to the native crystal structure.

\textbf{Q-factor.} Fraction of native contacts preserved, computed using the Best--Hummer--Eaton formulation~\cite{best2013} as implemented in mdtraj~\cite{mcgibbon2015mdtraj}. Two contact definitions are used in this work, tailored to different analyses:
\begin{itemize}
    \item \emph{Landscape and survey analyses} (Sections~\ref{sec:results-transition}--\ref{sec:results-topology}): heavy-atom contacts at 4.5~\AA{} distance cutoff, minimum sequence separation of 3 residues (884 contact pairs for ubiquitin). This follows the standard BHE convention.
    \item \emph{Unfolding pathway analysis} (Section~\ref{sec:results-unfolding}): C$\alpha$--C$\alpha$ contacts at 12~\AA{} cutoff, minimum sequence separation of 3 residues. The broader cutoff captures the same secondary-structure contacts identified in the folding literature, where contacts are typically reported as residue pairs rather than atom pairs.
\end{itemize}

\textbf{Radius of gyration ($R_g$).} Computed from C$\alpha$ positions as a measure of overall compactness.

\textbf{End-to-end distance ($R_{ee}$).} Euclidean distance between the N- and C-terminal C$\alpha$ atoms. Used alongside $R_g$ for intrinsically disordered proteins where no reference structure exists.

\textbf{Perturbation sensitivity.} For each perturbation condition, we measure how much each residue's predicted C$\alpha$ position varies across the structures produced by recycling. All structures are Kabsch-aligned to the native crystal structure, and per-residue positional spread is computed as the root mean square fluctuation (RMSF) from the ensemble mean of the aligned coordinates. We compute this separately at each perturbation depth, pooling across the 33 structures from a single run's recycling sequence, and report mean $\pm$ standard deviation across 3 seeds. We refer to this quantity as \emph{perturbation sensitivity} to distinguish it from physical flexibility. The former reflects what the model's weights appear to encode about a residue's conformational freedom; the latter measures actual conformational freedom in solution (see Section~\ref{sec:results-rmsf} for comparison). The same alignment and RMSF protocol is applied when computing MD flexibility from equilibrium trajectories, ensuring that differences reflect the ensembles rather than the analysis pipeline.

\textbf{Weighted contact number (WCN).} $\text{WCN}_i = \sum_{j \neq i} r_{ij}^{-2}$ measures residue burial and serves as a geometric baseline for partial-correlation analysis (Section~\ref{sec:results-rmsf}).

\textbf{Territorial overlap.} To quantify how much of the MD-accessible conformational space is also reached by \sgc{}, we use a $k$-nearest-neighbor support estimator that avoids arbitrary histogram binning. Both axes of the RMSD $\times$ $Q$ plane are affine-normalized to pooled range. For each MD point, the distance to its $k$-th nearest MD neighbor defines a local support radius; the 99th percentile of these radii defines the occupied territory of each dataset ($k = 20$). Coverage is then the area-weighted fraction of MD territory in which the AF2 dataset also has $k$ neighbors within the corresponding AF2 support threshold. Robustness to $k$ is documented in Supplementary Section~\ref{sec:supp-territory}.

\textbf{TM-score.} For KaiB, we use the template modeling score~\cite{zhang2005} as implemented in \texttt{tmtools}~\cite{tmtools} to assess structural similarity to the ground-state (PDB: 2QKE, chain~B) and fold-switched (PDB: 5JYT, chain~A) reference structures. TM-score $> 0.5$ indicates the same fold; the two references score 0.42 against each other, confirming distinct folds.

All primary structural analyses use raw, unminimized AlphaFold2 predictions. We developed restrained vacuum minimization pipelines in OpenMM~8.4.0 with AMBER14SB and CHARMM36. We do not use minimized structures in the main analyses because hydrogen placement and force-field choice would confound attribution of structural features to the learned weights. As a diagnostic, we minimized the full forward cumulative sweep at the primary operating point together with the matched unperturbed baseline. Outside the documented depth-3 block-2 anomaly, raw and minimized structures differ only modestly (median C$\alpha$ displacement 0.64~\AA{}; 90th percentile 0.94~\AA{}; Supplementary Fig.~S28; Supplementary Table~S8). We therefore report raw-structure observables throughout; the full depth-resolved validation is provided in Supplementary Section~\ref{sec:supp-unfolding-multicondition}.

\subsection{Molecular dynamics reference data}
\label{sec:methods-md}

We compare \sgc{} ensembles against published molecular dynamics trajectories from D.E.\ Shaw Research. Two datasets serve complementary roles.


\textbf{Ubiquitin unfolding (390~K).} Six independent millisecond-scale simulations of ubiquitin at 390~K, each starting from the native structure, from Piana, Lindorff-Larsen \& Shaw~\cite{piana2013}. Five trajectories span 1.14--1.24~ms (57.8~\AA{} box) and one is 0.68~ms (66.1~\AA{} box), totaling approximately 6.7~ms of simulation time. After striding every 10th frame, this yields 673{,}519 frames. Full-protein coordinates are used for heavy-atom $Q$-factor computation; C$\alpha$-only coordinates are used for RMSD and the supplementary C$\alpha$ $Q$-factor analysis. These trajectories reversibly sample states from the native fold through the denatured ensemble at 390~K, providing folding funnel coverage for landscape topology comparisons.

\textbf{Ubiquitin equilibrium (300~K).} Six equilibrium MD datasets from Robustelli, Piana \& Shaw~\cite{robustelli2018} span four AMBER-family force fields (a99SB-disp, a99SB-ILDN/TIP4P-D, a99SB*-ILDN/TIP3P, a99SB-UCB) and two CHARMM-family force fields (CHARMM22*/TIP3P, CHARMM36m). All used the NPT ensemble at 300~K with frames saved every 200~ps. Five contribute 8--10.5~$\mu$s each, while the CHARMM22*/TIP3P dataset provides 154.6~$\mu$s (77{,}280 frames after $10\times$ striding) and is therefore used as the primary RMSF reference. Of the six, five maintain native-like sampling (mean RMSD 2.8--3.3~\AA{}). The sixth, a99SB-UCB, shows extensive unfolding (mean RMSD $9.9 \pm 8.8$~\AA) and serves as an out-of-basin control. The remaining four native-state force fields provide a robustness check across force-field families. The correlation between \sgc{} perturbation sensitivity and MD RMSF is consistent across all five references, though the three AMBER variants are near-identical to each other (pairwise $r > 0.99$; Supplementary Fig.~S4).

Per-residue RMSF from the equilibrium trajectories serves as the reference for flexibility validation (Section~\ref{sec:results-rmsf}). Landscape comparisons (Section~\ref{sec:results-topology}) use the 390~K simulations for folding funnel coverage and the two CHARMM-family 300~K equilibrium datasets (82{,}280 frames total) for native-basin containment.

\subsection{Experiment summary}
\label{sec:methods-experiments}

We explored the \sgc{} parameter space systematically. For ubiquitin, a dense grid spanning $\sig \in \{0.00\} \cup [0.20, 0.30]$ at 0.01 intervals (12 values) and $\lam \in [0.70, 0.85]$ at 0.01 intervals (16 values) was run at all 48 perturbation depths with 3 seeds. This maps the transition boundary between native-like and unfolded predictions at high resolution (192 $(\sig, \lam)$ conditions $\times$ 48 depths $\times$ 3 seeds $= 27{,}648$ runs). To assess MSA depth sensitivity, a subset of conditions was repeated at MSA depths $32 \times 32$ and $128 \times 128$. Model independence was tested by repeating a representative subset of conditions across all five AlphaFold2 parameter sets (models~2--5). A reverse perturbation scheme (perturbing blocks $\{48{-}d, \ldots, 47\}$ instead of $\{0, \ldots, d{-}1\}$, so that $d$ denotes the number of perturbed blocks in both schemes) was also run at ($\sig = 0.30$, $\lam = 0.75$). Depth-resolved forward and reverse profiles are compared in Section~\ref{sec:results-transition} and Supplementary Section~\ref{sec:supp-forward-reverse}. All subsequent analyses use the forward scheme, which produces the richer conformational response.

Of the conditions surveyed, ($\sig = 0.30$, $\lam = 0.75$) produces the widest range of conformational response across perturbation depths while maintaining native-like structure at shallow depths; we use it as the primary condition for detailed analyses throughout this work. For KaiB and $\alpha$-synuclein, we selected this condition along with two others from the dynamics-relevant region: (0.30,~0.85) and (0.20,~0.75), and ran all five models across 48 depths and 3 seeds. Ubiquitin and KaiB used 32 recycling iterations, which we observed to be sufficient for the predicted structure to stabilize. $\alpha$-synuclein did not converge within the same 32-iteration window. Predicted coordinates continued to shift without reaching a plateau, so we doubled the recycling budget to 64 iterations, consistent with the absence of a single dominant structure for an IDP.

Noise controls used the primary \sgc{} condition ($\sig = 0.30$, $\lam = 0.75$) at all 48 depths across 3 seeds, with two matched controls (white noise and spectral noise) and two dosage variants (half-power and double-power white noise; Supplementary Methods).

\begin{table}[H]
\caption{\textbf{Summary of all experiments.} Each row's run count equals (conditions $\times$ depths $\times$ seeds $\times$ models). Each ubiquitin and KaiB run produces 33 structures (32 recycling iterations plus initial); each $\alpha$-synuclein run produces 65 (64 recycles).}
\label{tab:experiments}
\centering\footnotesize
\setlength{\tabcolsep}{3pt}
\begin{tabular}{@{}lllllcccrr@{}}
\toprule
 & Experiment & MSA & ($\sig$, $\lam$) conditions & Depths & Seeds & Models & Runs & Structures \\ 
\midrule
\multicolumn{9}{@{}l}{\textbf{Ubiquitin --- Model~1}} \\
 & Parameter grid & $64^2$ & $192 = 12\sig \times 16\lam$$^a$ & 48 & 3 & 1 & 27{,}648 & 912{,}384 \\
 & MSA depth sweep & $32^2$ & $14$$^b$ & 48 & 3 & 1 & 2{,}016 & 66{,}528 \\
 & MSA depth sweep & $128^2$ & $14$$^b$ & 48 & 3 & 1 & 2{,}016 & 66{,}528 \\
 & Reverse & $32^2\!/64^2\!/128^2$ & 1: (0.30, 0.75) & 48 & 3 & 1 & 432$^c$ & 14{,}256 \\
 & Baselines & $32^2\!/64^2\!/128^2$ & --- & --- & 3 & 1 & 9$^c$ & 297 \\
\cmidrule{8-9}
 & & & & & \multicolumn{2}{r}{\textit{Model~1 subtotal}} & 32{,}121 & 1{,}059{,}993 \\[2pt]
\multicolumn{9}{@{}l}{\textbf{Ubiquitin --- Models~2--5} (per model; $\times 4$)} \\
 & Cumulative & $64^2$ & $14$$^b$ & 48 & 3 & 1 & 2{,}016 & 66{,}528 \\
 & Cumulative & $32^2$ & $12$$^d$ & 48 & 3 & 1 & 1{,}728 & 57{,}024 \\
 & Cumulative & $128^2$ & $12$$^d$ & 48 & 3 & 1 & 1{,}728 & 57{,}024 \\
 & Reverse & $64^2$ & 1: (0.30, 0.75) & 48 & 1 & 1 & 48 & 1{,}584 \\
 & Baselines & $32^2\!/64^2\!/128^2$ & --- & --- & 3 & 1 & 9$^c$ & 297 \\
\cmidrule{8-9}
 & & & & & \multicolumn{2}{r}{\textit{Per-model subtotal}} & 5{,}529 & 182{,}457 \\
 & & & & & \multicolumn{2}{r}{\textit{$\times 4$ models}} & 22{,}116 & 729{,}828 \\[2pt]
\multicolumn{9}{@{}l}{\textbf{Ubiquitin --- Noise controls} (Model~1, $64^2$)} \\
 & White noise & $64^2$ & 1: matched to (0.30, 0.75) & 48 & 3 & 1 & 144 & 4{,}752 \\
 & Spectral noise & $64^2$ & 1: matched to (0.30, 0.75) & 48 & 3 & 1 & 144 & 4{,}752 \\
 & White $\tfrac{1}{2}\!\times$\,/\,$2\!\times$ & $64^2$ & 2 dosage variants & 48 & 3 & 1 & 288 & 9{,}504 \\
\cmidrule{8-9}
 & & & & & \multicolumn{2}{r}{\textit{Noise subtotal}} & 576 & 19{,}008 \\
\midrule
 & & & & & \multicolumn{2}{r}{\textbf{Ubiquitin total}} & \textbf{54{,}813} & \textbf{1{,}808{,}829} \\
\midrule
\multicolumn{2}{@{}l}{\textbf{KaiB}} & $64^2$ & 3$^e$ & 48 & 3 & 5 & 2{,}160 & 71{,}280 \\
 & Baselines & $64^2$ & --- & --- & 3 & 5 & 15 & 495 \\
\cmidrule{8-9}
 & & & & & \multicolumn{2}{r}{\textbf{KaiB total}} & \textbf{2{,}175} & \textbf{71{,}775} \\
\midrule
\multicolumn{2}{@{}l}{\textbf{$\alpha$-synuclein}} & $64^2$ & 3$^e$ & 48 & 3 & 5 & 2{,}160 & 140{,}400 \\
 & Baselines & $64^2$ & --- & --- & 3 & 5 & 15 & 975 \\
\cmidrule{8-9}
 & & & & & \multicolumn{2}{r}{\textbf{$\alpha$-syn.\ total}} & \textbf{2{,}175} & \textbf{141{,}375} \\
\midrule
 & & & & & \multicolumn{2}{r}{\textbf{Grand total}} & \textbf{59{,}163} & \textbf{2{,}021{,}979} \\
\bottomrule
\end{tabular}

\vspace{0.5em}
\begin{flushleft}\footnotesize
$^a$\,$\sig \in \{0.00\} \cup [0.20, 0.30]$ at $0.01$ intervals (12 values); $\lam \in [0.70, 0.85]$ at $0.01$ intervals (16 values). \\
$^b$\,14 conditions: $\sig \in \{0.20, 0.25, 0.30\} \times \lam \in \{0.65, 0.75, 0.85, 0.95\}$ plus $(\sig{=}0.00, \lam{=}0.75)$ and $(\sig{=}0.10, \lam{=}0.75)$. \\
$^c$\,Spans 3 MSA depths counted as separate runs: $3 \times 3 = 9$ (baselines), $3 \times 48 \times 3 = 432$ (reverse). \\
$^d$\,12 conditions: footnote~$b$ minus $(\sig{=}0.00, \lam{=}0.75)$ and $(\sig{=}0.10, \lam{=}0.75)$, run only at MSA $64^2$ for models~2--5. \\
$^e$\,$(0.20, 0.75)$, $(0.30, 0.75)$, $(0.30, 0.85)$.
\end{flushleft}
\end{table}

All runs include per-run provenance records with SHA-256 input checksums, perturbation specifications, and environment metadata. The full experiment corpus is indexed in machine-readable format for programmatic access.


\section{Results}
\label{sec:results}


\subsection{Perturbation depth induces a sharp structural transition}
\label{sec:results-transition}

All results in Sections~\ref{sec:results-transition}--\ref{sec:evo_pca} use ubiquitin (76 residues) as the test system; KaiB and $\alpha$-synuclein follow in Sections~\ref{sec:results-kaib}--\ref{sec:results-asyn}.

We begin with the primary controllability axis, perturbation depth. At the reference condition ($\sig = 0.30$, $\lam = 0.75$, MSA $64 \times 64$), we perturbed blocks $\{0, \ldots, d{-}1\}$ for each depth $d$ from 1 to 48 and measured the structural response of ubiquitin across 3 seeds.

\begin{figure}[!htbp]
    \centering
    \includegraphics[width=\textwidth]{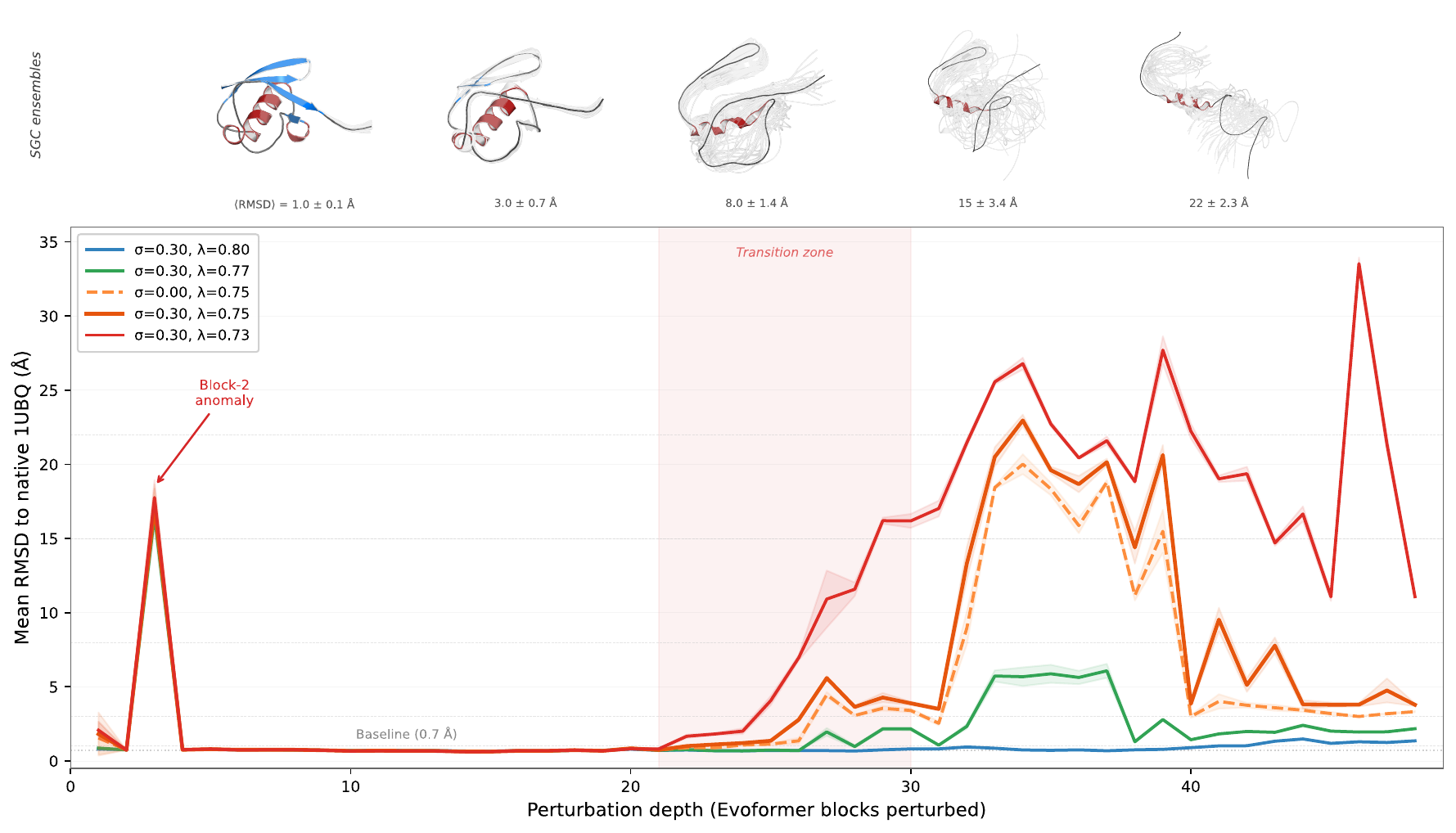}
    \caption{\textbf{SGC perturbation induces a tunable structural transition.} Mean C$\alpha$ RMSD to the native crystal structure as a function of perturbation depth for five representative ($\sig$, $\lam$) conditions (model~1, MSA $64 \times 64$, 3 seeds). $\lam$ controls the transition onset: at $\lam = 0.80$ (blue) the structure remains native-like at all depths; at $\lam = 0.73$ (red) collapse begins early. At matched $\lam = 0.75$, Gaussian smoothing ($\sig = 0.30$, solid orange) amplifies the perturbation beyond pure scaling ($\sig = 0.00$, dashed orange). The block-2 anomaly at depth~3 is visible across all conditions. Top: representative structural ensembles at key depths, aligned to native, showing progressive loss of secondary structure.}
    \label{fig:rmsd_vs_depth}
\end{figure}

The response is not uniform across depth (Figure~\ref{fig:rmsd_vs_depth}). Four regimes emerge:

\textbf{Enhancement zone (depths 4--20).} Perturbing up to 20 blocks leaves the predicted structure within 0.65--0.85~\AA{} of the native crystal structure (baseline: 0.72~\AA{}). The fraction of native contacts preserved remains high ($Q \approx 0.81$, baseline 0.806), and the radius of gyration is unchanged at 11.4~\AA{}. Over most of this range, the perturbed predictions are marginally \emph{closer} to the crystal structure than the unperturbed baseline, suggesting that mild perturbation may act as a regularizer on the structural variance that reduced MSA depth introduces.

\textbf{Block-2 anomaly (depth 3).} Perturbing exactly the first three blocks ($\{0, 1, 2\}$) produces catastrophic structural collapse: RMSD jumps to 17.2~\AA{} and $Q$ drops to 0.135 in the midst of an otherwise stable enhancement zone. Perturbing two blocks (depth~2) or four blocks (depth~4) produces no such disruption. The anomaly is specific to block~2 of the Evoformer and is reproduced across all three seeds. Including block~3 in the perturbation abolishes the anomaly. We exclude depths $\leq 5$ from the max-RMSD and min-$Q$ summaries in Sections~\ref{sec:results-boundary}--\ref{sec:results-topology} to avoid conflating this isolated anomaly with the systematic depth-dependent response.

\textbf{Possible mechanism.} The anomaly is strongly asymmetric: reverse perturbation of the last three Evoformer blocks ($\{45, 46, 47\}$) leaves ubiquitin near baseline (RMSD $\sim 0.78$~\AA{}), so the effect is specific to the early forward computation. The noise controls further suggest that the failure is not generic random damage. White and spectral noise at matched power produce stochastic, seed-dependent collapses at depth~3, whereas \sgc{} produces a deterministic all-seed collapse at the reference condition (Section~\ref{sec:results-noise}). We therefore interpret the anomaly provisionally as an early-interface compensation failure: perturbing blocks $\{0, 1, 2\}$ may shift the MSA--pair representation into a state that the unperturbed block~3, operating at full weight strength, amplifies or cannot process, while perturbing block~3 as well restores a self-consistent computation and returns the model to the smooth depth profile. This interpretation remains hypothetical. Distinguishing block-local fragility from boundary compensation will require windowed, single-block, and per-head perturbations, which we leave to follow-up work.

\textbf{Transition zone (depths 22--30).} Beyond depth~20, the structure begins to deform. RMSD rises from 0.85~\AA{} at depth~20 through 1.00~\AA{} at depth~22, 1.36~\AA{} at depth~25, and 3.89~\AA{} at depth~30---a 4.6-fold increase over ten blocks. $Q$ drops correspondingly from 0.82 to 0.39. The transition is continuous but steep. The model tolerates perturbation of the first 20 blocks with little effect, then loses structural integrity rapidly as the remaining unperturbed blocks can no longer compensate. The steepness of this transition suggests that, under perturbation, the Evoformer's learned processing divides into a tolerant early segment and a sensitive late segment, separated by a boundary in the low twenties.

\textbf{Deep perturbation and tail recovery (depths 30--48).} Beyond depth~30, RMSD reaches 23.0~\AA{} (at depth~34) and structures are largely unfolded. But a partial recovery occurs when nearly all blocks are perturbed. At depths 44--48, RMSD decreases to 3--5~\AA{} and the radius of gyration contracts back toward the native value. One interpretation is that the \emph{mismatch} between perturbed and unperturbed blocks, rather than perturbation per se, contributes to the disruption. When all blocks are uniformly attenuated, the model partially recovers folding ability.

\begin{figure}[!htbp]
    \centering
    \includegraphics[width=\textwidth]{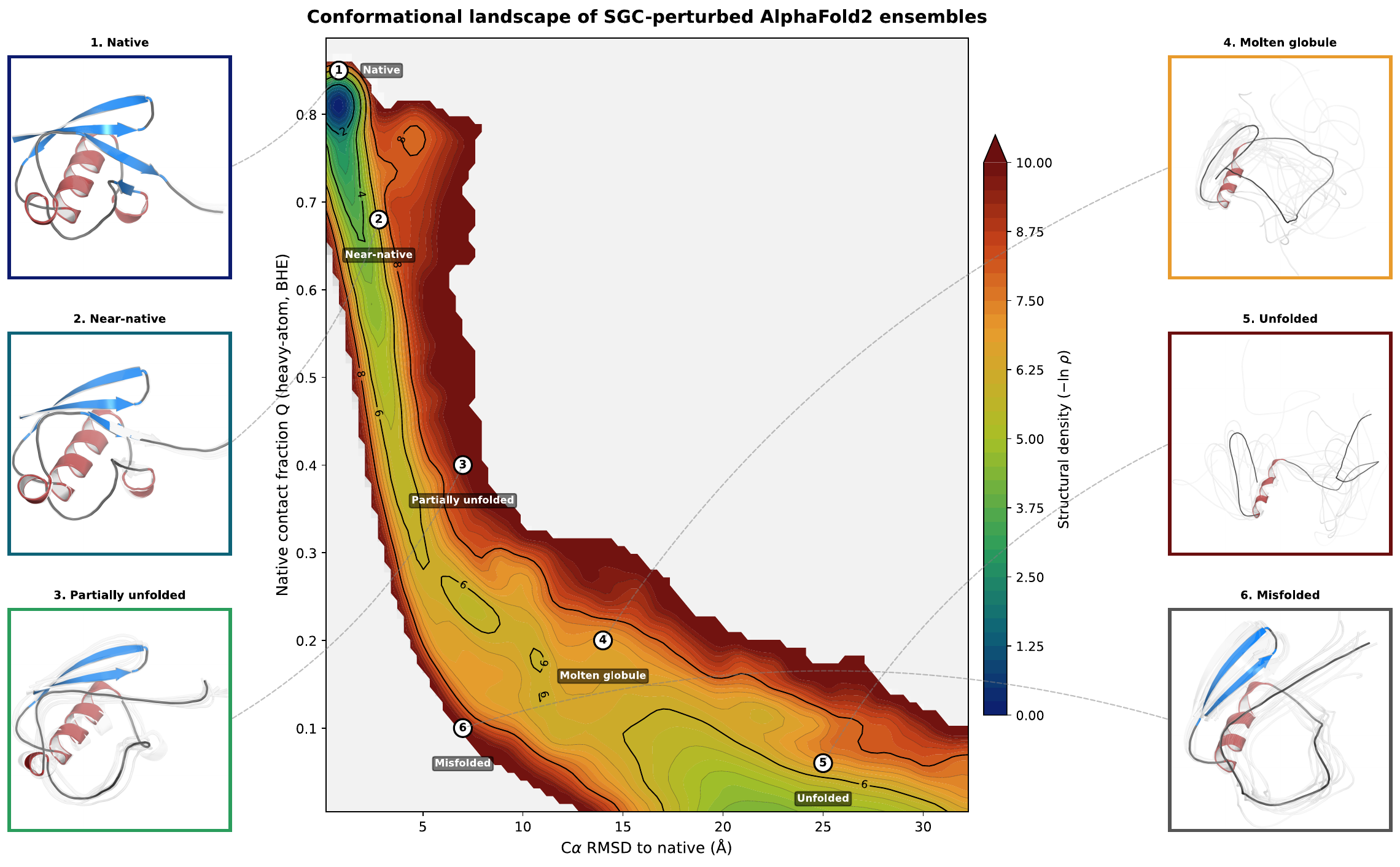}
    \caption{\textbf{Conformational landscape of SGC-perturbed ubiquitin.}
    Structural density ($-\ln\,\rho$) computed from 912{,}384 structures pooled across the primary parameter grid (model~1, MSA $64 \times 64$; 192 ($\sig$, $\lam$) conditions $\times$ 3 seeds $\times$ 48 depths $\times$ 33 recycles), projected onto the C$\alpha$ RMSD $\times$ heavy-atom $Q$-factor (BHE, 4.5~\AA{} cutoff) plane.
    The landscape exhibits an L-shaped funnel topology: a deep native basin (upper left) connected through a hinge near RMSD $\approx 10$~\AA{}, $Q \approx 0.25$ to the denatured ensemble at high RMSD and low $Q$.
    Numbered insets show representative structural ensembles from six regions, rendered as a medoid structure (cartoon) surrounded by ghost-trail members selected for structural diversity within each region's RMSD $\times$ $Q$ bounds.
    \textbf{1.~Native} ($Q > 0.75$, RMSD $< 1.5$~\AA{}): fully folded, all secondary structure intact.
    \textbf{2.~Near-native} ($Q$~0.55--0.80, RMSD~1.5--4~\AA{}): loop rearrangements and terminal fraying, $\beta$-grasp fold preserved.
    \textbf{3.~Partially unfolded} ($Q$~0.25--0.55, RMSD~4--10~\AA{}): $\alpha$1 helix and partial $\beta$-sheet retained.
    \textbf{4.~Molten globule} ($Q$~0.10--0.30, RMSD~10--18~\AA{}): compact but with disordered secondary-structure packing.
    \textbf{5.~Unfolded} ($Q < 0.12$, RMSD $> 18$~\AA{}): extended chain with only a helical fragment surviving.
    \textbf{6.~Misfolded} ($Q < 0.20$, RMSD~3--12~\AA{}): compact with non-native contact topology.
    Dashed lines connect each inset to its approximate landscape position. Bold contour lines at even-integer $-\ln\,\rho$ levels; fine contours every 0.5 units.}
    \label{fig:landscape}
\end{figure}

Even the most deformed structures retain recognizable secondary-structure elements. They are partially unfolded intermediates, not random coils. The perturbation does not destroy the model's capacity for folding; it modulates how far that capacity extends (see Figure~\ref{fig:landscape} for a structural progression across the landscape).

The direction of perturbation also matters. A reverse scheme (perturbing blocks $\{48{-}d, \ldots, 47\}$ instead of $\{0, \ldots, d{-}1\}$) was run at the same condition ($\sig = 0.30$, $\lam = 0.75$). Forward and reverse perturbation are strongly asymmetric. Forward perturbation produces the structural transition described above, with RMSD rising from 1.36~\AA{} at depth~25 through a peak of 23.0~\AA{} at depth~34. Reverse perturbation degrades the structure gradually, with RMSD rising from 0.7 to 3.8~\AA{} across all 48 depths (3-seed mean; the asymmetry is reproduced across all three MSA depths tested; depth-resolved numbers in Supplementary Section~\ref{sec:supp-forward-reverse}). At matched depth, corrupting the early $d$ blocks is much more effective than corrupting the late $d$ blocks at driving the prediction out of the native basin.

\subsection{The perturbation parameter space reveals a continuous transition boundary}
\label{sec:results-boundary}

The depth profile described above was measured at a single ($\sig$, $\lam$) condition. To map how the structural response of ubiquitin varies across the full parameter space, we computed the maximum RMSD and minimum $Q$-factor for each of the 192 ($\sig$, $\lam$) combinations in the parameter grid. Both metrics are averaged over 3 seeds with the block-2 anomaly excluded.

\begin{figure}[!htbp]
    \centering
    \includegraphics[width=\textwidth]{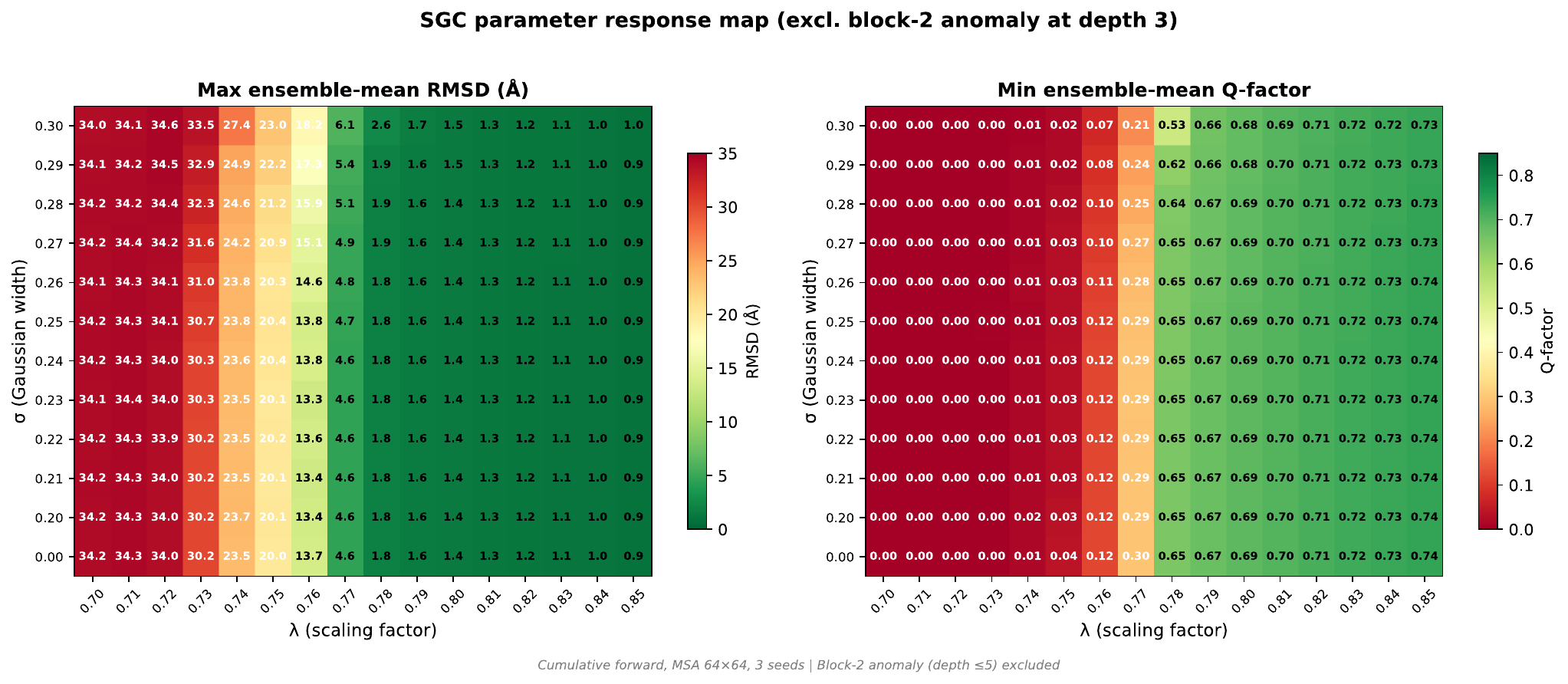}
    \caption{\textbf{Parameter response map of the SGC parameter space.} Left: maximum ensemble-mean RMSD to native (in \AA{}) across all perturbation depths at each ($\sig$, $\lam$) setting. Right: minimum ensemble-mean $Q$-factor (fraction of native contacts preserved). Both panels average over 3 seeds and exclude the block-2 anomaly (depths $\leq 5$). The transition boundary is dominated by $\lam$: a 9\% change in scaling ($\lam = 0.73 \to 0.80$) separates complete unfolding from native-like predictions, while $\sig$ shifts the boundary by at most one $\lam$ step. $Q$-factor resolves the boundary more sharply than RMSD. Data: 12 $\sig$ values $\times$ 16 $\lam$ values $\times$ 48 depths $\times$ 3 seeds $= 27{,}648$ runs (model\_1\_ptm, MSA $64 \times 64$).}
    \label{fig:phase_diagram}
\end{figure}

The result is a smooth transition boundary, not a cliff (Figure~\ref{fig:phase_diagram}). $\lam$ is the dominant axis: at $\sig = 0.30$, maximum RMSD increases from 1.5~\AA{} at $\lam = 0.80$ to 6.1~\AA{} at $\lam = 0.77$ to 33.5~\AA{} at $\lam = 0.73$. The corresponding minimum $Q$-factor drops from 0.68 to 0.21 to $<$0.01. This gradient spans roughly $\Delta\lam = 0.07$---a 9\% change in scaling factor separates near-native predictions from complete unfolding.


$Q$-factor resolves the transition more sharply than RMSD. The $Q$ heatmap shows a tighter band between the folded and unfolded regions, reflecting the sigmoidal relationship between native contact loss and global structural deviation. $Q$ saturates early (native contacts are either maintained or not) while RMSD continues to grow as unfolded structures expand. For identifying the transition boundary, $Q < 0.5$ is a more discriminating threshold than any fixed RMSD cutoff. Although $\lam$ dominates the boundary position, $\sig$ modulates the severity of the response within the transition zone; the next section examines this interaction.

\subsection{Sigma and lambda provide independent control}
\label{sec:results-sigma-lambda}

The parameter response map shows that both $\sig$ and $\lam$ influence the structural response, but their roles are mechanistically distinct.

\begin{figure}[!htbp]
    \centering
    \includegraphics[width=\textwidth]{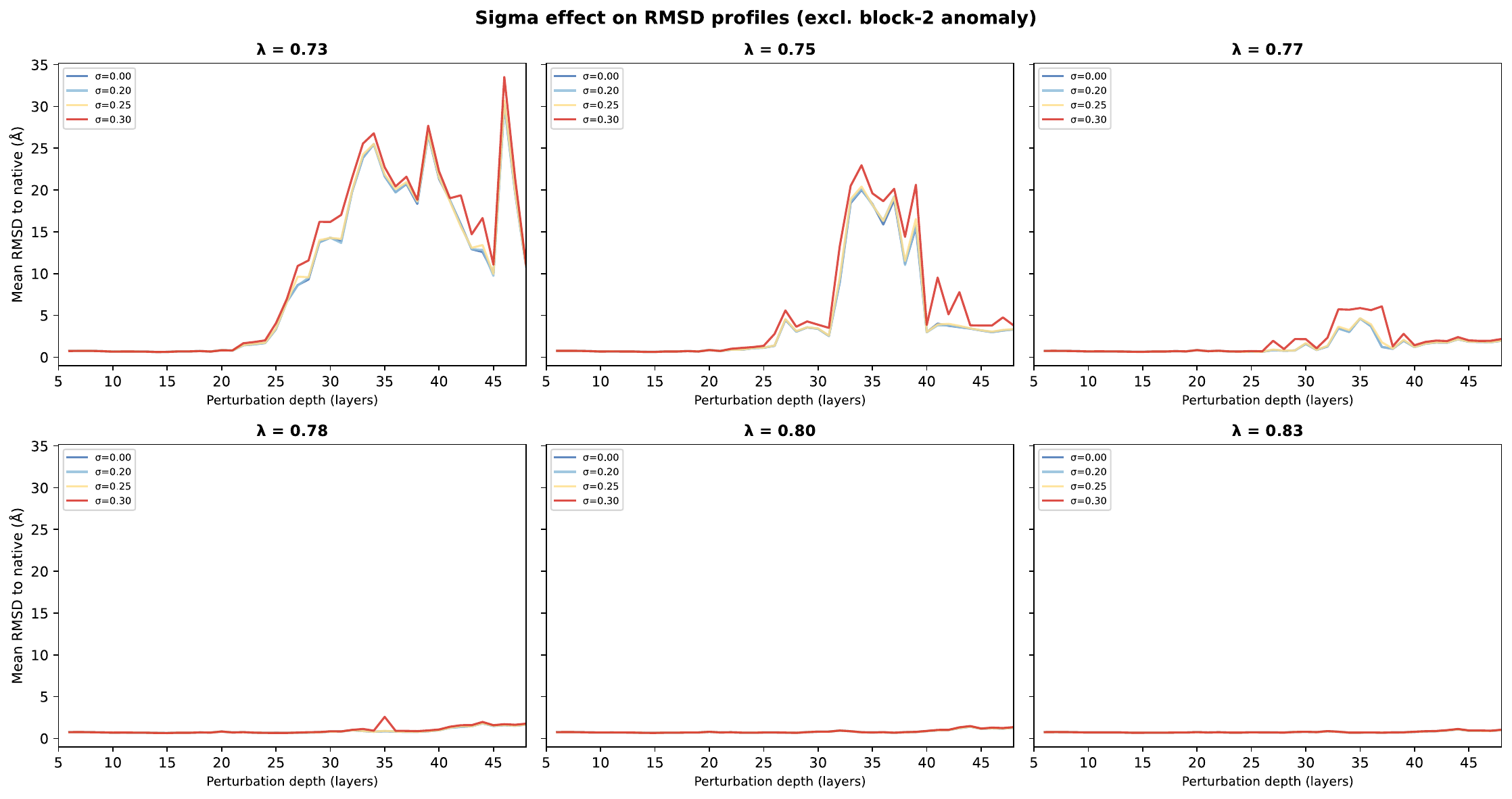}
    \caption{\textbf{Sigma and lambda provide independent control over the structural response.} Mean RMSD to native vs perturbation depth at six representative $\lam$ values (panels), with each curve colored by $\sig$ (0.00 = pure scaling, blue; 0.20, gold; 0.25, orange; 0.30, red). At high $\lam$ ($\geq 0.80$), all $\sig$ values produce identical, near-native profiles. At intermediate $\lam$ (0.75--0.78), $\sig$ amplifies the perturbation: at $\lam = 0.77$, pure scaling reaches 4.6~\AA{} while $\sig = 0.30$ reaches 6.1~\AA{}. At low $\lam$ ($\leq 0.73$), all conditions collapse regardless of $\sig$. Block-2 anomaly (depths $\leq 5$) excluded. Each curve averages over 3 seeds; model\_1\_ptm, MSA $64 \times 64$.}
    \label{fig:controllability}
\end{figure}

$\lam$ sets the transition depth. At $\sig = 0.30$, the depth at which mean RMSD first exceeds 1.5~\AA{} shifts continuously with $\lam$: from depth~22 at $\lam = 0.73$, to depth~26 at $\lam = 0.75$, to depth~27 at $\lam = 0.77$, to depth~35 at $\lam = 0.78$ (Figure~\ref{fig:controllability}). Scaling governs \emph{how much} perturbation the model can absorb before the structure deforms. This tunability is the primary controllability axis for the method: by adjusting $\lam$ by 0.01, one shifts the onset of structural disruption by 2--5 Evoformer blocks.

\begin{figure}[!htbp]
    \centering
    \includegraphics[width=0.85\textwidth]{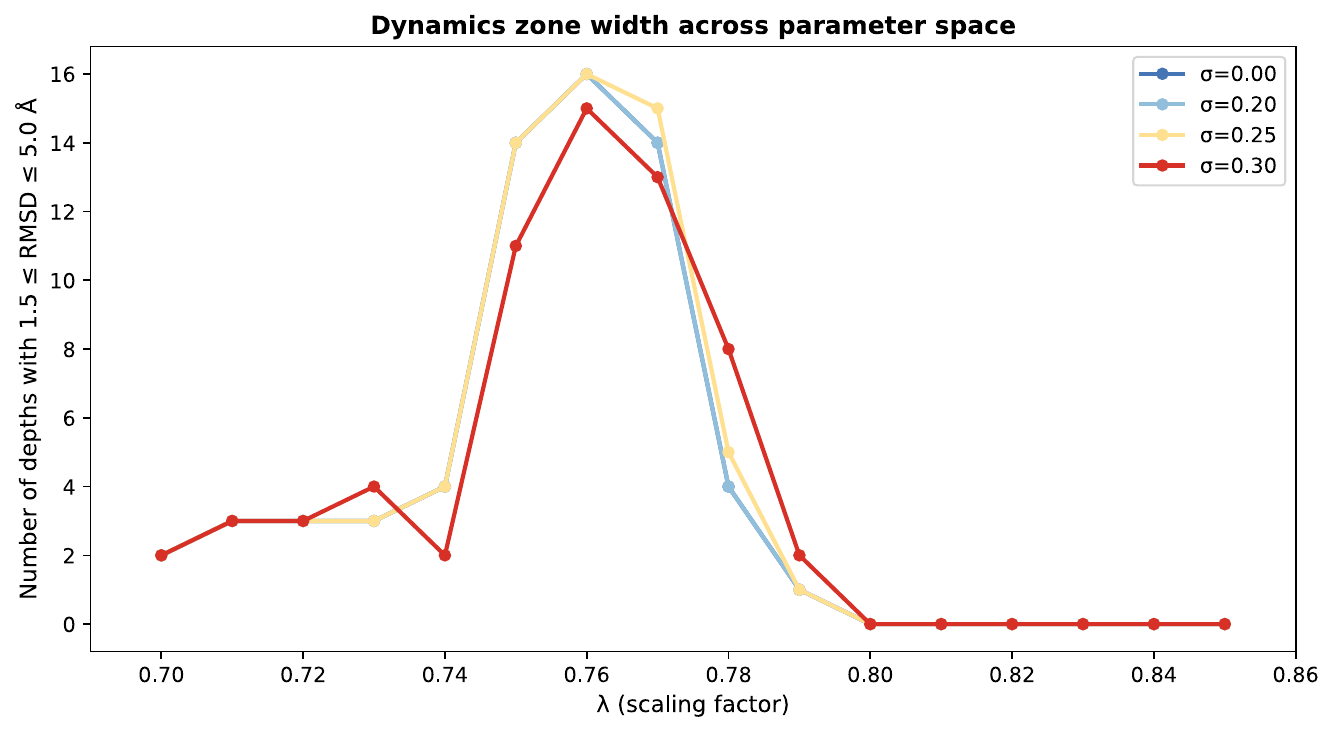}
    \caption{\textbf{The dynamics zone is narrow and tunable.} Number of perturbation depths producing moderate structural deformation (RMSD 1.5--5.0~\AA{}) as a function of $\lam$, colored by $\sig$. The zone peaks at $\lam = 0.76$--$0.77$, where 13--15 depths produce structures in this range. At $\lam \geq 0.80$, no depth exceeds 1.5~\AA{}; at $\lam \leq 0.73$, the model jumps directly from native-like to collapsed without passing through intermediate states. $\sig$ broadens the zone slightly at intermediate $\lam$ but does not shift its location. Block-2 anomaly excluded. Model\_1\_ptm, 3 seeds, MSA $64 \times 64$.}
    \label{fig:dynamics_zone}
\end{figure}

$\sig$ amplifies the perturbation at the transition boundary. At high $\lam$ ($\geq 0.80$), where the model remains native-like regardless of depth, $\sig$ has no detectable effect. Pure scaling ($\sig = 0.00$) and full smoothing ($\sig = 0.30$) produce identical RMSD profiles. At low $\lam$ ($\leq 0.73$), where the model collapses regardless, $\sig$ again makes no difference. But at intermediate $\lam$ (0.75--0.78), where the system sits near the transition boundary, $\sig$ has an outsized effect. At $\lam = 0.77$, adding Gaussian smoothing ($\sig = 0.30$) raises the peak recycle-averaged RMSD from 4.6 to 6.1~\AA{} and extends structural disruption across 18 depths compared to 14 for pure scaling. This broadening comes from an operation that contributes $<$1\% of the total perturbation power (Section~\ref{sec:methods-sgc}). At $\lam = 0.78$, where the model is otherwise nearly native-like, smoothing doubles the dynamics zone from 4 to 8 qualifying depths.

This asymmetry is visible in the ``dynamics zone'' (Figure~\ref{fig:dynamics_zone}), the number of perturbation depths producing moderate structural deformation (RMSD 1.5--5.0~\AA{}) at each ($\sig$, $\lam$) setting. At $\lam \leq 0.73$, most profiles pass directly from native-like to collapsed without extended intermediate sampling.

Although Gaussian smoothing is formally a low-pass filter, at $\sig = 0.30$ the frequency redistribution is negligible. The spectral shape shifts by fewer than five percentage points across all frequency bands (Supplementary Section~\ref{sec:supp-power-decomp}). The structural effect may instead reflect the change in direction of the smoothing residual in weight space. Where uniform scaling preserves each weight matrix's internal structure at lower amplitude, smoothing introduces a geometrically orthogonal component. It alters the geometry of the computation (Supplementary Section~\ref{sec:supp-directional}) rather than merely its amplitude. This directional component matters most near the transition boundary, where the model's folding decision depends on fine-grained weight structure that encodes the precise inter-residue contacts needed to lock the fold. Per-pair analysis of the Evoformer's pair representation suggests that smoothing sensitivity concentrates in a minority of residue pairs ($\sim$200--300 out of 2{,}850), with hotspots mapping to contacts at the boundaries between stable and fragile structural groups (the folding nucleus and the native-versus-intermediate boundary), while leaving the remaining pairs indistinguishable from pure scaling (Supplementary Figure~S17). The two operations are therefore complementary: scaling governs whether the model can fold; smoothing changes which contacts are most vulnerable under the perturbation (Section~\ref{sec:results-unfolding}).


\subsection{The conformational landscape mirrors known folding topology}
\label{sec:results-topology}

The individual depth profiles and parameter sweeps described above characterize the structural response one condition at a time. Pooling all 912{,}384 structures from the primary parameter grid (model~1, MSA $64 \times 64$) onto a single RMSD $\times$ $Q$ plane reveals the full conformational landscape that \sgc{} makes accessible (Figure~\ref{fig:landscape}).

The landscape has a characteristic L-shaped topology. The native basin occupies the upper-left corner ($Q \approx 0.81$, RMSD $< 1.5$~\AA{}). This is a deep, narrow well that contains the majority of structures from the enhancement zone. From this basin, the landscape extends along two arms. One is a vertical descent in $Q$ at moderate RMSD (2--10~\AA{}, corresponding to partial unfolding with progressive contact loss). The other is a horizontal extension to high RMSD at low $Q$ ($> 15$~\AA{}, corresponding to fully denatured conformations). The two arms meet at a hinge around RMSD $\approx 10$~\AA{}, $Q \approx 0.25$, where structures are best described as molten globules---compact but lacking native contacts.

Representative structures from six regions of this landscape illustrate the structural progression (Figure~\ref{fig:landscape}, insets). DSSP analysis quantifies what survives at each level of deformation. In the native basin, all secondary-structure elements are fully conserved. Near-native structures ($Q \approx 0.68$) show loop rearrangements and terminal fraying but retain the complete $\beta$-grasp fold. Partially unfolded structures ($Q \approx 0.39$) preserve the $\alpha$1 helix and portions of the $\beta$-sheet, while more peripheral contacts have been lost. Molten globules ($Q \approx 0.19$) retain recognizable helices and short $\beta$-strands, but their spatial arrangement is disordered. In the unfolded region ($Q \approx 0.05$), a fragment of the $\alpha$1 helix is the only surviving secondary-structure element, visible as the lone helical segment anchoring otherwise extended chains. The structural elements do not dissolve uniformly; some persist deep into the denatured region while others break early. Whether this hierarchy reflects the known stability architecture of ubiquitin is the question we turn to next.

\begin{figure}[!htbp]
    \centering
    \includegraphics[width=\textwidth]{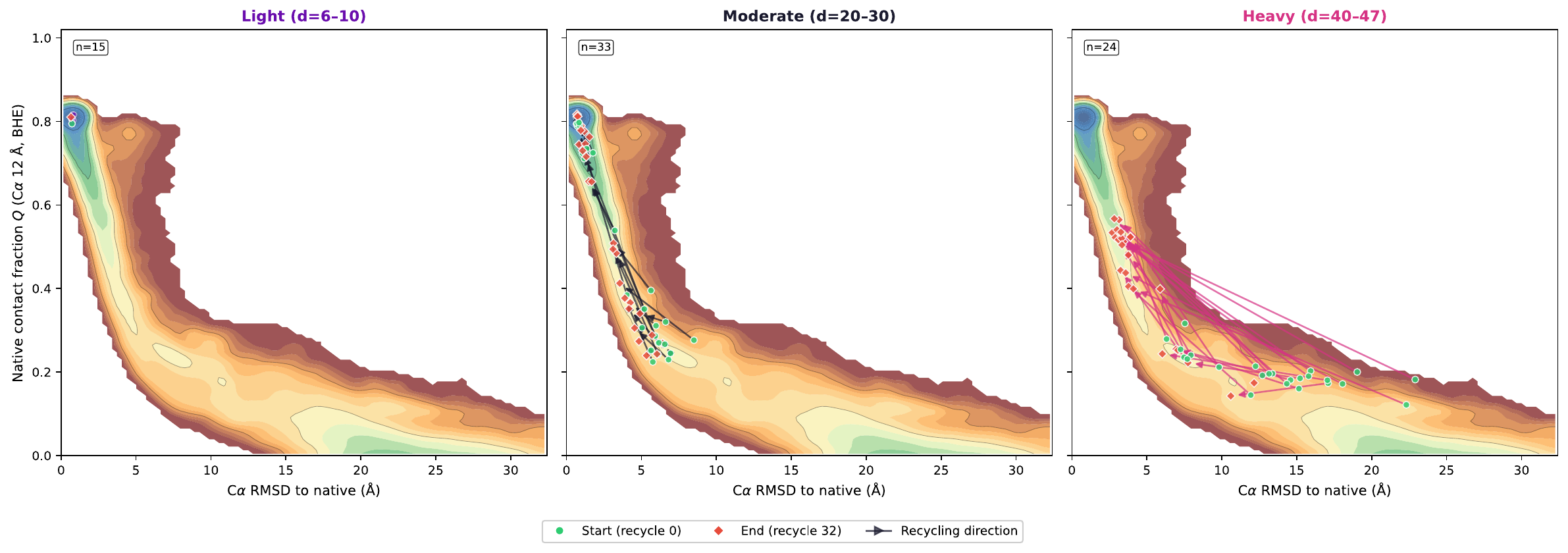}
    \caption{\textbf{Recycling trajectories on the conformational landscape.}
    Start (recycle~0, green circles) and end (recycle~32, colored diamonds) positions for each run at $\sig = 0.30$, $\lam = 0.75$, shown against the full conformational landscape (all conditions, model~1). Arrows indicate recycling direction.
    \textbf{Light} perturbation (depths 6--10): trajectories remain confined to the native basin with minimal displacement.
    \textbf{Moderate} perturbation (depths 20--30): trajectories fan along the L-shaped landscape, with early recycles partially unfolded and later recycles refolding toward native.
    \textbf{Heavy} perturbation (depths 40--47): trajectories reach the unfolded and molten-globule regions before partial recovery. All structural metrics from raw (unminimized) predictions after Kabsch superposition.}
    \label{fig:trajectories}
\end{figure}

Perturbation depth determines where on this landscape a given run's recycling trajectory lands (Figure~\ref{fig:trajectories}). At shallow depth (6--10 blocks), trajectories remain confined to the native basin, tracing small excursions that return to the starting point. At moderate depth (20--30 blocks), trajectories sweep through the transition zone, with early recycles producing partially unfolded structures that progressively refold across subsequent iterations. At deep perturbation (40--47 blocks), trajectories explore the full extent of the landscape, reaching the molten-globule and unfolded regions before partial recovery in later recycles.

\begin{figure}[!htbp]
    \centering
    \includegraphics[width=\textwidth]{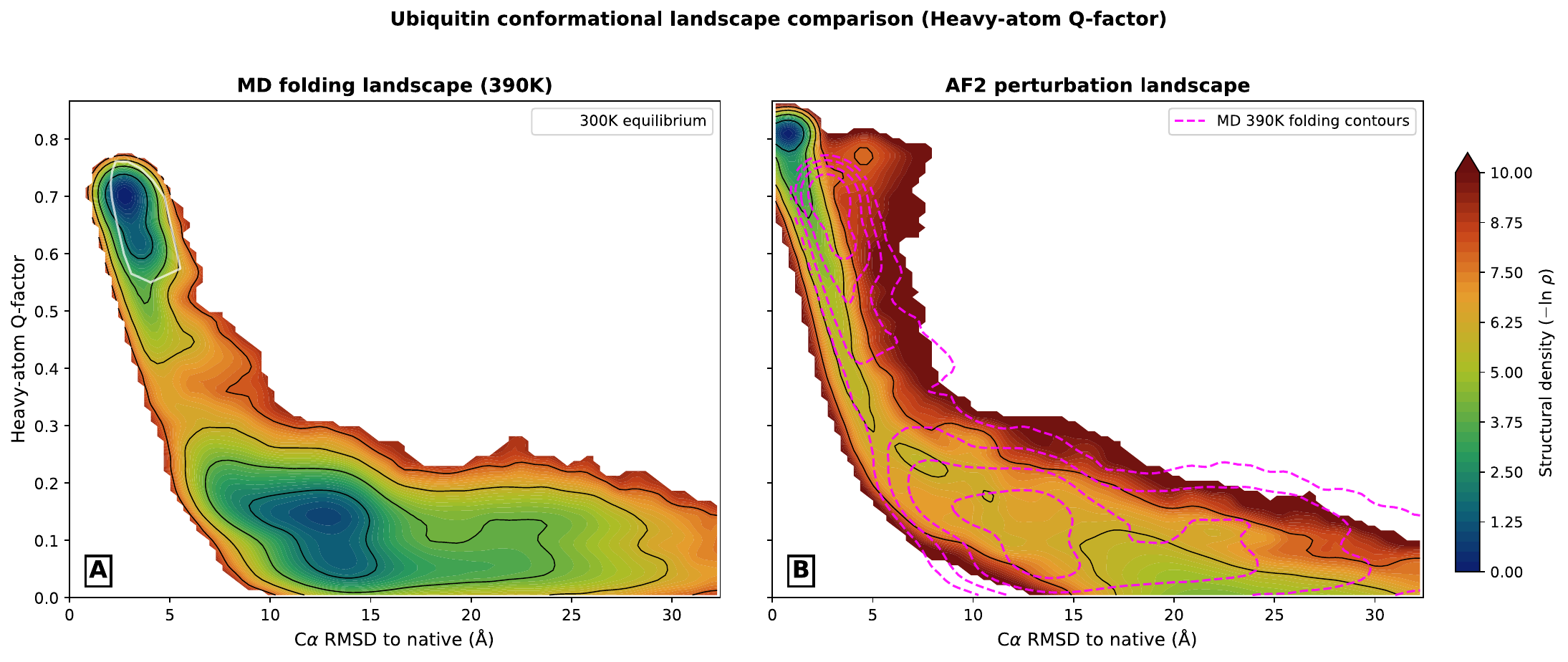}
    \caption{\textbf{The conformational landscape mirrors the MD folding funnel.} \textbf{(A)}~Structural density landscape from six independent millisecond-scale MD simulations of ubiquitin at 390~K (673{,}519 frames; Piana et al.~\cite{piana2013}), plotted as $-\ln\rho$ on C$\alpha$ RMSD $\times$ heavy-atom Q-factor axes. White outline: convex hull of 300~K equilibrium MD (82{,}280 frames, CHARMM22*/TIP3P + CHARMM36m~\cite{robustelli2018}). \textbf{(B)}~AF2 perturbation landscape (912{,}384 frames, model~1, all $\sig$/$\lam$ conditions). Dashed magenta contours: MD 390~K folding contours at $-\ln\rho = 2, 4, 6, 8$, overlaid for comparison. White outline: 300~K equilibrium hull. Both landscapes share the same L-shaped funnel topology. The 300~K equilibrium region is fully contained within the AF2 native basin. Q-factor computed using heavy-atom contacts (4.5~\AA{} cutoff, BHE, 884 contacts).\protect\footnotemark}
    \label{fig:md_landscape}
\end{figure}

The topology of this landscape is not unique to \sgc{}. We compared the AF2 perturbation landscape against the folding landscape obtained from millisecond-scale molecular dynamics simulations of ubiquitin at 390~K~\cite{piana2013}. These six independent trajectories start from the native state and sample reversible unfolding (Figure~\ref{fig:md_landscape}). Both landscapes share the same L-shaped funnel topology. A deep native basin connects by a continuous path to the denatured ensemble, with no isolated basins or disconnected regions in this RMSD--$Q$ projection. The MD folding contours, when overlaid on the AF2 landscape, trace the same arms and the same hinge region. Equilibrium MD at 300~K (154.6~$\mu$s, CHARMM22*/TIP3P~\cite{robustelli2018}) samples a compact region within the native basin that is fully contained within the AF2 landscape.

\footnotetext{The AF2 baseline Q-factor of 0.806 (vs.\ 1.0 for MD at the native state) reflects imprecise sidechain placement rather than missing backbone contacts: a C$\alpha$-only Q-factor gives 0.961 for the AF2 baseline. Topological agreement is broadly consistent across contact definitions (65.0\% kNN territory coverage under C$\alpha$-only contacts; Supplementary Fig.~S5).}

To quantify this agreement, we measured territorial overlap using the $k$-nearest-neighbor support estimator described in Section~\ref{sec:methods-structural}. After affine normalization of both axes, each dataset's occupied territory was defined by its 99\% self-support level set ($k = 20$). MD coverage by AF2 was then measured as the area-weighted fraction of MD support also covered by AF2 support. This yielded 72.6\% territorial coverage, stable to within 1 percentage point across $k = 5$--$80$ (Supplementary Fig.~S26). Nearly three-quarters of the conformational space visited by the 390~K MD trajectories is also accessible through \sgc{} perturbation.

The agreement is topological, not distributional. There is no reason to expect the density distributions to match. The two landscapes are generated by fundamentally different processes. What they share is the \emph{shape} of the accessible conformational space. The same regions are populated, the same regions are empty, and the path between the native and denatured states follows the same funnel geometry~\cite{bryngelson1995}. The territorial overlap is 72.6\%, and it is specific: matched-power noise controls do not reproduce it. The learned computation, when its rules are continuously deformed, generates structures that populate the same regions of the projected conformational landscape as millisecond-scale physical simulation. The next two sections test whether the ordering within this landscape also has physical correspondence.

\subsection{Perturbation depth traces a physically ordered unfolding pathway}
\label{sec:results-unfolding}

The conformational exploration induced by \sgc{} is not random. As perturbation depth increases, native contacts do not break uniformly. They break in a sequence that reflects the experimentally established stability ordering of the ubiquitin native state.

We classified native C$\alpha$ contacts (distance $< 12$~\AA, sequence separation $\geq 3$) into three groups drawn from the ubiquitin folding literature~\cite{went2005, sosnick2004, piana2013}. G1 is the folding nucleus: the $\beta$1--$\beta$2 hairpin docked against the $\alpha$1 helix (93 contacts). G2 is the intermediate boundary: $\alpha$1 and $\beta$2 contacts to the 3$_{10}$ helix and $\beta$3--$\beta$5 loop (45 contacts). G3 comprises the $\beta$5 long-range contacts between $\beta$1/$\beta$2 and $\beta$3 and the C-terminal $\beta$5 strand (112 contacts). For each of the 27,648 runs across the 192 ($\sig$, $\lam$) primary grid conditions, three seeds, and 48 perturbation depths, we computed the fraction of native contacts maintained in each group---a group-resolved Q-factor.

\begin{figure}[!htbp]
    \centering
    \includegraphics[width=\textwidth]{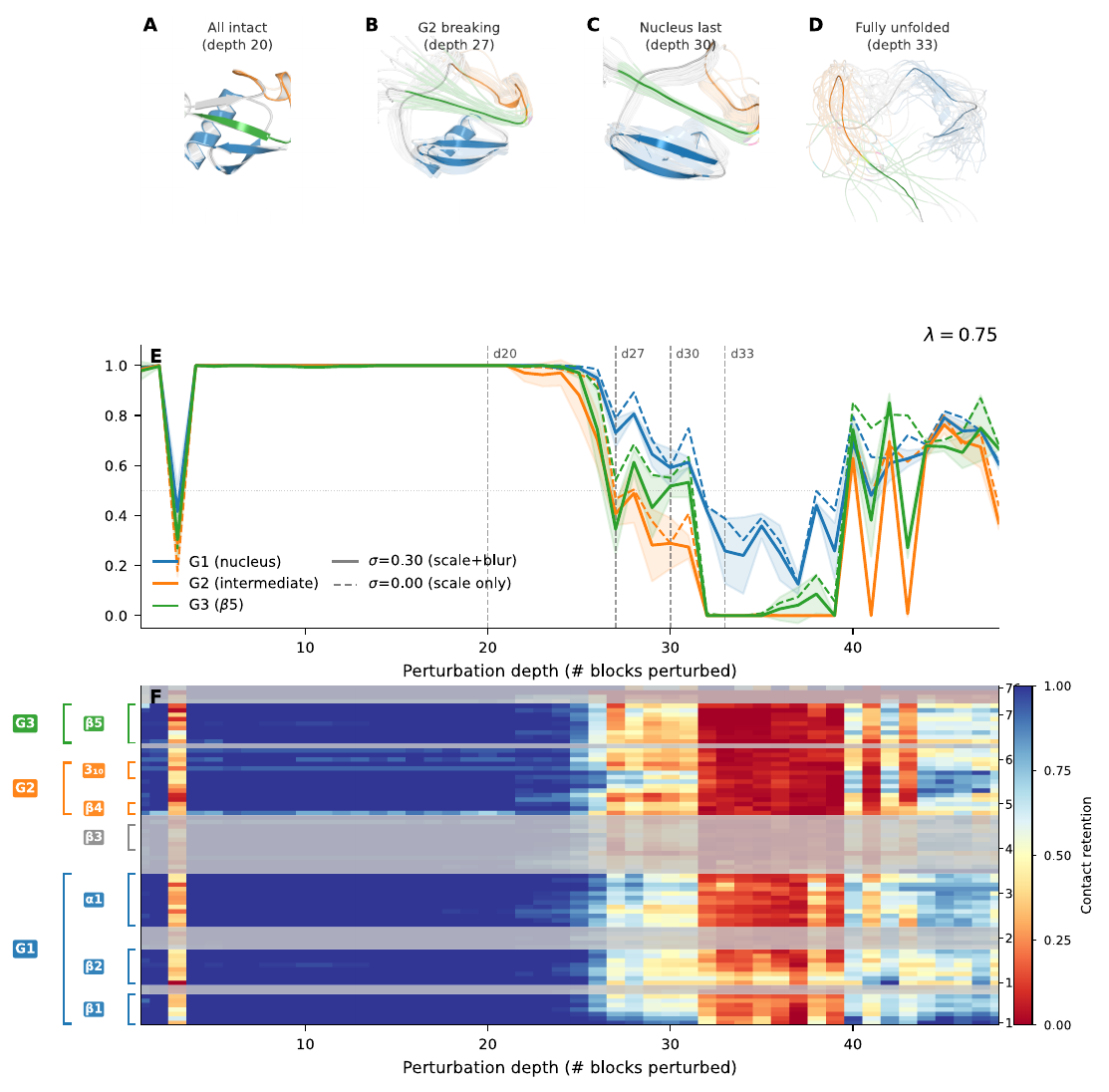}
    \caption{\textbf{Perturbation depth traces a physically ordered unfolding pathway.}
    \textbf{(A--D)}~Structural ensembles of ubiquitin at four perturbation depths ($\sig = 0.30$, $\lam = 0.75$), each showing the final-recycle structure from seed~43 (solid cartoon) overlaid on a ghost ensemble of three seeds $\times$ nine recycle snapshots (transparency = 0.70). Residues are coloured by contact group: G1/nucleus (blue; $\beta$1--$\beta$2 hairpin + $\alpha$1 helix), G2/intermediate (orange; $\beta$4 + 3$_{10}$ helix), G3/$\beta$5 (green; C-terminal strand), background (grey).
    \textbf{(A)}~At depth~20, all three groups retain native contacts and the fold is intact.
    \textbf{(B)}~At depth~27, G2 contacts are disrupted ($Q = 0.41$) while G1 remains largely intact ($Q = 0.73$).
    \textbf{(C)}~At depth~30, G2 and G3 are extensively disordered while the G1 nucleus persists ($Q = 0.59$).
    \textbf{(D)}~By depth~33, all groups have lost their native contacts.
    \textbf{(E)}~Fraction of native C$\alpha$--C$\alpha$ contacts ($Q$, cutoff 12~\AA, minimum sequence separation~3) retained by each group as a function of perturbation depth. Solid lines: $\sig = 0.30$ (scale + blur); dashed lines: $\sig = 0.00$ (scale only). Shaded bands show $\pm 1$ s.d.\ across three seeds. Vertical dashed lines mark the depths shown in panels A--D. The ordering G2-first, G3-second, G1-last is robust across all conditions; blur ($\sig = 0.30$) sharpens the separation between groups, particularly resolving G2 and G3 loss that scale alone conflates.
    \textbf{(F)}~Per-residue contact retention averaged across seeds ($\sig = 0.30$, $\lam = 0.75$). Each row is one residue (1--76); columns are perturbation depth. Contact-group residues are shown at full colour saturation; background residues (turns, linkers) are desaturated. Secondary structure elements and contact group membership are annotated at left. The nucleus (G1, blue rows) retains contacts at depths where G2 and G3 have already transitioned to red, consistent with the group-level ordering in panel~E.}
    \label{fig:unfolding}
\end{figure}

The intermediate boundary contacts break first. Of the 192 ($\sig$, $\lam$) conditions, 84 produce perturbation strong enough for all three groups to cross the 50\% $Q$ threshold. The remaining conditions either retain all contacts above 50\% at every depth or collapse all three groups simultaneously. Across those 84 informative conditions, G2 drops first in every case. The folding nucleus never breaks first. G1 survives to a greater or equal depth than every other group in all 84 conditions, and strictly outlasts the others in 67 (80\%). The remaining 17 are ties between G1 and G3, concentrated at $\lam = 0.75$--$0.76$ (12 and 5 conditions respectively), where the perturbation sits at the edge of the transition boundary and both groups lose their contacts at the same depth. Per-residue analysis reveals a spatial wave of contact loss that originates in the 3$_{10}$ helix and loop region and propagates inward toward the $\beta$1--$\beta$2 core (Figure~\ref{fig:unfolding}F).

This ordering is robust. It holds across three seeds, across $\sig$ from 0.00 to 0.30, and across $\lam$ from 0.70 to 0.77 (Figure~\ref{fig:unfolding}E; Supplementary Figs.~S14--S16). The transition depth shifts with $\lam$ (from depth~25 at $\lam = 0.73$ to depth~33 at $\lam = 0.77$), but the relative fragility of the three contact groups does not.

The two perturbation parameters play distinct roles in this hierarchy. At lower $\lam$ (0.70--0.74), where the perturbation is strong enough to break all three groups, scaling alone already separates the nucleus from $\beta$5 contacts: G1 persists 3--4 blocks deeper than G3. At the transition boundary ($\lam = 0.75$--$0.76$), this separation vanishes and the two groups lose their contacts at the same depth. Gaussian smoothing introduces a selective vulnerability at these boundary conditions. At $\lam = 0.75$ with $\sig = 0.30$, G3's group-resolved $Q$ dips to 0.35 at depth~27 while G1 remains at 0.73 (Figure~\ref{fig:unfolding}E, solid vs.\ dashed lines). Pure scaling at the same $\lam$ produces no such dip. G3 recovers before both groups permanently lose their contacts at depth~32, so the effect is a transient deepening of $\beta$5 contact loss rather than an earlier collapse. Per-pair analysis of the Evoformer pair tensor suggests the same pattern at the representational level. The residue pairs most sensitive to blur (up to 21\% fractional change vs.\ 3--5\% background) map to the folding nucleus and the native-vs-intermediate boundary contacts (Supplementary Figure~S17).

Several caveats apply. The unfolding ``pathway'' is not a trajectory. Each perturbation depth is a separate inference run, not a time point. The ordering reflects how robustly different contacts are encoded in the Evoformer weights, not a physical unfolding mechanism. The G2-before-G3 ordering matches the experimentally established stability ordering of the native state, not the kinetic folding pathway. The 3$_{10}$/loop region has the highest crystallographic B-factors and lowest NMR order parameters, whereas $\beta$5 contacts are kinetically more peripheral~\cite{went2005}. This is consistent with AlphaFold2 being trained on PDB structures and self-distilled predicted structure targets, rather than on kinetic or thermodynamic ensembles. Finally, this analysis covers one protein; generalization remains to be established.

The perturbation produces ordered, not random, structural change. It destabilizes the least robustly encoded contact groups first and preserves the most robustly encoded group last. We consider the implications of this ordering in the Discussion.

\subsection{Perturbation sensitivity tracks physical flexibility}
\label{sec:results-rmsf}

If the weights encode the conformational landscape, the least constrained residues should align with physically flexible regions. To test this, we compared per-residue perturbation sensitivity from \sgc{} ensembles against equilibrium MD RMSF from a 154.6-microsecond, 300~K simulation of ubiquitin (CHARMM22*/TIP3P; Robustelli et al.~\cite{robustelli2018}). All comparisons use MSA depth $64 \times 64$, following the design logic of Section~\ref{sec:methods-impl}: probing what the weights encode requires a regime where weight perturbation has a measurable effect despite the remaining coevolutionary input. For each perturbation depth, we computed perturbation sensitivity across the 33-structure recycling ensemble at $\sig = 0.30$, $\lam = 0.75$ and correlated the resulting 76-residue profile against the MD reference. Three MSA-subsampling seeds provide error estimates; the same depth-dependent pattern is reproduced across four additional native-state force fields (Supplementary Fig.~S4).

\textbf{The trivial correlation problem.} A naive Pearson correlation between perturbation sensitivity and MD RMSF risks capturing only the obvious: loops are flexible, the core is rigid. To test whether \sgc{} captures more than protein architecture, we report the partial Pearson~$r$ after controlling for weighted contact number (WCN; Section~\ref{sec:methods-structural}), a geometric measure of residue burial that anti-correlates with flexibility ($r = -0.69$ with MD RMSF).

\textbf{Shallow perturbation recovers the flexibility pattern (depths 4--24).} Perturbation sensitivity strongly tracks MD flexibility (Pearson $r = 0.83$--$0.89$), and this correlation is non-trivial: partial $r$\textbar WCN remains $0.66$--$0.79$ (Table~\ref{tab:rmsf}). At depths 8--20, \sgc{} explains 29--32\% of MD RMSF variance beyond WCN alone ($\Delta R^2_\text{adj}$), exceeding a Gaussian Network Model geometry-only baseline ($\Delta R^2_\text{adj} = 0.19$; Supplementary Fig.~S2). Within loop residues alone ($n = 27$), Pearson $r$ reaches 0.95 (Supplementary Fig.~S3), confirming that \sgc{} resolves which loops are more flexible than others, not merely that loops move more than the core. However, absolute magnitudes disagree: Lin's concordance correlation coefficient is near zero (CCC $\approx 0.08$), reflecting that mean perturbation sensitivity at these depths (${\sim}0.07$~\AA) is roughly 18 times smaller than mean MD RMSF (${\sim}1.3$~\AA). The \sgc{} response is correlated with \emph{which} residues are flexible, but perturbation sensitivity at shallow depths is not a calibrated measure of \emph{how much} they move.

\begin{figure}[!htbp]
    \centering
    \includegraphics[width=\textwidth]{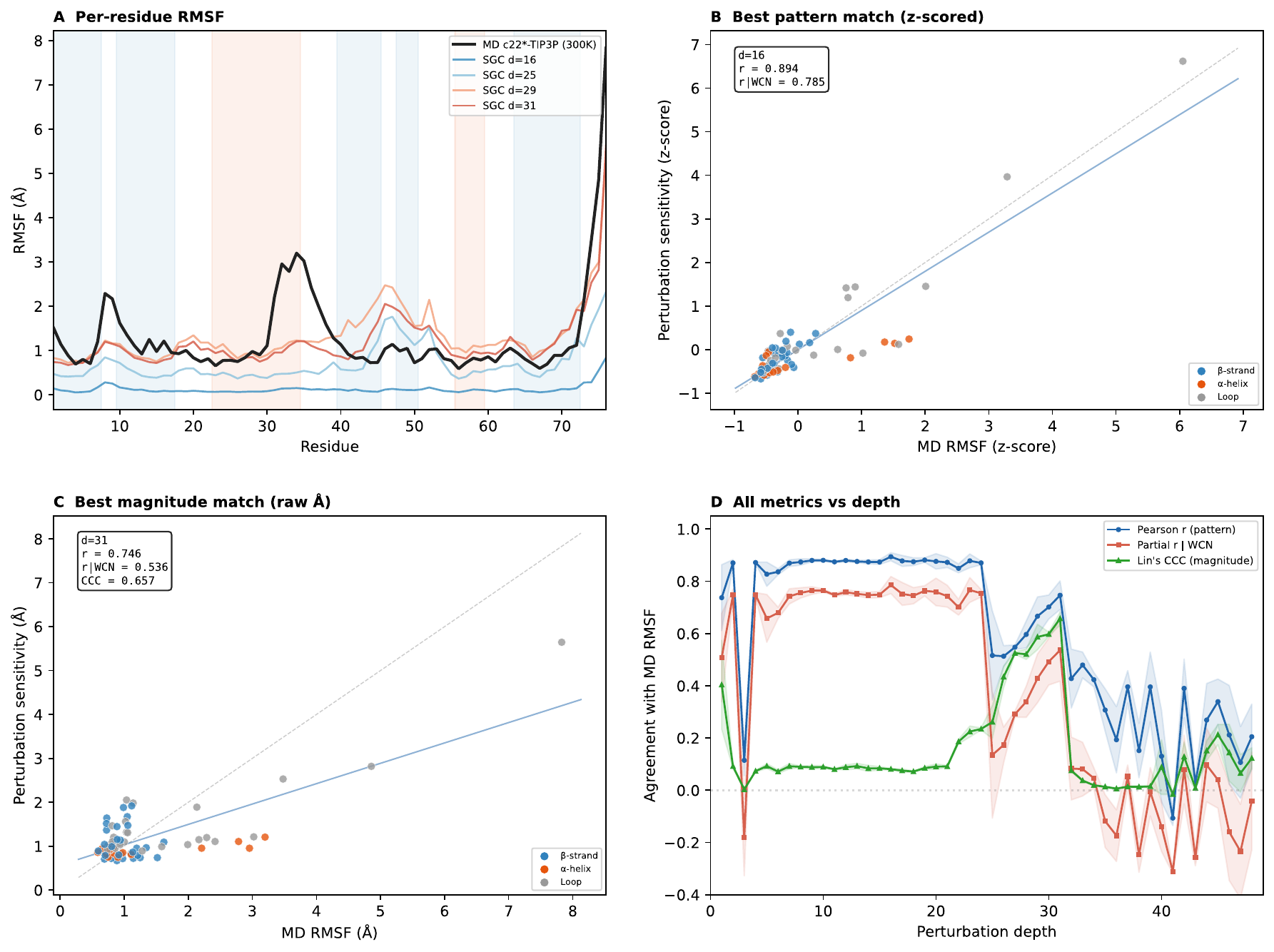}
    \caption{\textbf{Perturbation sensitivity tracks physical flexibility.}
    \textbf{(A)}~Per-residue profiles: MD RMSF from 154.6~$\mu$s CHARMM22*/TIP3P equilibrium (black) and \sgc{} perturbation sensitivity at four representative depths (colored). Secondary structure elements shaded.
    \textbf{(B)}~Z-scored scatter at the depth of best non-trivial agreement (selected by partial $r$\textbar WCN), colored by secondary structure. Residues are colored by structural element ($\beta$-strand, $\alpha$-helix, loop).
    \textbf{(C)}~Raw-scale scatter at the depth of best magnitude agreement (selected by CCC among depths with $r \geq 0.5$). Identity line shown.
    \textbf{(D)}~Agreement metrics vs perturbation depth: Pearson~$r$ (pattern), partial~$r$\textbar WCN (non-trivial pattern), and Lin's CCC (magnitude). Ribbons show $\pm 1$~s.d.\ across 3 seeds.
    All panels use $\sig = 0.30$, $\lam = 0.75$, MSA $64 \times 64$.}
    \label{fig:rmsf}
\end{figure}

\textbf{Deeper perturbation converges in magnitude (depths 27--31).} At intermediate depths, perturbation sensitivity magnitudes converge to the MD RMSF range: CCC peaks at 0.66 (depth~31) with Pearson $r = 0.75$, partial $r$\textbar WCN $= 0.54$, and RMSE $= 0.75$~\AA{} (Figure~\ref{fig:rmsf}, panel~C). However, these structures are moderately unfolded (RMSD 3--5~\AA{} from native) and lie outside the native basin that the 300~K equilibrium MD samples. The magnitude agreement may reflect scale convergence (both quantities happen to fall in the 1--3~\AA{} range) rather than sampling of the same conformational ensemble.

\begin{table}[htbp]
\centering\small
\caption{\textbf{RMSF validation metrics at sampled perturbation depths.}
Pearson~$r$ and Spearman~$\rho$ measure pattern agreement between per-residue \sgc{} perturbation sensitivity and MD RMSF. Partial~$r$\textbar WCN controls for weighted contact number (a geometric proxy for residue burial). $\Delta R^2_\text{adj}$ is the variance explained by \sgc{} beyond WCN in a nested OLS model. CCC is Lin's concordance correlation (magnitude agreement). RMSE is in \AA. All values are means over 3 seeds at $\sig = 0.30$, $\lam = 0.75$.}
\label{tab:rmsf}
\begin{tabular}{rcccccc}
\toprule
Depth & $r$ & $\rho$ & $r$\textbar WCN & $\Delta R^2_\text{adj}$ & CCC & RMSE (\AA) \\
\midrule
 1  & 0.74 & 0.48 &  0.51 & 0.13 & 0.40 &  1.20 \\
 3  & 0.11 & 0.19 & $-$0.18 & 0.02 & 0.00 & 14.68 \\
 5  & 0.83 & 0.74 &  0.66 & 0.22 & 0.09 &  1.53 \\
10  & 0.88 & 0.74 &  0.77 & 0.31 & 0.09 &  1.55 \\
16  & 0.89 & 0.76 &  0.79 & 0.32 & 0.08 &  1.55 \\
20  & 0.88 & 0.77 &  0.76 & 0.30 & 0.09 &  1.54 \\
25  & 0.52 & 0.29 &  0.14 & 0.02 & 0.26 &  1.13 \\
29  & 0.67 & 0.24 &  0.43 & 0.09 & 0.59 &  0.82 \\
31  & 0.75 & 0.27 &  0.54 & 0.15 & 0.66 &  0.75 \\
35  & 0.31 & 0.06 & $-$0.12 & 0.00 & 0.01 & 12.20 \\
40  & 0.13 & $-$0.15 & $-$0.14 & 0.01 & 0.09 &  1.93 \\
48  & 0.21 & $-$0.21 & $-$0.04 & 0.01 & 0.12 &  1.66 \\
\bottomrule
\end{tabular}
\end{table}

\textbf{Beyond the transition boundary, the comparison breaks down.} At depth~25, pattern metrics drop sharply (Pearson $r$ to 0.52, partial $r$\textbar WCN to 0.14) while CCC \emph{improves} from 0.09 to 0.26. This is a pattern-to-magnitude tradeoff rather than a universal collapse. This reflects the structures leaving the native basin. The 300~K equilibrium MD samples near-native states, and once \sgc{} pushes the model into partially unfolded conformations, the two ensembles no longer occupy the same region of conformational space. Beyond depth~32, partial $r$\textbar WCN oscillates near zero. The block-2 anomaly (depth~3) produces a characteristic dip ($r = 0.11$), consistent with the catastrophic sensitivity of the earliest Evoformer blocks described in Section~\ref{sec:results-transition}.

Perturbation sensitivity is a correlate of physical flexibility, not a substitute. Across depths 4--24, the non-trivial pattern correlation persists after controlling for native-state geometry and exceeds the GNM baseline, suggesting that the Evoformer weights carry a residue-level constraint signal correlated with physical flexibility and not reducible to burial alone. We note that the 33-structure recycling ensemble is a deterministic convergence trajectory, not an independent thermodynamic sample; the per-residue spread captures which residues the model's learned representation fails to constrain, and Section~\ref{sec:discussion-limits} discusses what this implies for the comparison.



\subsection{Noise controls show that matched-power incoherent perturbations do not reproduce the \sgc{} response}
\label{sec:results-noise}

The preceding sections demonstrate that \sgc{} produces structured, physically interpretable conformational responses. But a skeptic can ask: is the structure-specific response a property of Gaussian smoothing, or would any perturbation of comparable magnitude do the same? To answer this, we compare \sgc{} against two matched noise controls (Section~\ref{sec:methods-noise}) that deliver identical perturbation power but destroy different aspects of the perturbation's relationship to the original weights.

The three conditions form a clean experimental decomposition. \sgc{} preserves both the spectral profile and the phase relationship between perturbation and weights (the perturbation $\Delta W \approx -0.25W$ is coherent with the weight tensor). Spectral noise preserves the spectral profile but randomizes the phases. White noise destroys both. If the spectral shape matters, spectral noise should reproduce \sgc{}'s behavior. If the total perturbation energy matters, white noise should too. If neither does, the effect depends on the deterministic, weight-aligned structure of the perturbation, which these controls lack.

\begin{figure}[!htbp]
    \centering
    \includegraphics[width=\textwidth]{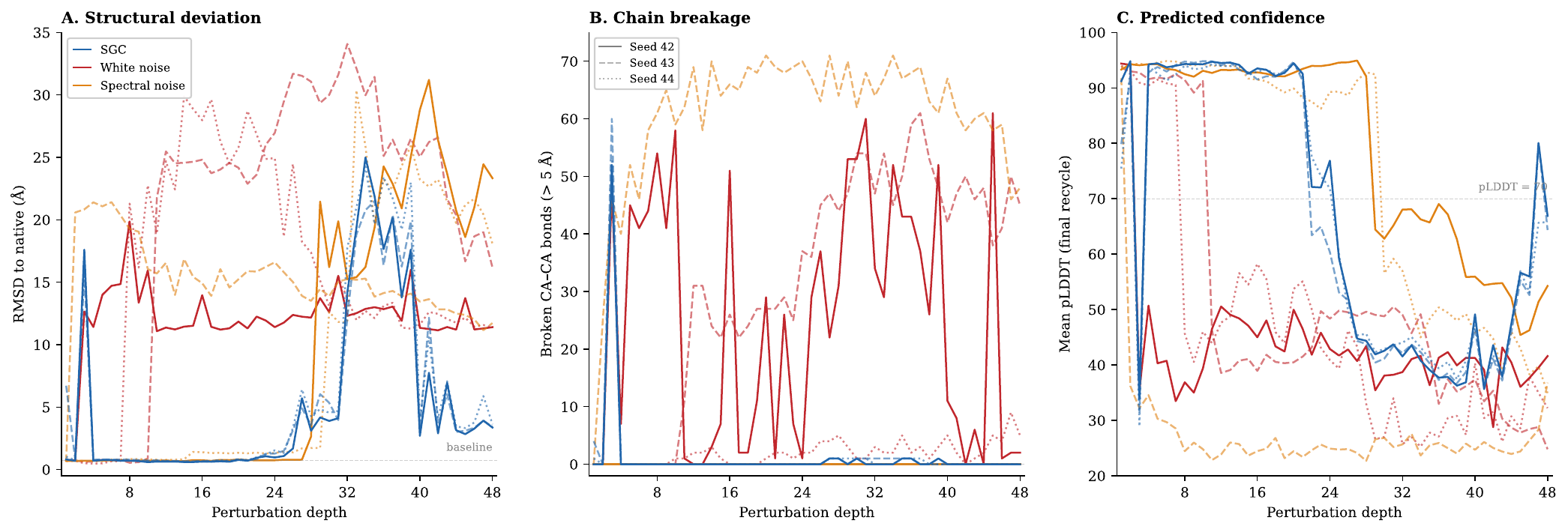}
    \caption{\textbf{Noise controls reveal that weight-aligned coherence is necessary for controllable perturbation.}
    Per-seed traces (solid = seed~42, dashed = seed~43, dotted = seed~44) for three perturbation types at matched power ($\sig = 0.30$, $\lam = 0.75$, MSA $64 \times 64$).
    \textbf{(A)}~C$\alpha$ RMSD to native. \sgc{} (blue) produces reproducible, smooth depth dependence ($\pm 0.1$--$0.6$~\AA{} across seeds). White noise (red) and spectral noise (orange) are sharply seed-sensitive: some seeds fold, others collapse, depending on the random draw.
    \textbf{(B)}~Chain breakage, measured as the number of consecutive C$\alpha$--C$\alpha$ distances exceeding 5~\AA{} (typical peptide bond: 3.8~\AA{}). Outside the block-2 anomaly (depth~3), \sgc{} structures maintain connected chains at all depths. White noise atomizes the chain (up to 58 broken bonds); spectral noise is intermediate (collapsed seed shows broken chains, surviving seeds remain connected).
    \textbf{(C)}~Predicted confidence (\plddt{}). The pLDDT head retains its trained weights but receives out-of-distribution representations under noise corruption; high pLDDT can coexist with broken chains. Block-2 anomaly visible at depth~3 for \sgc{}. For noise controls, seeds determine the random noise pattern (dedicated RNG); for \sgc{}, seeds control MSA subsampling.}
    \label{fig:noise_controls}
\end{figure}

\textbf{SGC is seed-robust; noise controls are seed-sensitive.} For noise controls, the ``seed'' determines the random noise pattern applied to the weights via a dedicated RNG; for \sgc{}, it controls MSA subsampling. At the reference condition ($\sig = 0.30$, $\lam = 0.75$), \sgc{} produces highly reproducible outcomes across three seeds. RMSD to native varies by $\pm 0.1$~\AA{} at depths 5--20 and by $\pm 0.6$~\AA{} at depth~30 (Figure~\ref{fig:noise_controls}A). The trajectory is smooth and graded, tracing a continuous path from the native basin through partial unfolding. By contrast, both noise controls produce sharply seed-dependent outcomes with two fate classes. Under white noise at depth~5, one seed folds normally (RMSD 0.8~\AA{}) while another produces an atomized structure (RMSD 14.0~\AA{}), a 17-fold difference from the same perturbation power. Spectral noise shows a similar split. One seed collapses from depth~3 onward (RMSD $> 15$~\AA{}, chains broken) while the other two remain near-native until depth~28. The outcome depends on the random draw, not on perturbation strength.

\textbf{White noise destroys chain integrity.} The most striking distinction is structural, not statistical. Under \sgc{}, excluding two isolated early-depth anomalies (depths~1 and~3; Section~\ref{sec:results-transition}), every structure at depths~2 and~4--24 maintains a fully connected polypeptide chain with all C$\alpha$--C$\alpha$ bonds below 5~\AA{}. At deep perturbation (depths 27--48), some structures show a single stretched bond (typically 5--7~\AA{}), but the chain remains topologically intact. These are recognizable unfolded or partially folded proteins. The scale of the damage is diagnostic. Across 141 runs each (excluding depth~3), 80\% of white-noise runs produce broken chains with a median of 25 broken bonds and a median maximum C$\alpha$--C$\alpha$ distance of 36.5~\AA{}, nearly 10$\times$ a peptide bond. \sgc{} produces broken chains in only 18\% of runs, nearly all limited to a single bond stretched to 5--7~\AA{} (median 6.8~\AA{}; Figure~\ref{fig:noise_controls}B).\footnote{The 18\% figure includes two depth-1 anomalies and 23 runs at depths 27--39 where a single bond stretches to 5--7~\AA{} in heavily unfolded structures. At milder $\lam$ (0.78--0.85), no run produces any broken chain across 414 structures.} White-noise structures are not unfolded proteins. They are atomized debris, physically meaningless arrangements from the model's attempt to fold with corrupted weights.

Spectral noise occupies an intermediate position with the same bimodal seed sensitivity. At shallow depths, two of three seeds produce intact structures indistinguishable from baseline (RMSD $< 1$~\AA{}, chain fully connected). At depth~30 and beyond, these two surviving seeds show high RMSD (12--16~\AA{}) but the chains remain connected. These are real, if unfolded, protein conformations, not atomized structures. The third seed (seed~43) shows broken chains from depth~3 onward, with up to 71 of 75 bonds broken, matching white-noise behavior (Figure~\ref{fig:noise_controls}B). The same perturbation type can produce either outcome depending on the random phase draw.

\textbf{A note on model confidence.} pLDDT, the model's predicted confidence, is unreliable under weight corruption (Figure~\ref{fig:noise_controls}C). The pLDDT head and Structure Module retain their trained weights; only Evoformer blocks are perturbed. But the representations they receive are far outside the training distribution. Under white noise, pLDDT values of 91 coexist with atomized structures (RMSD $> 15$~\AA{}, broken chains). The confidence head operates faithfully on out-of-distribution input and cannot be trusted. We report pLDDT for completeness but rely on RMSD and chain integrity for the noise control comparison.

\textbf{Why coherent perturbation works.} The scaling component has a clear mechanistic account. Because $\Delta W \approx (\lam - 1) W$ dominates the \sgc{} perturbation (Section~\ref{sec:methods-sgc}), every dot product in the perturbed network is rescaled by approximately~$\lam$. Attention logits, gating values, and projections are all attenuated by a predictable, weight-proportional factor. Large weights receive large kicks, small weights receive small kicks, all in the same direction. The network's information routing shifts smoothly. The smoothing component is harder to account for. It contributes $<$1\% of perturbation power yet has outsized structural effects at the transition boundary (Section~\ref{sec:results-sigma-lambda}), selectively destabilizing specific contacts while leaving others unchanged. Why local averaging of weight values produces this selectivity is not captured by the scaling argument, and we do not have a satisfactory mechanistic explanation for it.

Why do random perturbations at matched power not produce this coherent shift? One plausible mechanism involves concentration of measure. Uncorrelated noise cancels across the Evoformer's 256-dimensional projections, suppressing per-output perturbation by $\sim\!1/\sqrt{d}$ relative to total norm. This argument is suggestive but not conclusive, and we have not verified that $1/\sqrt{d}$ cancellation is the operative mechanism rather than some other property of the network's nonlinearities.\footnote{The argument applies cleanly to uncorrelated noise but less directly to spectral noise after inverse FFT. The Evoformer's nonlinear operations (softmax, layer normalization, sigmoid gating) each respond differently to correlated versus uncorrelated perturbations; a full analysis is deferred to future work.}

\textbf{Block-2 anomaly diagnostic.} The block-2 anomaly (Section~\ref{sec:results-transition}) provides an additional diagnostic. Under \sgc{}, perturbing exactly blocks $\{0, 1, 2\}$ produces catastrophic collapse (RMSD 16.7~\AA{}, 55 broken bonds). Both white and spectral noise at matched power produce stochastic single-seed collapses at depth~3 (one of three seeds in each), whereas \sgc{} produces the deterministic all-seed collapse at the reference condition. This pattern suggests that block~2 is specifically sensitive to coherent, weight-aligned attenuation rather than to random damage of equivalent magnitude.

These controls do not prove that weight-aligned coherence is the sole mechanism underlying \sgc{}'s structured output. They show that matched-power perturbations lacking a deterministic relationship between $\Delta W$ and $W$ produce either stochastic failure or no effect, not the controllable, reproducible, physically correlated response that \sgc{} delivers.

\textbf{The smoothing kernel is not unique.} A natural follow-up question is whether the Gaussian kernel specifically is required, or whether any local smoothing operation produces the same effect. We tested two alternatives, median filtering (a nonlinear rank-order statistic) and uniform (box) filtering (the simplest local average), each power-matched to the Gaussian reference on a per-tensor basis. At $\sig = 0.30$, $\lam = 0.75$, all three kernels produce indistinguishable structural outcomes: RMSD to native within 0.03~\AA{}, pLDDT within 0.4, and identical secondary structure content and chain integrity (Supplementary Table~S5). Independently parameterized dose--response sweeps (varying smoothing strength without reference to the Gaussian) confirm smooth, monotonic structural responses for both alternative kernels. This equivalence is expected. At $\sig = 0.30$, the Gaussian kernel's center weight is 0.992 and its footprint spans only three elements, a regime where all local averaging operations converge. We chose the Gaussian kernel for its analytical tractability (continuous parameterization, well-characterized mathematical properties) rather than any unique structural effect.


\subsection{The encoding is model-independent}
\label{sec:results-model-indep}

AlphaFold2 ships five separately trained models with different random initializations. The preceding sections use model~1 exclusively. If the conformational landscape we observe were a quirk of one model's weight configuration, it would undermine every claim about what the architecture has learned. We tested all five models under identical conditions: 7 ($\sig$, $\lam$) combinations, 48 perturbation depths, 3 seeds each (5{,}040 runs, 166{,}320 recycle frames total).

\begin{figure}[!htbp]
    \centering
    \includegraphics[width=\textwidth]{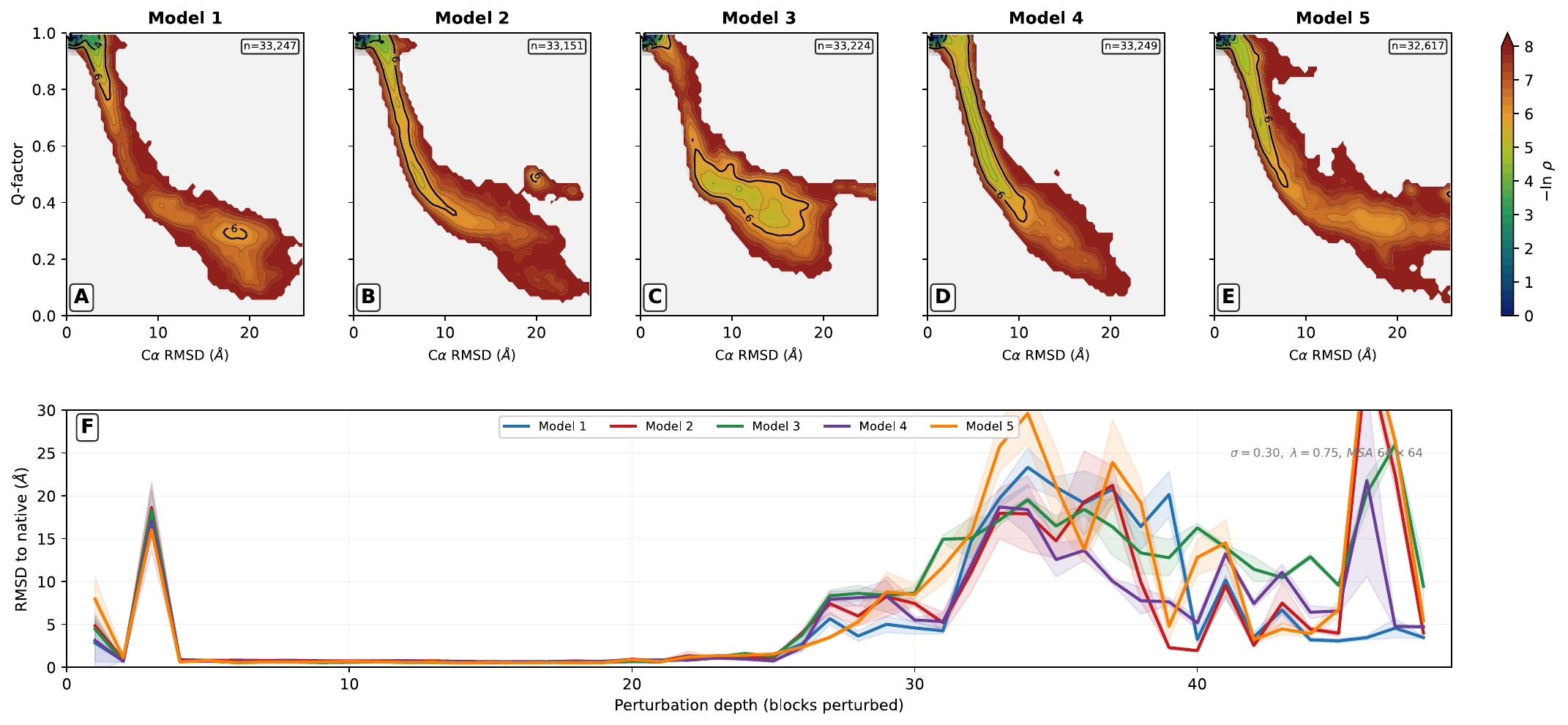}
    \caption{\textbf{The encoding is model-independent.}
    \textbf{(A--E)}~Conformational landscapes ($-\ln\,\rho$ density in RMSD--$Q$ space) for each of the five AlphaFold2 models, pooling all recycles across 7 ($\sig$, $\lam$) combinations, 48 depths, and 3 seeds (${\sim}33{,}000$ frames per model). All five models produce the same L-shaped topology: a dense native basin at low RMSD / high $Q$, a continuous ridge of partially unfolded intermediates, and an unfolded tail. White star marks the native crystal structure. RMSD axis truncated at the global 99.5th percentile (26~\AA{}).
    \textbf{(F)}~Final-recycle RMSD to native vs.\ perturbation depth at the reference condition ($\sig = 0.30$, $\lam = 0.75$), mean $\pm$ s.d.\ over 3 seeds. The structural transition occurs at depth~24--26 in all models (spread of 2 blocks). The block-2 anomaly (depth~3) is universal. Models diverge in the deep-perturbation tail, where amplitude varies but topology does not. $y$-axis capped at 30~\AA{}; maximum values reach 41.8~\AA{} (model~5).}
    \label{fig:multimodel}
\end{figure}

The five models produce the same landscape (Figure~\ref{fig:multimodel}). In the enhancement zone (depths 6--20 at $\sig = 0.30$, $\lam = 0.75$), per-model mean RMSD ranges from 0.58 to 0.74~\AA{} ($Q \approx 0.99$). All five models show two isolated early-depth anomalies. Depth~1 produces a weaker excursion (mean RMSD 2.8--8.0~\AA{} across models). Depth~3 triggers the universal block-2 collapse (mean RMSD 16.1--18.7~\AA{}), followed by recovery at depth~4. Excluding both, the structural transition, defined as the first depth where mean RMSD exceeds 1.5~\AA{}, begins at depth~24--26 in every model, a spread of two blocks. At the mild condition ($\sig = 0.30$, $\lam = 0.85$), no model exceeds 1.13~\AA{} RMSD at any depth outside the two anomalies. Pairwise rank correlations of the depth--RMSD profiles average $\rho = 0.92$ (Spearman; minimum 0.89). The models differ in the deep-perturbation tail. Seed-mean peaks range from 23.3~\AA{} (model~1) to 38.9~\AA{} (model~5), with corresponding single-seed maxima of 25.0 and 41.8~\AA{}. The L-shaped landscape topology, the transition depth, and the enhancement zone are shared.

This agreement means the conformational landscape is not stored in any single model's idiosyncratic weight values. Five independent training runs, starting from different random initializations, converged on the same landscape topology for ubiquitin's conformational space. The landscape is a property of the architecture and training data, not of one realization. Whether this convergence holds for all proteins is a separate question. In Sections~\ref{sec:results-kaib} and~\ref{sec:results-asyn}, we will see that it does not, and that the presence or absence of inter-model agreement is itself informative about what the network has learned.


\subsection{SGC shifts pair-activation magnitudes along a single axis}
\label{sec:evo_pca}

So far we have examined \sgc{} through the structures it produces. But the structures are outputs, the end of a 48-block iterative computation inside the Evoformer, whose pair representation $z_{ij} \in \mathbb{R}^{128}$ encodes the model's evolving belief about every residue pair before the Structure Module turns it into coordinates. If \sgc{} is degrading the network's ability to fold, we would expect the pair representation to deteriorate in complex, condition-dependent ways. If it is instead modulating a continuous degree of freedom in the learned representation, we would expect something simpler: a graded, parameter-ordered shift along a low-dimensional axis.

\begin{figure}[!htbp]
    \centering
    \includegraphics[width=\textwidth]{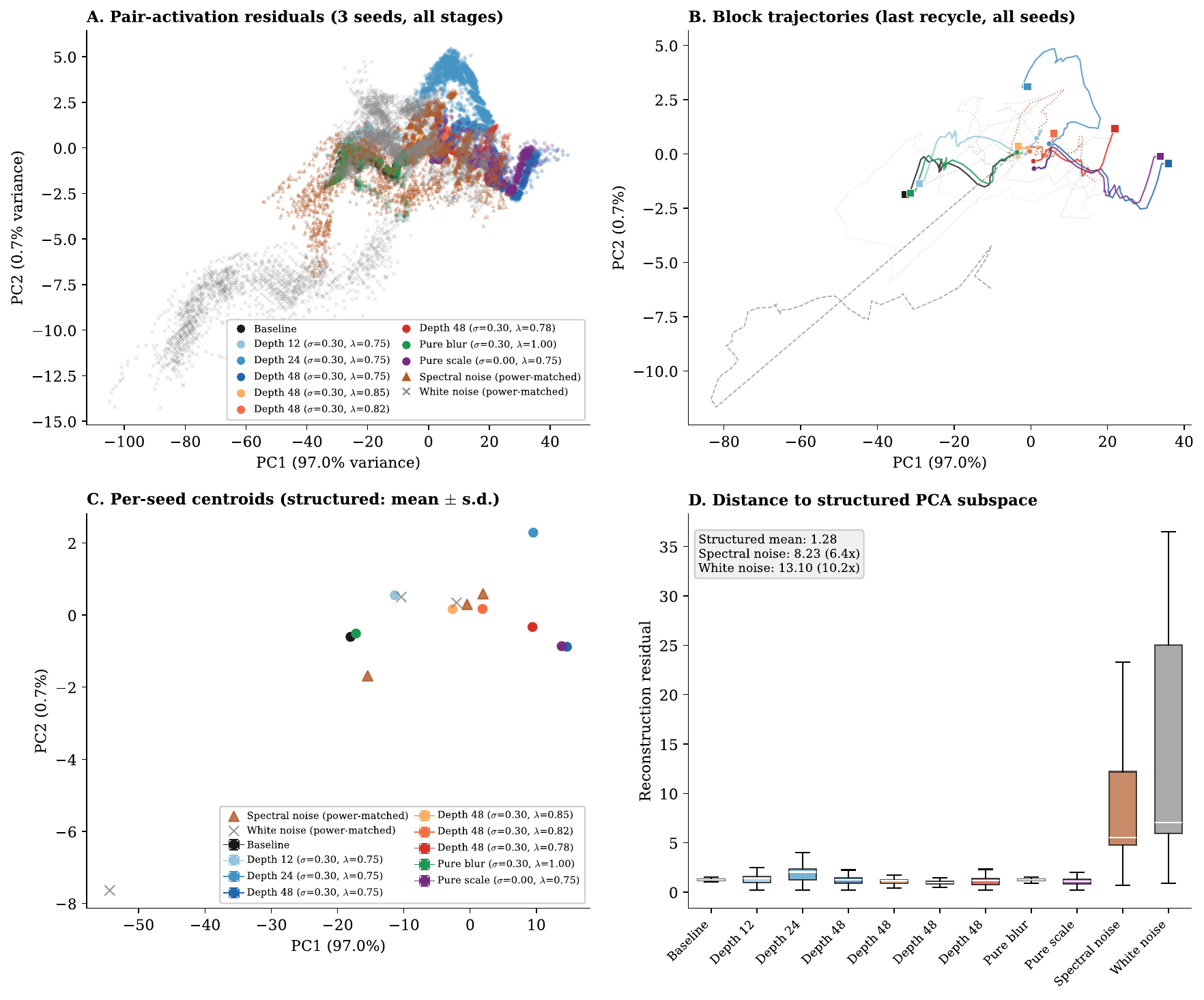}
    \caption{\textbf{SGC shifts pair-activation magnitudes along a single structured axis.}
    \textbf{(A)}~PCA of stage-residualized pair-activation magnitudes ($\operatorname{asinh}(\|z_{ij}\|_2)$, $|i-j| \geq 6$) for nine structured conditions across 3 seeds (42{,}768 snapshots). Spectral noise (orange triangles) and white noise (gray crosses) are projected out-of-sample. PC1 captures 97.0\% of condition-specific residual variance.
    \textbf{(B)}~Block trajectories (seed~42, last recycle) showing how each condition traverses the residualized PC space through the 48 Evoformer blocks. Circle = block~0; square = block~47.
    \textbf{(C)}~Condition centroids with inter-seed standard deviation as error bars (3 seeds = 3 independent replicates). Structured conditions have negligible inter-seed spread (s.d.\ 0.02--0.27 on PC1); spectral noise s.d.\ = 7.7; white noise s.d.\ = 23.0. Conditions arrange by perturbation strength along PC1.
    \textbf{(D)}~Reconstruction residual (distance to the 10-component PCA subspace fit on structured conditions). White noise: $10.2\times$ larger than structured mean (13.1 vs.\ 1.3); spectral noise: $6.4\times$ (8.2 vs.\ 1.3).
    All SGC conditions use $\sig = 0.30$, $\lam = 0.75$ unless stated otherwise. Residualization subtracts the cross-condition mean at each exact (seed, recycle, block) stage, removing the shared Evoformer processing trajectory.}
    \label{fig:evo_pca}
\end{figure}

We find the latter. We capture $z$ at all 48 blocks across 33 recycles for nine \sgc{} conditions and two noise controls, each replicated across three random seeds (52{,}272 snapshots total). For each snapshot we compute pair-activation magnitudes ($\operatorname{asinh}(\|z_{ij}\|_2)$ for all long-range pairs $|i-j| \geq 6$) and remove the shared Evoformer processing trajectory. At each exact (seed, recycle, block) stage, we subtract the mean across conditions, so that what remains is purely how each condition deviates from the average. We fit PCA on the nine structured conditions and project the two noise controls out-of-sample. A single principal component captures \textbf{97.0\%} of the condition-specific residual variance (Figure~\ref{fig:evo_pca}A).

This axis is not an artifact of the analysis. By removing the across-condition mean at each computational stage, we eliminate the shared Evoformer processing trajectory; what remains is driven by condition identity, which explains 72\% of PC1 variance. The conditions arrange along this axis in a physically interpretable order (Figure~\ref{fig:evo_pca}C). Baseline sits at one end; the depth conditions progress smoothly away from it ($d = 12 \to 24 \to 48$); the three $\lam$ variants ($0.85, 0.82, 0.78$) span a parallel gradient. Two features stand out. First, pure scale ($\sig = 0$, $\lam = 0.75$) occupies nearly the same position as the full perturbation at depth~48 ($\sig = 0.30$, $\lam = 0.75$), while pure blur ($\sig = 0.30$, $\lam = 1.00$) barely departs from baseline, confirming at the representational level that $\lam$ controls how far from native the model ventures, while $\sig$ acts as a subtle specificity modulator. Second, the depth-12 condition ($d = 12$; only blocks 0--11 perturbed) sits close to baseline despite receiving the same per-block perturbation as the depth-48 condition, because unperturbed downstream blocks partially restore the baseline trajectory (Supplementary Fig.~S9). The representational effect accumulates through perturbed blocks and is partially reversible when perturbation stops.

The noise controls tell the contrasting story. Projected out-of-sample into the structured PCA subspace, white noise has a reconstruction residual $10.2\times$ larger than structured conditions (13.1 vs.\ 1.3; Figure~\ref{fig:evo_pca}D); spectral noise is intermediate at $6.4\times$ (8.2 vs.\ 1.3). Neither noise type moves along the same representational axis as \sgc{}. Both depart into regions the structured PCA subspace cannot reconstruct.

The three-seed design adds a reproducibility dimension. For structured \sgc{} conditions, the inter-seed population standard deviation on PC1 is 0.02--0.27; three seeds of the same perturbation land in effectively the same place. For spectral noise, the inter-seed standard deviation is 7.7; for white noise, 23.0 (Figure~\ref{fig:evo_pca}C). A different random noise draw sends the representation to a different region of the subspace; a different MSA subsample under \sgc{} does not. This is the representational counterpart of the structural seed-robustness observed in Section~\ref{sec:results-noise}: the determinism originates not in the Structure Module's response, but in how the perturbation navigates the Evoformer's pair representation.

The implication is that, for ubiquitin, the conformational changes we observe under \sgc{} are not a downstream consequence of generic network failure. They begin inside the Evoformer as a continuous, one-dimensional shift in pair-activation magnitude, which the Structure Module translates into the graded structural responses described in the preceding sections.


\subsection{KaiB: convergent limits of the encoded landscape}
\label{sec:results-kaib}

KaiB is a rare fold-switching protein: its ground state adopts an unusual $\alpha/\beta$ fold, while its active state rearranges into a thioredoxin-like fold that mediates circadian clock signaling in cyanobacteria. \af{} predicts the fold-switched (thioredoxin-like) state. If the weights encode a conformational landscape around that prediction, \sgc{} should reveal whether the ground state is accessible (hidden behind a barrier that unperturbed inference cannot cross) or simply absent from the encoding.

We tested 5 models under three perturbation conditions (71{,}775 frames total). The two reference folds are from related but distinct species: the ground state from \emph{Synechococcus elongatus} (PDB 2QKE) and the fold-switched state from \emph{Thermosynechococcus elongatus} (PDB 5JYT). We therefore use TM-score rather than RMSD or $Q$-factor, as it is structure-based, length-normalized, and handles sequence-length differences natively.\footnote{The AF2 input (\emph{S.\ elongatus} PCC~7942 KaiB, UniProt Q79PF5, 102 residues) shares $\sim$85\% sequence identity with both references; species, PDB, and identity details are in Supplementary Table~S1. Repeating the experiment with the 2QKE sequence directly produces the same result: TM$_\text{gs}$ remains below 0.5 at all depths (Supplementary Fig.~\ref{fig:supp_kaib_2qke}).} A TM-score above 0.5 indicates the same fold; cross-reference TM(2QKE, 5JYT) $= 0.42$, confirming the two folds are structurally distinct.

\begin{figure}[!htbp]
    \centering
    \includegraphics[width=\textwidth]{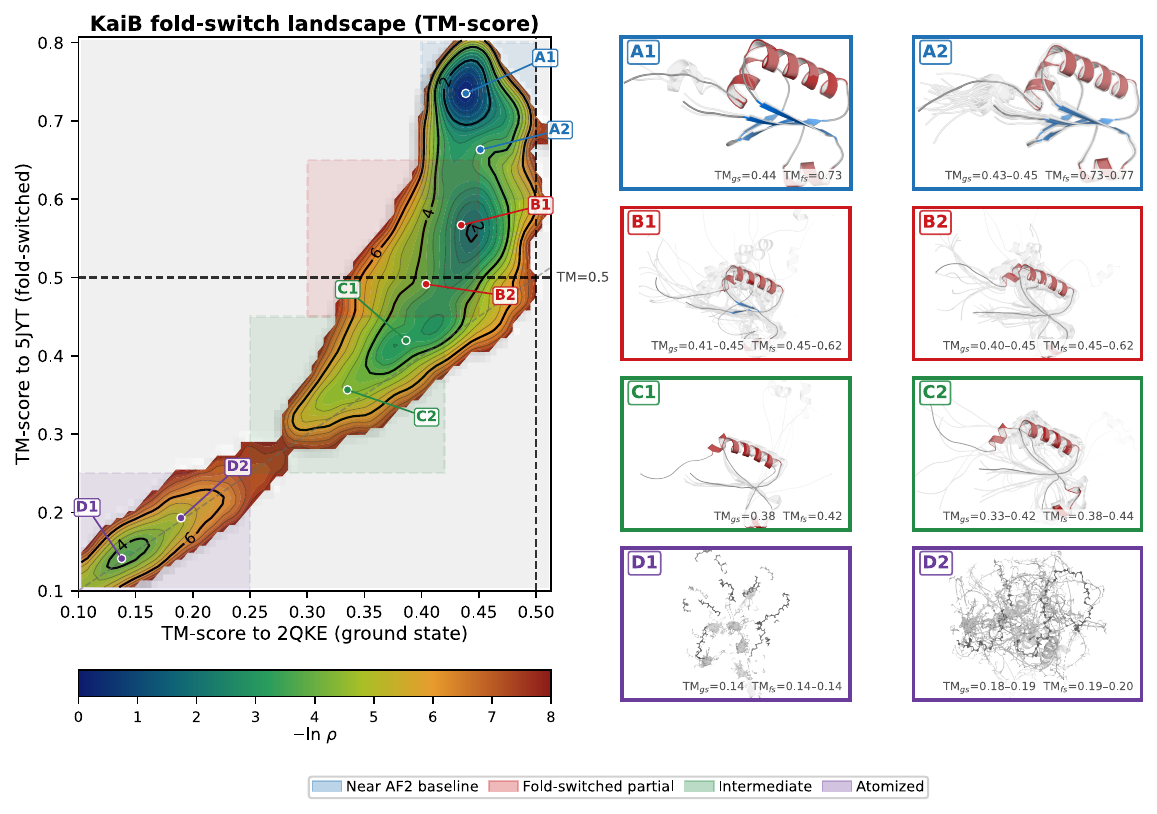}
    \caption{\textbf{KaiB: the conformational landscape is one-dimensional denaturation, not fold-switching.}
    Density landscape in TM-score space (TM to ground state 2QKE vs.\ TM to fold-switched state 5JYT), pooling 71{,}775 frames from all 5 AlphaFold2 models. The \af{} baseline (blue square) sits in the fold-switched basin (TM$_\text{fs} > 0.5$, TM$_\text{gs} < 0.5$). Perturbation drives structures diagonally downward, losing similarity to both folds simultaneously, with no orthogonal branch toward the ground-state corner ($\text{TM}_\text{gs} > 0.5$, lower right). Structural insets show representative conformations from four landscape regions: \textbf{(A)}~near-baseline helical bundle, \textbf{(B)}~partial helices, \textbf{(C)}~disordered intermediates, \textbf{(D)}~fully denatured. Organized $\beta$-sheet (hallmark of the ground state) never appears. Red triangle marks the ground-state reference; blue square marks the fold-switched reference.}
    \label{fig:kaib}
\end{figure}

The landscape is fundamentally one-dimensional (Figure~\ref{fig:kaib}). TM-scores to the ground state and fold-switched state are \emph{positively} correlated across all five models ($r = 0.70$--$0.80$), meaning that perturbation drives both scores down simultaneously. The structures denature rather than switch. At no perturbation depth does an orthogonal branch emerge toward the ground-state basin. The structural progression is monotonic. Intact helical bundle $\to$ partial helices $\to$ disordered loops $\to$ fully denatured (Figure~\ref{fig:kaib}, insets). Organized $\beta$-sheet, the hallmark of the ground-state fold, never appears.

\begin{figure}[!htbp]
    \centering
    \includegraphics[width=\textwidth]{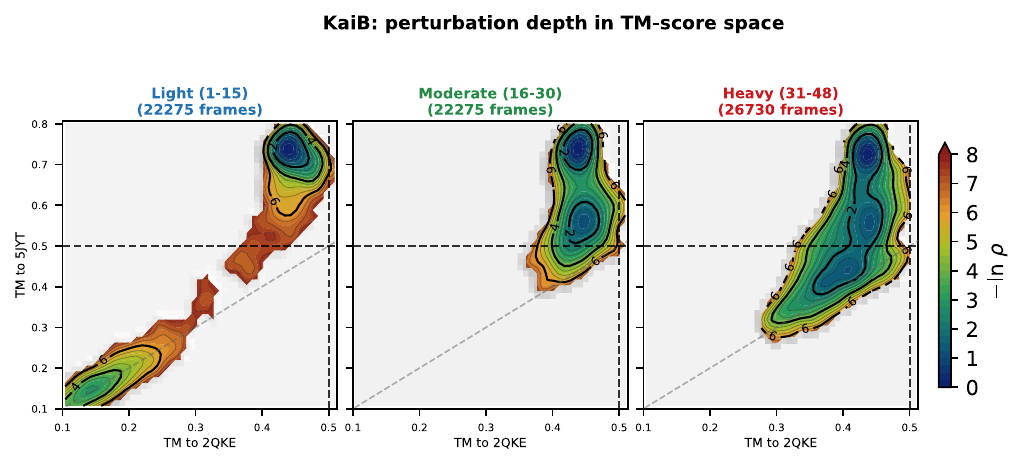}
    \caption{\textbf{Perturbation depth controls denaturation extent, not switching direction.}
    KaiB TM-score landscapes split by perturbation depth (all 5 models, $\sig = 0.30$, $\lam = 0.75$). Most light-perturbation depths (1--15, excluding the block-2 anomaly at depth~3 and a condition-dependent dip at depth~7) remain near baseline. Moderate depths (16--30) migrate diagonally, losing similarity to both folds. Heavy perturbation (31--48) produces a broad denatured basin. At no depth does the landscape branch toward the ground-state corner.}
    \label{fig:kaib_depth}
\end{figure}

Perturbation depth controls how far along this denaturation axis the model ventures (Figure~\ref{fig:kaib_depth}), but two early-depth anomalies interrupt the progression. Depth~3 produces a sharp collapse across all conditions and all five models, paralleling the block-2 anomaly observed for ubiquitin (Section~\ref{sec:results-transition}). Depth~7 is a second local dip at the two $\lam = 0.75$ conditions, recovering by depth~8; at $\lam = 0.85$ it is absent. Outside those anomalies, light perturbation (depths 1--2 and 4--15, excluding depth~7 at $\lam = 0.75$) remains near the \af{} baseline (TM$_\text{fs} \approx 0.65$--$0.75$, TM$_\text{gs} \approx 0.40$--$0.48$). At moderate depths (16--30), structures migrate diagonally, losing similarity to both folds. At heavy depths (31--48), the cloud expands into a broad denatured basin at TM $< 0.3$ for both references.

Only 31 of 71{,}775 frames (0.04\%) cross the nominal TM$_\text{gs} > 0.5$ threshold, and these are not ground-state-like structures. The first are shallow-depth threshold fluctuations (models~1, 4, 5): TM$_\text{gs}$ barely exceeds 0.50 while TM$_\text{fs}$ remains $\sim$0.67, still firmly fold-switched. The second are moderate-depth structurally intermediate frames from model~3 (TM$_\text{gs} \approx$ TM$_\text{fs} \approx 0.50$), equidistant from both folds with no evidence of 2QKE-topology $\beta$-sheet arrangement. Model~2 never crosses the threshold at any depth. Per-model landscapes are provided in Supplementary Figure~\ref{fig:supp_kaib_per_model}.

All five models agree on this result. The absence of fold-switching is not a quirk of one model's weights. It is shared across all five independently trained variants. This convergent absence mirrors the convergent presence observed for ubiquitin (Section~\ref{sec:results-model-indep}). Where the training data appears to encode a conformational landscape, all models recover it under \sgc{}; where no landscape was detected under the parameters tested, none of the models produced one. Within the tested parameter regime, \sgc{} detects no alternative-fold basin.


\subsection{$\alpha$-synuclein: structured disagreement at the edge of training support}
\label{sec:results-asyn}

$\alpha$-synuclein is a 140-residue intrinsically disordered protein implicated in Parkinson's disease, whose misfolding and aggregation into amyloid fibrils is a hallmark of the pathology. In solution it has no single native state, sampling a broad ensemble of compact and extended conformations. Because there is no reference structure, we replace RMSD and $Q$-factor with radius of gyration ($R_g$) and end-to-end distance ($R_{ee}$), coordinates that characterize chain compactness without assuming a target fold. We tested all five models under three perturbation conditions (141{,}375 frames total).

\begin{figure}[!htbp]
    \centering
    \includegraphics[width=\textwidth]{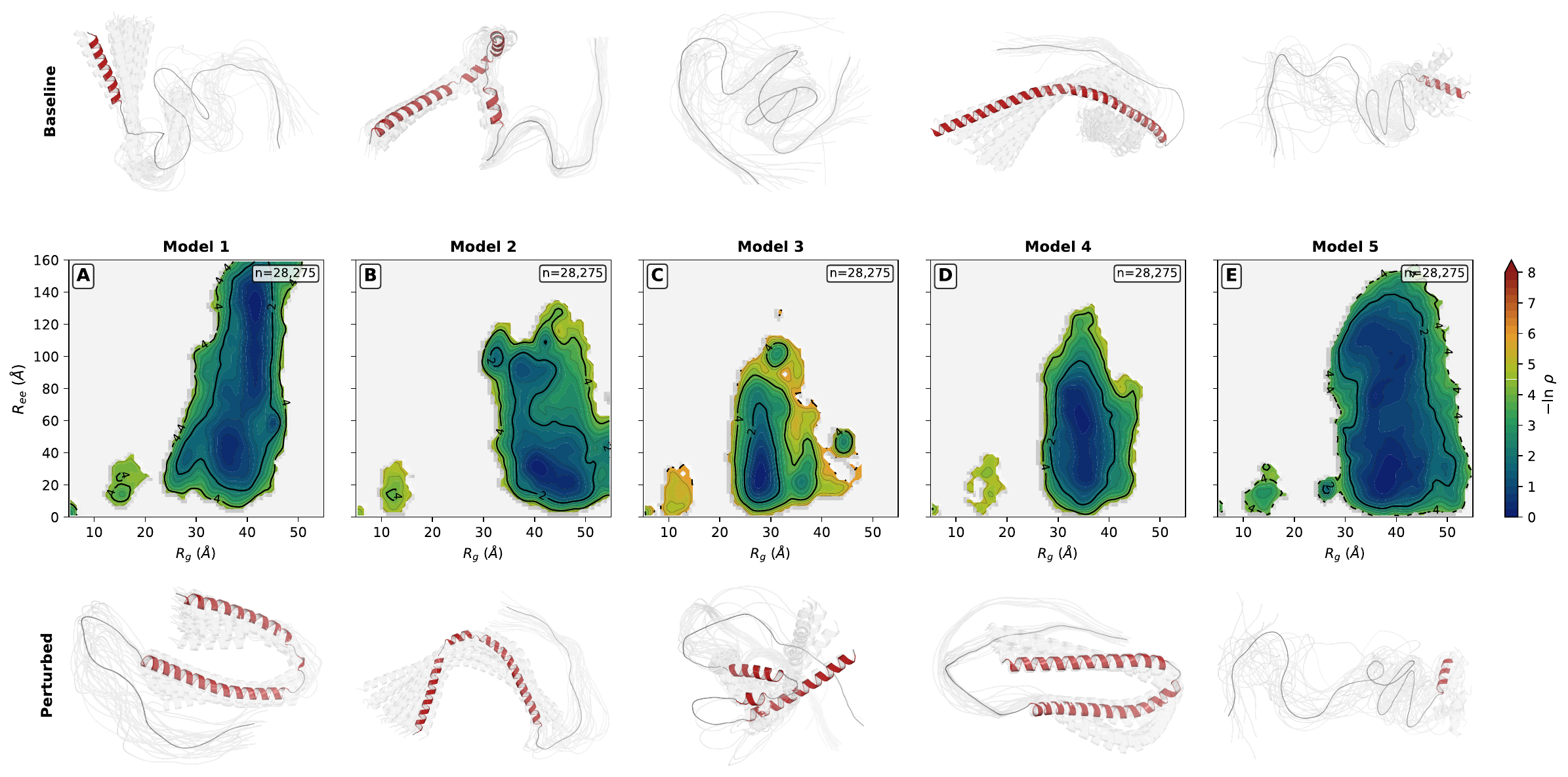}
    \caption{\textbf{$\alpha$-synuclein: five models, five landscapes.}
    \textbf{(A--E)}~Per-model conformational landscapes ($-\ln\,\rho$ density in $R_g$--$R_{ee}$ space), pooling all recycles across 3 perturbation conditions, 48 depths, and 3 seeds (${\sim}28{,}000$ frames per model). Each model produces a distinct landscape topology: model~2 favors extended helical states, model~3 favors compact conformations, models~1, 4, 5 are intermediate. Inter-model coefficient of variation on $R_g$ is 14.6\% (6$\times$ the ubiquitin value).}
    \label{fig:asyn}
\end{figure}

The five models produce five different landscapes (Figure~\ref{fig:asyn}). Even before perturbation, the baselines diverge. Model~2 predicts an extended helical conformation ($R_g = 31.9$~\AA, 45\% helix). Model~3 predicts a compact state with no helix and 8\% $\beta$-sheet ($R_g = 25.6$~\AA). Model~4 predicts nearly 50\% helix at $R_g = 33.5$~\AA. Under \sgc{}, these differences amplify. The inter-model coefficient of variation on final-recycle mean $R_g$ is 14.6\%, compared to 2.3\% for ubiquitin, a sixfold increase (one-way ANOVA on final-recycle values, $F = 332$, $p < 10^{-222}$; ubiquitin $F = 7.8$). This is not seed noise. At the primary operating point ($\sig = 0.30$, $\lam = 0.75$), the per-model inter-seed coefficient of variation on $R_g$ is 0.1--0.9\%. At the mild condition ($\lam = 0.85$), individual model baselines show greater seed sensitivity (Supplementary Table~\ref{tab:supp_ss}). The disagreement is between models, not between seeds of the same model. This per-model divergence is already present at the unperturbed baseline and persists at every perturbation depth (Supplementary Fig.~\ref{fig:supp_asyn_depth}), ruling out a depth-specific perturbation artefact.

\begin{figure}[!htbp]
    \centering
    \includegraphics[width=\textwidth]{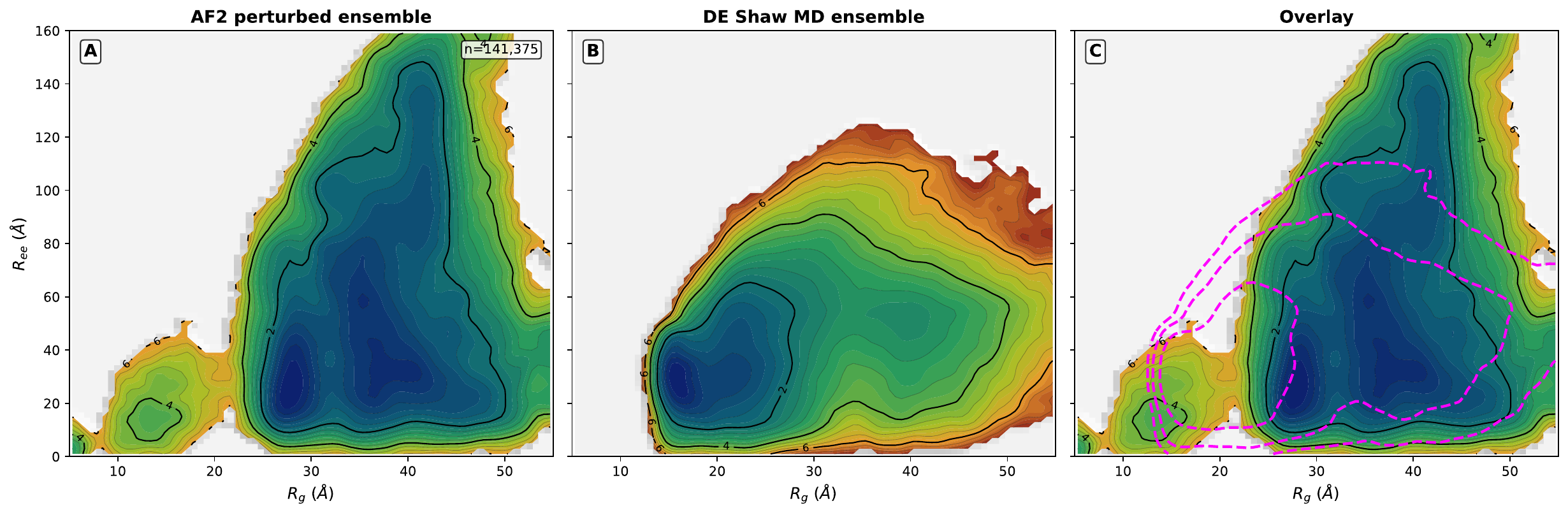}
    \caption{\textbf{$\alpha$-synuclein: AF2 ensemble covers the MD-sampled core.}
    \textbf{(A)}~Pooled AF2 landscape across all 5 models (141{,}375 frames) in $R_g$--$R_{ee}$ space.
    \textbf{(B)}~DE~Shaw explicit-solvent MD ensemble pooled across eight force-field variants ($\sim$273~$\mu$s total; Supplementary Table~S1) in the same coordinate space.
    \textbf{(C)}~Overlay: AF2 density (filled) with MD $-\ln\,\rho = 2, 4, 6$ contours (dashed magenta). The AF2 ensemble encompasses the MD-accessible region and extends beyond it in model-specific directions.}
    \label{fig:asyn_md}
\end{figure}

Yet the models are not generating arbitrary structures. When pooled across all five models (Figure~\ref{fig:asyn_md}A), the \sgc{} ensemble covers the core of the $R_g$--$R_{ee}$ landscape sampled by explicit-solvent molecular dynamics (DE~Shaw, multiple force fields; Figure~\ref{fig:asyn_md}B). The overlay (Figure~\ref{fig:asyn_md}C) shows that the AF2 ensemble encompasses the MD-accessible region and extends beyond it in model-specific directions. The shared core, where all models agree and MD also finds density, is consistent with constraints shared across training data and physical simulation. The model-specific extensions are structured (each with characteristic secondary-structure content) but not shared. The pooled MD reference is itself heterogeneous. Across the eight force fields, some trajectories remain compact-biased whereas others explore broader $R_g$--$R_{ee}$ space, and no single trajectory recapitulates the pooled landscape (Supplementary Fig.~\ref{fig:supp_asyn_md_per_ff}). A per-model, per-force-field comparison would be more informative than comparison to a single pooled reference, but we do not attempt that mapping here.

This is the opposite end of the spectrum from ubiquitin. Where the PDB provided an unambiguous structural target, all five models converged on the same conformational landscape (Section~\ref{sec:results-model-indep}). Where the training signal is underdetermined (as it is for an IDP with no single stable fold), each model arrives at a different but internally coherent representation. We interpret these differences as structured model-dependent extrapolations rather than random failures, five independently derived responses to the same training signal. Each is consistent with structural patterns the model appears to have learned from training data, but not uniquely determined by those patterns. What all five models agree on is testable. We predict that the shared-core region of ($R_g$, $R_{ee}$) space (Figure~\ref{fig:asyn_md}C) will contain the experimentally occupied region of any future ensemble with sufficient sampling of $\alpha$-synuclein's disordered state. The model-specific extensions are open hypotheses about under-determination, not competing predictions. We return to this observation in the Discussion.

\section{Discussion}
\label{sec:discussion}


\textbf{Status of claims.} The results in this paper operate at three epistemic levels. The structural observations---contact-loss ordering, flexibility correlations, noise-control contrasts, inter-model convergence---are measurements, reproducible across seeds and models. The interpretation that these reveal a physically structured encoding in the Evoformer weights is well-supported but not proven. Alternative accounts, including smooth failure mode and architectural artifact, are not excluded. The broader claims that the encoding reflects evolutionary constraints, and that weight-perturbation probing generalizes beyond this model, are speculation. We flag them as such in Section~\ref{sec:discussion-speculation}.

\subsection{Neural spectroscopy and the structure of the response}
\label{sec:discussion-spectroscopy}

\textbf{The signal is structured, not random.} AlphaFold2 was trained on static structure targets, yet under a perturbation this simple its weights yield a structured conformational response rather than generic network capacity loss. The evidence is specific. Native contacts break in the experimentally established stability ordering, with the folding nucleus outlasting peripheral contacts in every informative condition (Section~\ref{sec:results-unfolding}). Perturbation sensitivity tracks molecular dynamics RMSF beyond what native-state geometry alone explains (Section~\ref{sec:results-rmsf}). Matched-power noise controls produce qualitatively different outputs: atomized debris or stochastic coin flips, never graded progressions (Section~\ref{sec:results-noise}). The response is kernel-independent and consistent across five independently trained models (Sections~\ref{sec:results-noise},~\ref{sec:results-model-indep}). A similar enhancement zone persists across MSA depths $32 \times 32$ through $128 \times 128$, though behaviour at default depth is untested. These observations are inconsistent with indiscriminate corruption.

\textbf{We frame the observation in terms of three levels.} To make sense of this response, we find it useful to distinguish three levels. At the first, \sgc{} deforms what we call the weight-space geometry: the specific configuration of learned parameters across the Evoformer's 48 blocks. Scaling attenuates each block's parameters. Smoothing perturbs their local correlations. At the second level, the Evoformer applies these modified rules to a protein's sequence and alignment and constructs a different representational geometry. The pair and MSA representations, which exchange information across all 48 blocks, jointly encode the model's belief about the protein. PCA on the pair representation captures the effect of perturbation on one of these streams. Structured perturbations arrange along a single axis accounting for 97\% of variance, graded by perturbation strength. Noise controls project outside this subspace entirely (Section~\ref{sec:evo_pca}). At the third level, the Structure Module decodes the altered representation into three-dimensional coordinates. The resulting structures form a conformational landscape whose topology converges with millisecond-scale molecular dynamics. We can characterize each level empirically. What we do not yet have is a theory connecting them.

\textbf{This response is not built into the training objective.} The model was trained to minimize coordinate error on static structures, using evolutionary covariance as input. It has no temporal axis, no thermodynamic loss, no concept of folding pathways. The 48 Evoformer blocks are a computational sequence. Recycling is iterative refinement toward a single prediction. The perturbation depth axis, along which we observe ordered structural change, is a parameter of our probe that the model was never trained to respond to. The correspondence with physical processes emerged from a training process that rewarded only endpoint accuracy.

\textbf{Two hypotheses, neither proven.} Why does continuous deformation of a computation trained on static snapshots produce output that traces the folding landscape? We offer two compatible hypotheses. First, generalization from evolutionary data may require learning which structural constraints are critical to a fold. The MSA encodes which residue-residue relationships are under selection. The model may have internalized this as differential encoding robustness: constraints most strongly conserved across the evolutionary record may be encoded most robustly. These may be the same constraints that stabilize the fold physically, because evolutionary conservation and thermodynamic stability may both favor the contacts that hold it together. This would account for why perturbation sensitivity recovers the pattern of MD-measured flexibility (which residues move) without recovering its magnitude (how much they move). The training objective penalizes coordinate error, giving the model every reason to learn which residues anchor the fold and no reason to learn how far they stray. Second, the sequential residual architecture may impose a computational hierarchy that, under cumulative perturbation, produces a natural degradation order. This is architectural speculation. Single-block perturbation profiling could test it directly.

\textbf{Memorization alone does not explain the connectivity.} A natural concern is that ubiquitin, among the most heavily represented proteins in the PDB, could simply be memorized. We cannot rule this out with the present data. But memorization of individual conformations does not explain what we observe. The PDB contains snapshots: individual structures at various levels of resolution. It does not contain the path connecting native to denatured through specific intermediates. A lookup table, corrupted, produces a wrong entry or garbage. It does not produce a continuous, ordered progression through recognizable structural states. What requires explanation is not which conformations are accessible but that they are connected by a trajectory whose ordering has physical correspondence. That connectivity emerged from training. Whether it reflects learned generalization, an architectural property of deep residual networks, or their interaction is the question that data ablation and architectural comparison could address.

\textbf{The signal sits in dominant directions.} That a blunt perturbation recovers this much structure constrains where the information sits. \sgc{} is dominated by a near-uniform rescaling, $W \leftarrow 0.75W$, with $\sig$ adding a small orthogonal local-smoothing component. A near-uniform rescaling acts most strongly on the dominant, high-magnitude structure of the weights. If the conformational signal were instead stored in fine-grained component interactions, uniform scaling would not find it. The two parameters probe distinct aspects of the response. $\lam$ moves the representation along the exploration axis through uniform attenuation. $\sig$ perturbs local weight correlations, resolving contact-level distinctions despite contributing $<$1\% of the perturbation power (Supplementary Section~\ref{sec:supp-power-decomp}).

We propose the term \emph{neural spectroscopy} for this approach. It denotes systematic, controlled perturbation of a trained network's weights to characterize what it has learned, by analogy with physical spectroscopy. Apply a controlled stimulus, measure the structured response, infer internal properties. Mechanistic interpretability traces specific circuits and attention heads~\cite{elhage2021,olsson2022}; targeted model-editing methods such as ROME~\cite{meng2022} test whether a particular association is stored at a particular site. \sgc{} instead perturbs the learned system globally and asks whether the aggregate response has interpretable structure. In that sense it is closer to a continuous weight-space lesion with a structured functional readout than to circuit tracing. The biological result here is specific to AlphaFold2, but the approach is not. Architecturally distinct protein structure prediction models (ESMFold~\cite{lin2023}, AlphaFold3~\cite{abramson2024}) are natural next candidates; more broadly, any trained network whose outputs carry domain-specific structure may be amenable to similar probing.

The formal account connecting weight-space geometry, through the construction of representational geometries, to the emergence of conformational topology is work we defer---not because it is unimportant, but because we do not yet have it. We have characterized the three levels empirically: the weight deformation is continuous and structure-preserving; the representational response is low-dimensional and graded; the conformational output has physical correspondence in topology, ordering, and sensitivity pattern. That these levels chain together to produce physically structured output from a model never trained on physical processes is the finding we understand least.

\subsection{Shared, absent, and model-contingent representations}
\label{sec:discussion-representations}

\textbf{The three proteins probe the same computation under different input constraints.} Section~\ref{sec:discussion-spectroscopy} developed the three-level framework from ubiquitin, where the training signal is strong and unambiguous. The weights are the same 93 million parameters for all three proteins. The computation is the same 48-block sequential refinement. What changes is the input. The sequence and the evolutionary information it carries differ across proteins. Ubiquitin, KaiB, and $\alpha$-synuclein provide three different substrates for the same learned rules, and the outcomes differ because the inputs differ in strength and specificity. These outcomes---convergent presence, convergent absence, and structured disagreement---tell us what the training signal provided for each protein.

\textbf{Inter-model agreement under perturbation goes beyond prediction agreement.} Five independently trained models share the same perturbation-response geometry under \sgc{} (Section~\ref{sec:results-model-indep}). In the three-level framing, Level~1 varies: five training initializations produce five different weight-space geometries. Yet the conformational landscapes converge. The input is strong enough to wash out the variation in learned weights. Ubiquitin and structurally similar folds are abundantly represented in the training data. All five training runs encountered this fold class frequently enough to internalize the rules of computation that generate a rich representational geometry for it. Whether those rules are specific to well-represented fold classes or are general principles that yield rich representations whenever the input is sufficiently constrained remains an open question.

\textbf{Consistent non-recovery across models is also informative.} KaiB fold-switches as its biological function. The switch is intrinsic. The protein interconverts between folds in isolation, without binding partners~\cite{waymentsteele2024kaib}. The ground-state fold is real and biologically accessible. And yet, \sgc{} produces only one-dimensional denaturation from the fold-switched state, with no branch toward the ground-state basin at any perturbation depth, consistently across all five models (Section~\ref{sec:results-kaib}). Even with the ground-state cognate sequence as input, the result is unchanged (Supplementary Figure~\ref{fig:supp_kaib_2qke}). Chakravarty and Porter~\cite{chakravarty2024} showed that \af{}'s fold-switching predictions are driven by structure memorization rather than learned coevolutionary signal. The convergent absence we observe is consistent with this finding. The general capability to construct $\beta$-sheet geometries exists in the weights. \af{} builds them routinely for other proteins. What appears to be absent is the specific pathway that would harness this general capability and direct it toward a ground-state representational geometry for KaiB's input.\footnote{Within the parameter ranges tested here ($\sig = 0.20$--$0.30$, $\lam = 0.75$--$0.85$, 48 perturbation depths, 3 seeds per condition). A broader parameter sweep or a qualitatively different perturbation scheme could in principle reach regions of the manifold that \sgc{} at these settings does not.} \sgc{} maps this as a boundary of the representational manifold.

\textbf{Structured disagreement reveals underdetermination.} $\alpha$-synuclein has no single stable fold. No monomeric solution structure exists for it in the PDB~\cite{ruff2021}. The input does not uniquely determine what the computation should produce for this protein. Five models, confronted with this ambiguity, produce five different conformational landscapes (Section~\ref{sec:results-asyn}). The divergence is not seed noise. Inter-seed coefficient of variation on radius of gyration is 0.1--0.9\%. Inter-model coefficient of variation is 14.6\%. The disagreement is present at the unperturbed baseline and persists at every perturbation depth. \sgc{} amplifies a pre-existing representational difference. At Level~3, the five models produce visibly different landscapes. Per our framework, this implies that the representational manifolds at Level~2 differ as well, though we have not measured them directly for $\alpha$-synuclein. Each model has derived its own internally coherent interpretation of the same underdetermined input. These five interpretations are structured extrapolations from the training signal, shaped by stochastic differences in learning history.

\textbf{The shared core yields the most falsifiable prediction of this study.} When all five models are pooled, the \sgc{} ensemble covers the core of the $(R_g, R_{ee})$ landscape sampled by explicit-solvent molecular dynamics (Section~\ref{sec:results-asyn}; Figure~\ref{fig:asyn_md}C). The shared-core region in ($R_g$, $R_{ee}$) space is where the input, despite its ambiguity, was strong enough to constrain all five models. We predict that this region will contain the experimentally occupied ensemble for any future characterization of $\alpha$-synuclein's disordered state with sufficient sampling. The model-specific extensions are structured, reproducible, and internally coherent, but they are not shared. They are open hypotheses about what the input leaves underdetermined.

The three-protein pattern reveals how the input's information content modulates the three-level emergence. Strong signal constrains the representational manifold across models, producing a shared landscape with physical correspondence. Absent signal produces convergent absence, where the manifold encodes only what the input supports and \sgc{} reports what is missing. Ambiguous signal produces structured disagreement, with the shared core marking the floor of what the input reliably encodes. These distinctions exist in the learned representations.

\subsection{Speculation}
\label{sec:discussion-speculation}

What follows extends beyond what our evidence establishes. We offer it as a framework for thinking about what we observed.

\textbf{The emergence puzzle.} The open question is why continuous deformation of a computation trained on static snapshots produces conformational output with physical correspondence. One reading is that the training process, optimizing for accurate structure prediction across diverse proteins, was forced to learn structural constraints with differential robustness. Contacts critical to a fold must be encoded robustly because getting them wrong incurs the largest training loss. Contacts peripheral to the fold can be encoded less precisely. This differential encoding, when probed by continuous attenuation, would produce ordered contact loss. The most robustly encoded contacts would survive the deepest perturbation. That this ordering matches the experimentally known stability hierarchy would then follow from the evolutionary signal in the input, which encodes constraint importance through the same selective pressure that determines physical stability. We cannot establish this from three proteins. But it is the reading most consistent with our observations, and the proteome-scale screen proposed in Section~\ref{sec:discussion-future} is designed to test it.

\textbf{The architecture has error-correction properties.} An observation about the Evoformer is relevant to interpreting KaiB. The pair representation is not a collection of independent pairwise beliefs. Triangle multiplicative updates and triangle attention refine every pairwise relationship through all possible intermediary residues, enforcing geometric consistency across residue triplets throughout the protein~\cite{jumper2021}. This produces a self-reinforcing computation that converges on a globally coherent structure. The same property may contribute to the observation that \af{}'s predicted structures are often insensitive to single-residue mutations~\cite{pak2023}, though structural effects of individual mutations can sometimes be detected~\cite{mcbride2023}. Triangle relationships may restore consistency for most local perturbations. \sgc{} operates differently. It attenuates the enforcement mechanism itself across many blocks simultaneously, weakening the computation globally rather than perturbing it locally. This distinction matters for understanding why certain conformational states may be accessible under \sgc{} while remaining invisible to local input changes.

\textbf{Why is the ground-state valley missing?} KaiB's convergent absence raises a question that Section~\ref{sec:discussion-representations} deliberately left open. The ground-state fold is real. The computation has the general capability to construct $\beta$-sheet geometries. Why is the specific pathway to a ground-state representation absent for this protein?

Several factors could contribute, and we suspect they compound. Consider first the structural distance between the two states. KaiB's fold switch involves different secondary structures in the same residues across roughly 40\% of the protein (TM-score between the two folds is 0.42). This is not a loop rearrangement or a domain rotation. The alternative fold requires qualitatively different pairwise relationships across a large fraction of the residue pairs. Whether the representational manifold can span such distances is an open question. It is possible that the manifold is locally rich, able to represent conformational variants near the states it was constructed from, but does not extend to topologically distant regions of structure space.

Consider next the training signal. Chakravarty and Porter~\cite{chakravarty2024} showed that \af{}'s fold-switching predictions track structure memorization from the PDB. If the ground-state fold is underrepresented in the training data, the model may never have established the specific input-to-representation mapping needed to construct it. But training data alone may not be the full story. The evolutionary signal in the input could in principle encode constraints for both folds, but the strength and separability of that signal for KaiB is something we have not measured. Schafer and Porter~\cite{schafer2025} showed that even when \af{} does predict fold-switched conformations, the success is powered by training-set presence rather than learned coevolutionary signal, suggesting that the MSA alone may not carry sufficient information to distinguish the two folds.

Finally, consider the architecture. The triangle operations enforce geometric consistency across the entire protein. Once the early Evoformer blocks establish a set of pairwise relationships consistent with the fold-switched state, the subsequent blocks refine and reinforce this geometry. A minority signal pointing toward an alternative fold, even if present, would be geometrically inconsistent with the majority and could be suppressed through this reinforcement. Whether this actually happens in the KaiB computation is untested. Per-head perturbation profiling could probe it directly: selectively isolating individual attention heads might reveal whether any carry sub-dominant signals compatible with the ground state, or whether the signal is entirely absent at every level of the computation.

We cannot currently distinguish these factors. The structural distance, the training signal, and the architectural reinforcement may all contribute. What we can say is that the manifold, as mapped by \sgc{} at the parameter ranges tested here, does not extend to KaiB's ground-state fold.

\textbf{What might the model-specific landscapes encode?} For $\alpha$-synuclein, the speculation takes a different form. Five models produce five structured, internally coherent landscapes. Section~\ref{sec:discussion-representations} attributed this to underdetermination of the input. The manifold framework suggests a further question: are these five interpretations biologically meaningful, or are they arbitrary consequences of stochastic training differences?

If the representational manifold reflects evolutionary constraints carried by the input, and the input for a multi-functional IDP encodes diffuse constraints from diverse structural contexts (membrane binding, partner interactions, disorder maintenance), then different models, resolving the ambiguity differently, may have converged on different functional facets of the conformational space. Model~2's 45\% helical structures are consistent with $\alpha$-synuclein's known membrane-binding mode~\cite{ulmer2005}. Model~3's compact, $\beta$-containing structures share features associated with aggregation-prone conformations~\cite{tuttle2016}. We do not claim these correspondences are established. We lack experimental validation. But the prediction is specific: if model-specific landscapes correspond to identifiable functional contexts, that would support the hypothesis that the manifold reflects structural constraints from the evolutionary training signal even when the input is ambiguous. If the landscapes correspond to nothing biologically interpretable, the manifold for underdetermined proteins reflects training noise rather than evolutionary signal. Either outcome would be informative.

\textbf{The representational geometry as the deeper object.} These speculations converge on a single point. The representational geometry that the Evoformer constructs for each protein is the object that determines what \sgc{} can and cannot reveal. We have glimpsed this geometry through pair-activation PCA on one representation stream, along one axis, for conditions on one protein. Its full structure remains unexplored. Its dimensionality, its topology, how it varies across proteins, and how it relates to physical properties are all open questions.

A methodological challenge accompanies the scientific one. Five independently trained models produce five different weight-space geometries. For ubiquitin, these converge at Levels~2 and~3. For $\alpha$-synuclein, they diverge. Understanding what drives this convergence or divergence requires an analytical framework for comparing representational geometries across models. Are two models' manifolds related by some transformation, or are they fundamentally different? How might we characterize the difference? We do not yet have such a framework.

Protein language models such as ESM~\cite{lin2023}, which predict structure from sequence alone without MSA input, offer the sharpest comparison for separating what the sequence encodes from what the coevolutionary signal adds. Weight perturbation applied to ESMFold would test whether the manifold's richness depends on the MSA channel or is already present in the sequence-trained weights. More broadly, each protein's representational manifold may encode a compressed version of its conformational landscape, capturing accessible states shaped by structural constraints reflected in the evolutionary and structural training data. If so, the representational manifold should be analyzed directly, and neural spectroscopy provides one empirical window into it. The formal theory connecting weight-space geometry to representational geometry to conformational topology is the work we defer to future investigation.

\subsection{Limitations}
\label{sec:discussion-limits}

The primary operating regime is reduced-MSA. All primary analyses use MSA depth $64 \times 64$ to place structural control in the learned weights rather than the input alignment (Section~\ref{sec:methods-impl}). Unfolding order and landscape topology persist across MSA depths $32 \times 32$, $64 \times 64$, and $128 \times 128$, though the transition onset shifts to deeper blocks at higher MSA depth (Supplementary Section~\ref{sec:supp-msa-depth}; Supplementary Fig.~\ref{sfig:msa-sensitivity}). But that comparison covers one protein at 7 of 192 parameter settings, spanning a narrow range relative to the default ($512 \times 5{,}120$). Sensitivity already falls at $128 \times 128$: atomization at $\lam = 0.75$ is 0.2\% versus 3.0\% at $64 \times 64$. Behaviour at default MSA depth, and the extent to which MSA reduction itself modulates perturbation sensitivity, remain open questions.

Ubiquitin and close structural relatives are heavily represented in the PDB-derived training corpus, and several experimental reference structures relevant to this system predate AlphaFold2's training cutoff (30 April 2018)~\cite{jumper2021}. A training-exposure explanation for the ubiquitin result is therefore plausible, though the connectivity and ordering of the recovered landscape are harder to attribute to memorization alone (Section~\ref{sec:discussion-spectroscopy}). The three-protein design cannot distinguish these accounts; the proteome-scale screen proposed in Section~\ref{sec:discussion-future} is designed to pressure-test a pure training-exposure account against a broader learned-structure account.

\sgc{} does not produce thermodynamic ensembles. The 33 structures at each condition are deterministic recycling iterates, not independent conformational samples. The observed spread is a path-dependent convergence readout. A different recycle window would produce a different ensemble. We therefore use ``perturbation sensitivity'' as a model diagnostic, not a physical fluctuation measure, and we do not compare against NMR $S^2$ order parameters, which require equilibrium sampling.

The training objective rewards predicting static structures, giving the model every reason to learn \emph{which} residues are flexible but no reason to learn \emph{how much} they move. The contact-loss ordering (Section~\ref{sec:results-unfolding}) reflects differential encoding robustness in the Evoformer weights, not free-energy differences or kinetic barriers. The landscape overlap (Section~\ref{sec:results-topology}) is topological, not Boltzmann-weighted. \sgc{} measures what the weights encode; whether that encoding corresponds to a physical energy function is a question we pose but do not yet answer.

Methodologically, this study probes one architecture family with one perturbation scheme. Most mechanistic analyses use cumulative forward perturbation (blocks $0 \to d$), and several are reported only at the primary operating point ($\sig = 0.30$, $\lam = 0.75$). Single-block contributions and parameter generality beyond the survey grid remain unresolved. The three proteins were chosen as boundary probes, not a representative sample. For $\alpha$-synuclein, the model-specific extensions are structured but unvalidated, and the pooled MD reference is itself force-field-heterogeneous (Supplementary Fig.~\ref{fig:supp_asyn_md_per_ff}). Neural spectroscopy remains an empirical framework. The formal theory connecting perturbation response to representational structure is the subject of ongoing work.

\subsection{Future directions}
\label{sec:discussion-future}

The sharpest test of this paper's central interpretation is a proteome-scale screen. Section~\ref{sec:discussion-representations} proposed that inter-model agreement under \sgc{} should track training-data representation density. Running \sgc{} across a curated protein panel spanning fold class, disorder, and PDB coverage would characterize where inter-model agreement is high, partial, and absent. This is the strongest available test for distinguishing a pure training-exposure account from a broader learned-structure account. Under the first, recovered structure should scale tightly with training density; under the second, structural regularities should extend beyond explicit coverage. Training-data density is itself a proxy, not a clean discriminator, so the screen pressure-tests the hypothesis rather than settling it. A well-designed version of the screen would also calibrate perturbation sensitivity against experimental or simulation references in matched systems, testing whether pattern-level agreement extends to magnitude under controlled conditions.

A second open question is where the conformational information resides. It could be in the Evoformer's learned weights, in the MSA-derived coevolutionary input, or in their interaction. \sgc{} modulates the processing. MSA subsampling modulates the input. Chang and Perez removed MSA and template inputs entirely~\cite{chang2026}. A matched comparison across these regimes, calibrated to comparable structural diversity, would separate the three contributions more directly than any single approach. The sharpest available experiment is weight perturbation applied to ESMFold~\cite{lin2023}, which predicts structure from sequence alone without coevolutionary input. If perturbation of ESMFold's weights produces structured conformational responses, the encoding is already present in sequence-trained weights and the MSA adds specificity rather than generating the landscape. If it does not, the coevolutionary input is load-bearing. AlphaFold3~\cite{abramson2024}, with its diffusion-based architecture and modified attention structure, would test whether the approach generalizes beyond the Evoformer. A finer-grained question exists within the Evoformer itself. Each block contains multiple components: MSA row and column attention, outer-product mean, triangle multiplicative updates, triangle attention, and pair transitions. \sgc{} currently perturbs all of them simultaneously. Component-targeted perturbation could identify which operations carry the conformational signal, whether some are dispensable, and whether selective perturbation yields more stable or more interpretable responses than the uniform scheme used here.

The representational geometry that the Evoformer constructs for each protein is the deeper object of study. We have characterized it along a single axis, in one representation stream, for one protein. Multi-stream analysis across proteins and models is the natural extension. For ubiquitin, where five models converge at Level~3, the representational manifold may be low-dimensional and tightly constrained. For $\alpha$-synuclein, where they diverge, it may admit more solutions. Cross-model representational comparison, through methods such as centered kernel alignment or representational similarity analysis, could test whether two models' manifolds are related by a learnable transformation or are genuinely different objects. The $\alpha$-synuclein landscapes offer the most concrete test. Section~\ref{sec:discussion-speculation} proposed that model-specific landscapes may correspond to identifiable functional contexts. Mapping each model's representational manifold to its structural output and asking whether those correspondences are biologically meaningful would distinguish structured extrapolation from training noise. Either outcome clarifies what the manifold encodes.

The block-2 anomaly, the KaiB depth-7 dip, and the forward/reverse asymmetry suggest that the Evoformer's 48 blocks are not functionally equivalent. Jumper and colleagues trained Structure Module probes at each Evoformer block and showed that intermediate predictions evolve across the 48-block stack~\cite{jumper2021}. Those probes are consistent with a coarse early-to-late progression, but they do not by themselves establish a global-topology-versus-local-geometry division. Per-block and per-head perturbation profiling is therefore the natural next test. It could identify which blocks set coarse fold organization, which refine contacts, and where compensation occurs. The block-2 collapse and block-3 rescue mark one such compensation point. The KaiB depth-7 anomaly suggests another, condition-dependent bottleneck. More broadly, per-head profiling for any protein would reveal what individual attention heads contribute to the conformational encoding. For KaiB, the question is sharpest: do any heads carry sub-dominant signals compatible with the ground-state fold, or was no such signal recovered under the \sgc{} protocol tested here? Selectively perturbing triangle attention while leaving other components intact would test whether the error-correction properties discussed in Section~\ref{sec:discussion-speculation} actively suppress a minority signal or whether no such signal was recoverable under the perturbation scheme used here.

AlphaFold2 is a learned representation of protein folding that warrants study beyond its use as a prediction tool. Decades of PDB depositions and evolutionary sequence alignments appear to have left conformational organization in these 93 million parameters that is not visible in unperturbed predictions. The \sgc{} perturbation reads out stability orderings, flexibility patterns, and landscape topology. The quality of these signals varies with the strength of the training signal. Their pattern is accessible to global perturbations because it lives in dominant directions of the learned representational geometry. That a model trained only on static structures encodes this conformational organization, and that a simple perturbation can read it out with physical correspondence, remains unexplained. The formal connection between weight-space geometry, representational construction, and conformational emergence remains to be established. We call the approach of probing it neural spectroscopy, and expect that alternative perturbation protocols will reveal complementary aspects of the encoding.

\section*{Funding}

This research received no specific grant from any funding agency in the public, commercial, or not-for-profit sectors.

\section*{Competing interests}

The author declares no competing interests.

\section*{Code and data availability}

Source code for the \sgc{} perturbation, the analysis and figure-generation scripts, the YAML run manifests, and the processed metric tables are available from the author on request. Representative predicted structures and the indexed run metadata are provided; the full generated structure corpus is large (approximately two million structures) and is available on request. The molecular dynamics reference trajectories are from D.E.\ Shaw Research and are available from the original authors under their terms.

\clearpage

\setcounter{section}{0}
\setcounter{subsection}{0}
\setcounter{figure}{0}
\setcounter{table}{0}
\setcounter{equation}{0}
\setcounter{algocf}{0}

\renewcommand{\thesection}{S\arabic{section}}
\renewcommand{\thesubsection}{S\arabic{section}.\arabic{subsection}}
\renewcommand{\thefigure}{S\arabic{figure}}
\renewcommand{\thetable}{S\arabic{table}}
\renewcommand{\theequation}{S\arabic{equation}}
\renewcommand{\thealgocf}{S\arabic{algocf}}

\renewcommand{\theHsection}{supp.\arabic{section}}
\renewcommand{\theHfigure}{supp.\arabic{figure}}
\renewcommand{\theHtable}{supp.\arabic{table}}
\renewcommand{\theHequation}{supp.\arabic{equation}}

\fancyhead[L]{\small\sffamily\color{gray} Supplementary Information}

\thispagestyle{plain}
\begin{center}
    {\LARGE\bfseries\sffamily Supplementary Information}

    \vspace{0.6em}

    {\large Neural spectroscopy of AlphaFold2 reveals\\[4pt]
    encoded protein conformational landscapes}

    \vspace{0.6em}

    {\normalsize Kaustav Mehta}
\end{center}

\vspace{1.4em}

\section{Experimental systems}
\label{sec:supp-systems}

\begin{table}[H]
\centering
\caption{\textbf{Protein systems: sequences, MSA depths, reference structures, and experimental data.}
All AlphaFold2 inputs are the canonical full-length sequences listed below the table. MSA depth is reported as $m_\text{seq} \times m_\text{extra}$, where $m_\text{seq}$ is the number of cluster representatives and $m_\text{extra}$ is the number of extra sequences (default is $512 \times 5{,}120$; all experiments reported here use reduced depth, see Section~\ref{sec:methods-impl}).}
\label{tab:supp_systems}
\small
\renewcommand{\arraystretch}{1.2}
\begin{tabular}{p{0.12\textwidth}p{0.08\textwidth}p{0.06\textwidth}p{0.22\textwidth}p{0.37\textwidth}}
\toprule
Protein & UniProt / source & Length (aa) & Reference structure(s) & Experimental / MD data \\
\midrule
Ubiquitin (\emph{H.\ sapiens}) & P0CG48 & 76 & 1UBQ~\cite{vijaykumar1987} & 6$\times$MD at 390~K (6.7~ms total, individual trajectories 0.68--1.24~ms; Piana, Lindorff-Larsen \& Shaw 2013~\cite{piana2013}); 6$\times$equilibrium MD at 300~K spanning 4 AMBER and 2 CHARMM force fields (five trajectories at 8--10.5~$\mu$s each, plus CHARMM22*/TIP3P at 154.6~$\mu$s used as the primary RMSF reference; Robustelli, Piana \& Shaw 2018~\cite{robustelli2018}) \\
\midrule
KaiB (\emph{S.\ elongatus} PCC~7942) & Q79PF5 & 102 & Ground state: 2QKE chain~B (\emph{S.\ elongatus})~\cite{pattanayek2008}; fold-switched state: 5JYT chain~A (\emph{T.\ elongatus} with stabilizing mutations)~\cite{tseng2017} & No MD or NMR ensemble used in this work; fold-switching literature~\cite{tseng2017,chakravarty2022} provides qualitative context \\
\midrule
KaiB (2QKE cognate) & PDB 2QKE seq & 108 & 2QKE chain~B~\cite{pattanayek2008}; 5JYT chain~A~\cite{tseng2017} & Supplementary control only (Fig.~\ref{fig:supp_kaib_2qke}); confirms that main KaiB results do not depend on the \mbox{$\sim$85\%} sequence identity gap between Q79PF5 and the reference structures \\
\midrule
$\alpha$-synuclein (\emph{H.\ sapiens}) & P37840 & 140 & None (intrinsically disordered) & Eight explicit-solvent MD trajectories, $\sim$273~$\mu$s total, spanning six AMBER and two CHARMM force-field variants (a03ws, a99SB-ILDN/TIP4P-D, a99SB-UCB, a99SBdisp, a99SBdisp-extended, a99SB*-ILDN/TIP3P; c22*/TIP3P, CHARMM36m; each $\sim$30~$\mu$s except a99SB-ILDN/TIP4P-D at $\sim$21~$\mu$s and a99SBdisp-extended at $\sim$73~$\mu$s; Robustelli, Piana \& Shaw 2018~\cite{robustelli2018}); no experimental consensus ensemble exists \\
\bottomrule
\end{tabular}
\end{table}

\noindent\textbf{Canonical FASTA sequences used as AlphaFold2 input.}
Sequences below are provided verbatim as submitted to the inference pipeline. Per-residue indexing in the main text uses one-based numbering from the starting methionine.

{\small\ttfamily
\textgreater{}1UBQ $|$ \textit{H.\ sapiens} ubiquitin (P0CG48) $|$ 76 aa\\
MQIFVKTLTGKTITLEVEPSDTIENVKAKIQDKEGIPPDQQRLIFAGKQLEDGRTLSDYN\\
IQKESTLHLVLRLRGG

\vspace{0.4em}
\textgreater{}KaiB $|$ \textit{S.\ elongatus} PCC~7942 (Q79PF5) $|$ 102 aa\\
MSPRKTYILKLYVAGNTPNSVRALKTLKNILEVEFQGVYALKVIDVLKNPQLAEEDKILA\\
TPTLAKVLPLPVRRIIGDLSDREKVLIGLDLLYGELQDSDDF

\vspace{0.4em}
\textgreater{}KaiB\_2QKE $|$ \textit{S.\ elongatus} ground-state cognate $|$ 108 aa\\
MAPLRKTYVLKLYVAGNTPNSVRALKTLNNILEKEFKGVYALKVIDVLKNPQLAEEDKIL\\
ATPTLAKVLPPPVRRIIGDLSNREKVLIGLDLLYEEIGDQAEDDLGLE

\vspace{0.4em}
\textgreater{}$\alpha$-synuclein $|$ \textit{H.\ sapiens} (P37840) $|$ 140 aa\\
MDVFMKGLSKAKEGVVAAAEKTKQGVAEAAGKTKEGVLYVGSKTKEGVVHGVATVAEKTK\\
EQVTNVGGAVVTGVTAVAQKTVEGAGSIAAATGFVKKDQLGKNEEGAPQEGILEDMPVDP\\
DNEAYEMPSEEGYQDYEPEA
\par}

\section{Supplementary Methods}
\label{sec:supp_methods}

\subsection{Noise control construction}

Both noise controls deliver perturbation power matched to the reference \sgc{} condition ($\sig = 0.30$, $\lam = 0.75$). For each Evoformer weight tensor $W$:

\textbf{White noise.} We compute the squared Frobenius norm of the \sgc{} perturbation $\|\Delta W_{\text{SGC}}\|_F^2$ and draw an i.i.d.\ Gaussian tensor $\delta$ of matching shape, with per-element variance set so that the expected squared Frobenius norm $\mathbb{E}\,\|\delta\|_F^2$ equals $\|\Delta W_{\text{SGC}}\|_F^2$ (Algorithm~\ref{alg:white}). Matching is therefore exact in expectation per tensor, not per individual draw. The noise is added to $W$ in place. A dedicated random number generator (independent of the inference RNG) ensures that the noise pattern is determined solely by the noise seed, not by the model's internal state.

\textbf{Spectral noise.} We compute the FFT of $\Delta W_{\text{SGC}}$ along each tensor axis, randomize the phases (replacing each complex phase with a uniform draw on $[0, 2\pi)$), and inverse-FFT to produce a real-valued perturbation with the same power spectrum but destroyed phase structure. This preserves the spectral profile of the original perturbation while eliminating its deterministic relationship to $W$. Note that the FFT assumes periodic boundaries along each tensor axis; since Evoformer weight tensors do not have a natural spatial periodicity, the spectral noise control tests sensitivity to phase structure specifically, not to general spectral properties.

\subsection{Algorithms}
\label{sec:supp-algorithms}

Algorithms~\ref{alg:supp-sgc}, \ref{alg:white}, and~\ref{alg:spectral} give the exact in-place operations on Evoformer weights used in this work. In all three schemes, the Structure Module, recycling embedder, input embedder, and auxiliary heads (pLDDT, pTM, distogram) are untouched.

\begin{spacing}{1.2}
\begin{algorithm}[H]
\DontPrintSemicolon
\SetAlgoLined
\SetInd{0.6em}{1.0em}
\SetKwInOut{Input}{inputs}
\SetKwInOut{Output}{outputs}
\caption{\textsc{SGC}\,---\,Scaled Gaussian Convolution of Evoformer weights}
\label{alg:supp-sgc}
\Input{AlphaFold model $M$ with 48 Evoformer blocks; blur width $\sigma \geq 0$; scale $\lambda > 0$; depth set $L \subseteq \{0, 1, \ldots, 47\}$}
\Output{Model $M$ with perturbed Evoformer weights in place}
\ForEach{block index $i \in L$}{
    \ForEach{parameter tensor $W$ in \emph{M.evoformer.blocks}[$i$]}{
        $\widetilde{W} \leftarrow \mathrm{gaussian\_filter}(W;\, \sigma,\, \mathrm{order}{=}0)$ \tcp*[r]{scipy.ndimage, applied to every tensor axis}
        $W \leftarrow \lambda \cdot \widetilde{W}$ \tcp*[r]{in-place on \emph{param.data}; same dtype and device}
    }
}
\Return $M$\;
\end{algorithm}
\end{spacing}

\vspace{-0.3em}
\noindent\emph{Notes.} (i)~\texttt{gaussian\_filter} treats the tensor as an N-D voxel grid; $\sigma$ is in element units and applies identically to every axis. (ii)~Every named parameter inside an Evoformer block is perturbed: MSA row/column attention projections (q/k/v/o), outer-product mean, triangle attention (start/end), pair transition, and all associated biases. (iii)~For the sweeps reported in the main text, ``cumulative forward'' at depth $k$ corresponds to $L = \{0, 1, \ldots, k-1\}$. (iv)~Original weights are checkpointed and restored between runs so the same model object can be reused across the sweep.

\begin{spacing}{1.2}
\begin{algorithm}[H]
\DontPrintSemicolon
\SetAlgoLined
\SetInd{0.6em}{1.0em}
\SetKwInOut{Input}{inputs}
\SetKwInOut{Output}{outputs}
\caption{\textsc{WhiteNoise}\,---\,per-tensor power-matched Gaussian noise}
\label{alg:white}
\Input{Model $M$; depth set $L$; SGC parameters $\sigma, \lambda$ used to define the reference residual; power multiplier $\kappa \geq 0$ (default $\kappa = 1$); dedicated RNG $\mathcal{R}$}
\Output{Model $M$ with additive Gaussian noise on Evoformer weights}
\ForEach{block index $i \in L$}{
    \ForEach{parameter tensor $W$ in \emph{M.evoformer.blocks}[$i$]}{
        $\Delta W_{\mathrm{SGC}} \leftarrow \lambda \cdot \mathrm{gaussian\_filter}(W;\, \sigma) - W$ \tcp*[r]{the SGC residual that defines reference power}
        $P \leftarrow \sum_{a} \Delta W_{\mathrm{SGC}}[a]^2$ \tcp*[r]{squared Frobenius norm, per tensor}
        \If{$P < 10^{-30}$}{\textbf{continue} \tcp*[r]{skip tensors unaffected by smoothing (e.g.\ scalar biases)}}
        $\sigma_\eta \leftarrow \sqrt{\kappa \cdot P / \mathrm{numel}(W)}$ \tcp*[r]{so $\mathbb{E}\,\lVert\eta\rVert_F^2 = \kappa \cdot P$}
        $\eta \sim \mathcal{N}(0,\, \sigma_\eta^2 \mathbf{I})$ \tcp*[r]{drawn via $\mathcal{R}$; shape of $W$}
        $W \leftarrow W + \eta$ \tcp*[r]{additive on original weights}
    }
}
\Return $M$\;
\end{algorithm}
\end{spacing}

\vspace{-0.3em}
\noindent\emph{Notes.} (i)~Power matching is per-tensor, not global, to prevent large tensors from absorbing the noise budget. (ii)~$\mathcal{R}$ is seeded from \texttt{run["seed"]} and is independent of the MSA-subsampling RNG; this makes the noise pattern deterministic given the seed but decoupled from model internal state. (iii)~Dosage variants (Supplementary Fig.~\ref{fig:supp_noise_dosage}) correspond to $\kappa \in \{0.5,\, 1.0,\, 2.0\}$.

\begin{spacing}{1.2}
\begin{algorithm}[H]
\DontPrintSemicolon
\SetAlgoLined
\SetInd{0.6em}{1.0em}
\SetKwInOut{Input}{inputs}
\SetKwInOut{Output}{outputs}
\caption{\textsc{SpectralNoise}\,---\,phase-randomized perturbation with preserved magnitude spectrum}
\label{alg:spectral}
\Input{Model $M$; depth set $L$; SGC parameters $\sigma, \lambda$; dedicated RNG $\mathcal{R}$}
\Output{Model $M$ with phase-randomized additive perturbation on Evoformer weights}
\ForEach{block index $i \in L$}{
    \ForEach{parameter tensor $W$ in \emph{M.evoformer.blocks}[$i$]}{
        $\Delta W_{\mathrm{SGC}} \leftarrow \lambda \cdot \mathrm{gaussian\_filter}(W;\, \sigma) - W$\;
        $\widehat{\Delta} \leftarrow \mathrm{rfftn}(\Delta W_{\mathrm{SGC}})$ \tcp*[r]{real-input FFT; halves the last axis via Hermitian symmetry}
        \If{$\max |\widehat{\Delta}| < 10^{-15}$}{\textbf{continue}}
        $A \leftarrow |\widehat{\Delta}|$ \tcp*[r]{magnitude spectrum, preserved}
        $\theta \sim \mathrm{Uniform}(0,\, 2\pi)$ \tcp*[r]{independent phases for every half-spectrum coefficient, via $\mathcal{R}$}
        $\widehat{\Delta}^{\prime} \leftarrow A \odot \exp(i\,\theta)$ \tcp*[r]{element-wise}
        $\Delta^{\prime} \leftarrow \mathrm{irfftn}(\widehat{\Delta}^{\prime};\, \mathrm{out\_shape} = \mathrm{shape}(\Delta W_{\mathrm{SGC}}))$ \tcp*[r]{real-valued by construction of \texttt{irfftn}}
        $W \leftarrow W + \Delta^{\prime}$\;
    }
}
\Return $M$\;
\end{algorithm}
\end{spacing}

\vspace{-0.3em}
\noindent\emph{Notes.} (i)~Parseval's theorem guarantees $\|\Delta^{\prime}\|_F = \|\Delta W_{\mathrm{SGC}}\|_F$, so no explicit power-matching factor is required. (ii)~Using \texttt{numpy.fft.rfftn}/\texttt{irfftn} exploits Hermitian symmetry: phases for the mirrored half of the spectrum are reconstructed implicitly, and the output is guaranteed real. (iii)~The DC component (k = 0) is randomized along with the rest of the spectrum; in practice the residual's DC coefficient is negligible relative to the bulk of the power.

\paragraph{Determinism and seeding.}
\sgc{} (Algorithm~\ref{alg:supp-sgc}) is fully deterministic given $(\sigma, \lambda, L)$. Both noise controls consume a dedicated \texttt{numpy.random.Generator} seeded from \texttt{run["seed"]}, which is the same seed used for MSA subsampling in the inference run but obtained from an independent RNG object. A single weight checkpoint is written before perturbation and restored after each run, so perturbation, restoration, and re-perturbation across depths share the exact starting weights.

\subsection{Experiment orchestration}
\label{sec:supp-orchestration}

\begin{figure}[H]
\centering
\resizebox{\textwidth}{!}{%
\begin{tikzpicture}[
    node distance=0.9cm,
    stage/.style={rectangle, draw=black!55, fill=black!3,
                  rounded corners=3pt, line width=0.5pt,
                  minimum height=1.35cm, minimum width=2.55cm,
                  align=center, font=\small},
    emph/.style={rectangle, draw=black!70, fill=black!8,
                 rounded corners=3pt, line width=0.6pt,
                 minimum height=1.35cm, minimum width=2.8cm,
                 align=center, font=\small},
    fanout/.style={draw=none, align=center, font=\footnotesize\itshape, text=black!60},
    arrow/.style={-{Latex[length=3mm,width=2mm]}, line width=0.55pt, black!55},
    dashedarrow/.style={-{Latex[length=3mm,width=2mm]}, dashed, line width=0.5pt, black!45},
]
\node[stage] (plan) {YAML\\plan};
\node[stage, right=of plan] (expand) {\texttt{expand\_plan}\\grid $\times$ seeds};
\node[emph, right=of expand] (runs) {$N$ inference runs\\{\footnotesize checkpoint\,/\,restore}};
\node[stage, right=of runs] (idx) {per-run dir\,+\\\texttt{index.jsonl}};
\node[stage, right=of idx] (ana) {analysis\\scripts};

\draw[arrow] (plan) -- (expand);
\draw[arrow] (expand) -- (runs);
\draw[arrow] (runs) -- (idx);
\draw[arrow] (idx) -- (ana);

\node[fanout, below=0.15cm of expand] {seeds $\times$ $\sigma$ $\times$ $\lambda$ $\times$ layers};
\node[fanout, below=0.15cm of idx] {NDJSON\,+\,33 PDBs per run};

\draw[dashedarrow]
    (idx.north) to[out=90, in=90, looseness=1.2]
    node[pos=0.5, above=1pt, font=\scriptsize\itshape, text=black!55] {resume: existing runs filter planned combos}
    (expand.north);
\end{tikzpicture}%
}
\caption{\textbf{Sweep orchestration pipeline.} A single YAML plan is expanded into $N$ runs via Cartesian product over seeds, $(\sigma, \lambda)$, and layer sets. Each run writes a per-run directory containing 33 PDB structures plus metrics and provenance files, and appends one NDJSON record to \texttt{index.jsonl}. All downstream analysis consumes the index directly. On resume (dashed arrow), the driver consults existing per-run directories and removes already-completed combos from the expanded plan before inference, making interrupted sweeps safe to restart.}
\label{fig:supp_orchestration}
\end{figure}
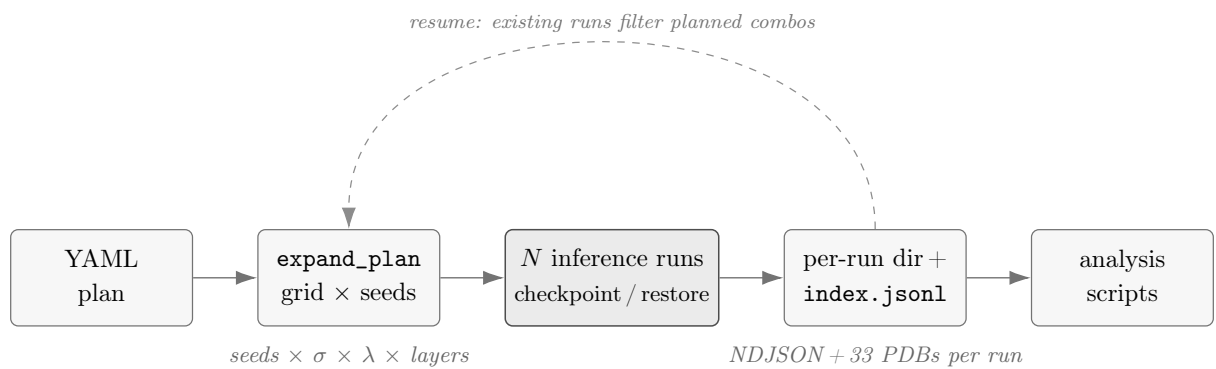

Each parameter sweep is defined by a single YAML plan file that specifies the target protein (FASTA and MSA paths), the OpenFold config and parameter archive, the number of recycles, the MSA-depth variants to test, and a list of sweep entries. Each sweep entry names a perturbation type (\texttt{gaussian\_smooth}, \texttt{noise\_white}, or \texttt{noise\_spectral}), scalar or list-valued $\sigma$ and $\lambda$, a perturbation scheme, and the seeds to use. At runtime, the driver expands every combination of (seed, MSA depth, $\sigma$, $\lambda$, depth-specific layer set) via Cartesian product. Figure~\ref{fig:supp_orchestration} summarizes the pipeline.

\paragraph{Schemes.} The layer selection scheme determines which Evoformer blocks are perturbed:
\begin{itemize}
\setlength{\itemsep}{0.15em}
    \item \texttt{baseline}: no perturbation (empty layer set).
    \item \texttt{single\_layer}: 48 runs, each perturbing exactly one block in turn.
    \item \texttt{cumulative\_forward}: 48 runs perturbing the first $k$ blocks ($k = 1, \ldots, 48$). This is the primary scheme used throughout the main text.
    \item \texttt{cumulative\_reverse}: 48 runs perturbing the last $k$ blocks.
    \item \texttt{explicit}: a single run applying a user-specified layer list from the YAML.
\end{itemize}

\paragraph{Per-run outputs.} Each run writes its outputs to a structured path keyed by protein, model config, parameter archive, seed, scheme, and layer set, under a shared results root. Each leaf directory contains: 33 PDB structures (\texttt{raw/recycle\_000.pdb} through \texttt{raw/recycle\_032.pdb}), per-recycle confidence metrics (\texttt{metrics.csv}), and a \texttt{provenance.json} recording the model config, plan SHA-256, input-file hashes, perturbation parameters, seed, and wall-clock duration. On completion, one NDJSON record is appended to the canonical index, \texttt{results/index.jsonl}, with fields: \texttt{protein}, \texttt{model\_config}, \texttt{perturbation\_type}, \texttt{sigma}, \texttt{scale}, \texttt{msa\_clusters}, \texttt{extra\_msa}, \texttt{seed}, \texttt{scheme}, \texttt{layers}, \texttt{num\_recycles}, \texttt{path}, \texttt{status}, \texttt{mean\_plddt\_final}, \texttt{ptm\_final}, \texttt{has\_minimized}, \texttt{duration\_s}, and \texttt{timestamp} (plus \texttt{power\_scale} for white-noise dosage runs). All downstream analysis in the paper consumes this index directly.

\paragraph{Resume safety.} The driver is idempotent. Before executing any run, it checks for the presence of \texttt{provenance.json} and the first and last recycle PDBs in the target directory; if all three exist, the run is silently skipped. Interrupted sweeps can therefore be restarted without loss or duplication, and completed results from prior sessions are never recomputed.

\paragraph{Monitoring.} A lightweight status reader walks the results tree and infers per-run state from filesystem artefacts (no database), reporting per-protein completion rates and, where applicable, per-force-field minimization counts (CHARMM36, AMBER14SB) alongside the inference runs.

\subsection{Pair-activation magnitude PCA}

For each captured pair tensor $z_{ij} \in \mathbb{R}^{128}$ (one per residue pair per Evoformer block per recycle), we compute the scalar magnitude $\operatorname{asinh}(\|z_{ij}\|_2)$. The $\operatorname{asinh}$ transform compresses the heavy-tailed distribution of pair norms while preserving zero and avoiding the singularity of $\log$ at zero. We retain only long-range pairs ($|i - j| \geq 6$), excluding trivial sequential neighbors, yielding a 4{,}970-dimensional feature vector per snapshot.

\textbf{Stage residualization.} To isolate condition-specific variation from the shared Evoformer processing trajectory, we subtract the cross-condition mean at each exact computational stage. For each (seed $s$, recycle $r$, block $b$) triple, the residual is:
\[
    x'_{c,s,r,b} = x_{c,s,r,b} - \frac{1}{|\mathcal{C}|} \sum_{c' \in \mathcal{C}} x_{c',s,r,b}
\]
where $\mathcal{C}$ is the set of structured conditions (excluding noise controls). Each seed is residualized independently, preserving inter-seed variation as a measure of reproducibility.

\textbf{PCA fitting and projection.} PCA is computed via the truncated singular value decomposition (retaining only the top 10 components) and fit on the residualized structured conditions only ($|\mathcal{C}| = 9$ conditions $\times$ 3 seeds = 42{,}768 samples). Noise controls are residualized using the same structured stage means and projected into the fitted PCA subspace. The reconstruction residual $\|x'_{\text{oos}} - \hat{x}'_{\text{oos}}\|_2$ measures how far each out-of-sample point lies from the 10-component structured subspace.

\textbf{Panel C interpretation.} Error bars on structured condition centroids represent the standard deviation of per-seed centroids (3 seeds = 3 independent replicates, where ``seed'' controls MSA subsampling). For noise controls, each seed is plotted as an individual point rather than as a mean $\pm$ s.d., because with $n = 3$ and visibly non-Gaussian spread, error bars would be misleading.

\textbf{Panel D interpretation.} Boxplots show the distribution of reconstruction residuals across all (seed, recycle, block) snapshots for each condition. For structured conditions these are in-sample residuals; for noise controls they are out-of-sample. Both are in the same feature-space units ($\operatorname{asinh}$ pair-norm distance).

\section{RMSF Validation: Supplementary Figures}

\begin{figure}[H]
    \centering
    \includegraphics[width=\textwidth]{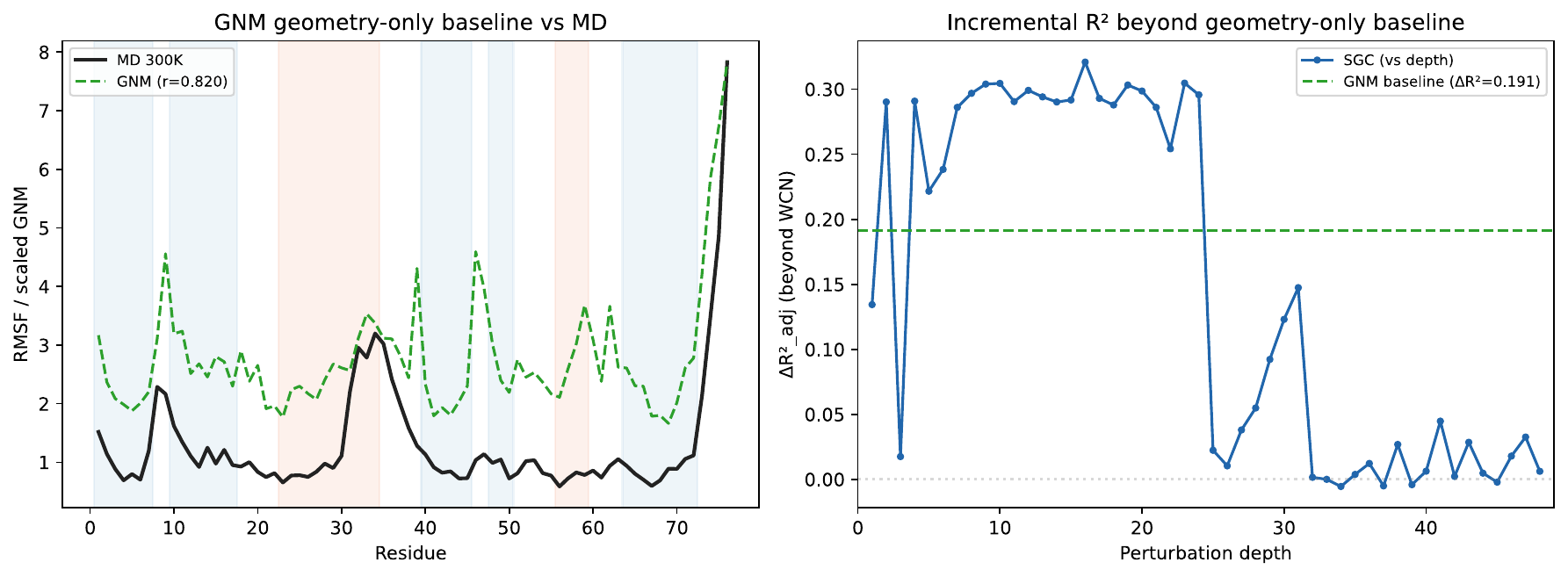}
    \caption{\textbf{Geometry-only baseline: Gaussian Network Model.}
    \textbf{(Left)}~Per-residue GNM fluctuations (20 slowest modes, cutoff 7.3~\AA{}) scaled to MD RMSF range, compared to 300~K CHARMM22*/TIP3P equilibrium RMSF. GNM achieves Pearson~$r = 0.82$ with MD, capturing the broad loop-vs-core pattern from native geometry alone.
    \textbf{(Right)}~Incremental explained variance ($\Delta R^2_\text{adj}$) beyond a WCN-only baseline model. \sgc{} perturbation sensitivity (blue) exceeds the GNM geometry-only baseline (green dashed line) across perturbation depths 4--24, demonstrating that the Evoformer encodes flexibility information beyond what native-state geometry provides. The GNM baseline was computed using ProDy~\cite{bakan2011prody}.}
    \label{fig:supp_gnm}
\end{figure}

\begin{figure}[H]
    \centering
    \includegraphics[width=0.85\textwidth]{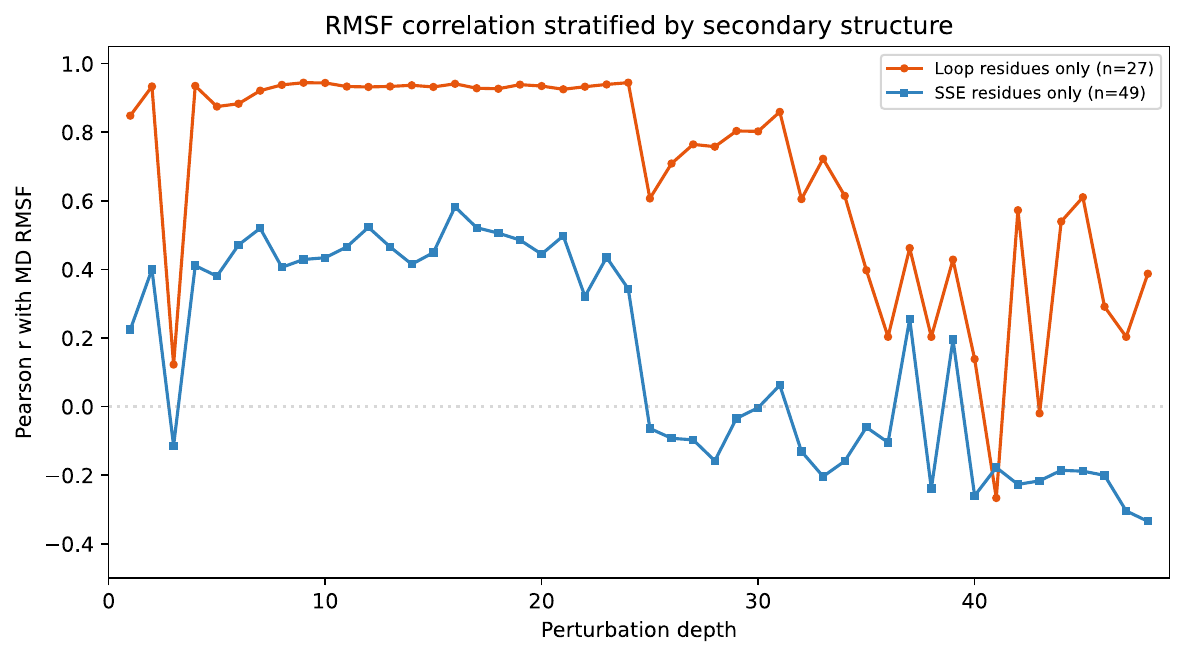}
    \caption{\textbf{RMSF correlation stratified by secondary structure.}
    Pearson~$r$ between \sgc{} perturbation sensitivity and MD RMSF computed separately for loop residues (orange, $n = 27$) and secondary structure element residues (blue, $n = 49$). At shallow depths (4--24), loop-only correlation reaches $r = 0.95$, confirming that \sgc{} resolves \emph{which} loops are more flexible than others---not merely that loops move more than the core. SSE-only correlation is lower ($r \approx 0.2$--$0.6$), consistent with less dynamic range among rigid elements.
    MD reference: CHARMM22*/TIP3P, 300~K equilibrium. \sgc{} condition: $\sig = 0.30$, $\lam = 0.75$, 3 seeds.}
    \label{fig:supp_loop_vs_sse}
\end{figure}

\begin{figure}[H]
    \centering
    \includegraphics[width=\textwidth]{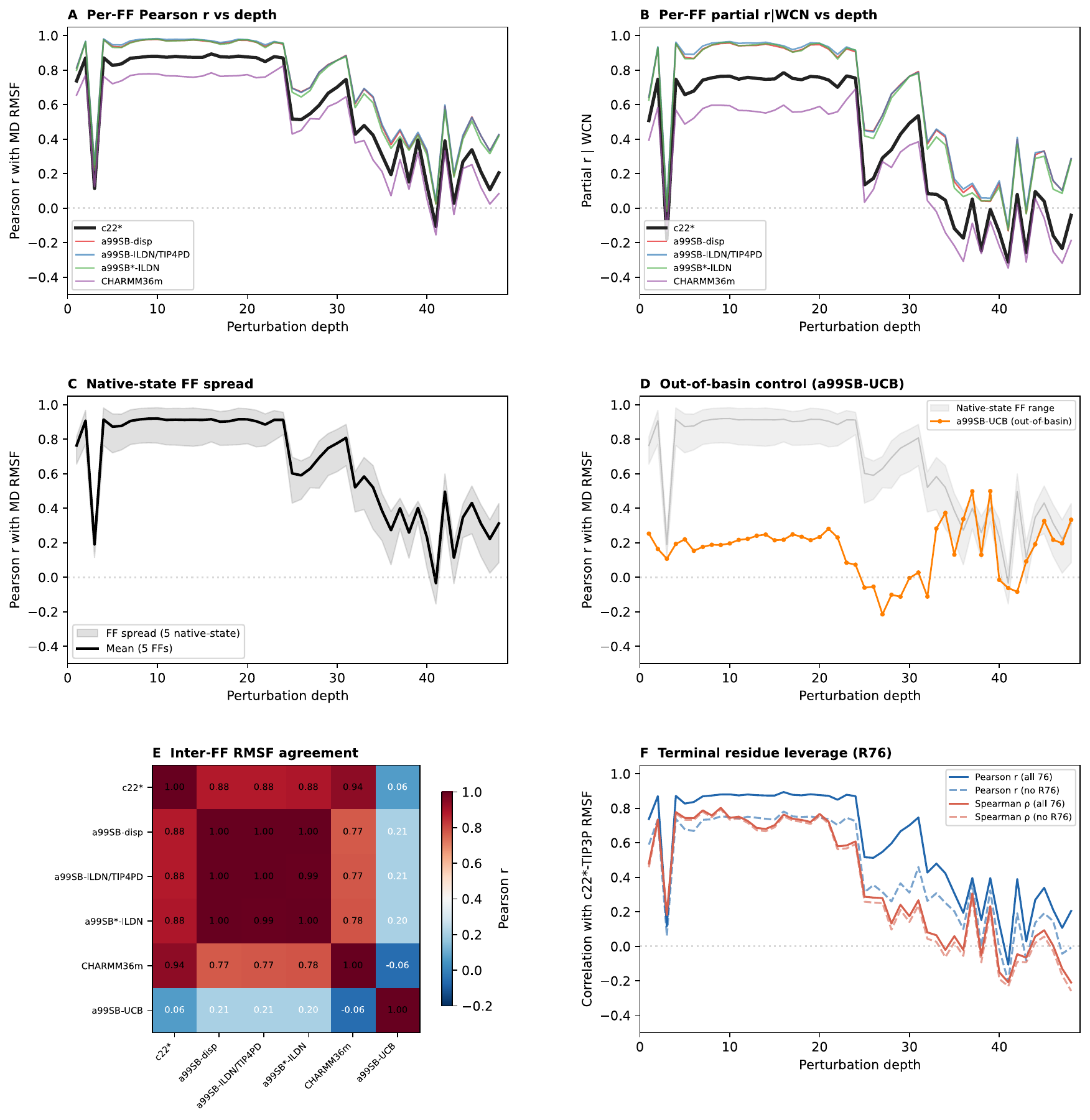}
    \caption{\textbf{Cross-force-field robustness of perturbation sensitivity correlation.}
    \textbf{(A)}~Pearson~$r$ between \sgc{} perturbation sensitivity and MD RMSF at each perturbation depth, computed against five native-state 300~K equilibrium force fields from Robustelli et al.~\cite{robustelli2018}. CHARMM22*/TIP3P (black, primary reference) is emphasized; all five show the same depth-dependent pattern.
    \textbf{(B)}~Partial~$r$\textbar WCN (non-trivial correlation after controlling for burial) across the same five force fields, confirming that the agreement reflects learned flexibility information, not protein architecture.
    \textbf{(C)}~Cross-FF summary: mean Pearson~$r$ (black line) and min--max range (grey band) across all five native-state references.
    \textbf{(D)}~Out-of-basin control: a99SB-UCB (orange), in which ubiquitin unfolds during the equilibrium simulation (mean RMSD $9.9 \pm 8.8$~\AA{}), shows weak, unsystematic correlation that fluctuates around zero at most depths. Grey envelope shows the native-state FF range for comparison.
    \textbf{(E)}~Pairwise Pearson~$r$ between MD RMSF profiles across all six force fields. The five native-state FFs cluster tightly ($r = 0.71$--$0.95$); the three AMBER variants are near-identical ($r > 0.99$). a99SB-UCB is the clear outlier ($r = -0.09$ to $0.18$).
    \textbf{(F)}~Terminal residue leverage analysis. Removing residue~76 (C-terminus) reduces Pearson~$r$ appreciably (dashed blue vs solid blue), especially at deeper depths, but Spearman~$\rho$ (red) is nearly unchanged. The rank-order flexibility pattern is not driven by the terminal residue.
    All panels use $\sig = 0.30$, $\lam = 0.75$, MSA $64 \times 64$, 3 seeds.}
    \label{fig:supp_multi_ff}
\end{figure}

\section{Ubiquitin landscape: C$\alpha$ analysis}

\begin{figure}[H]
    \centering
    \includegraphics[width=\textwidth]{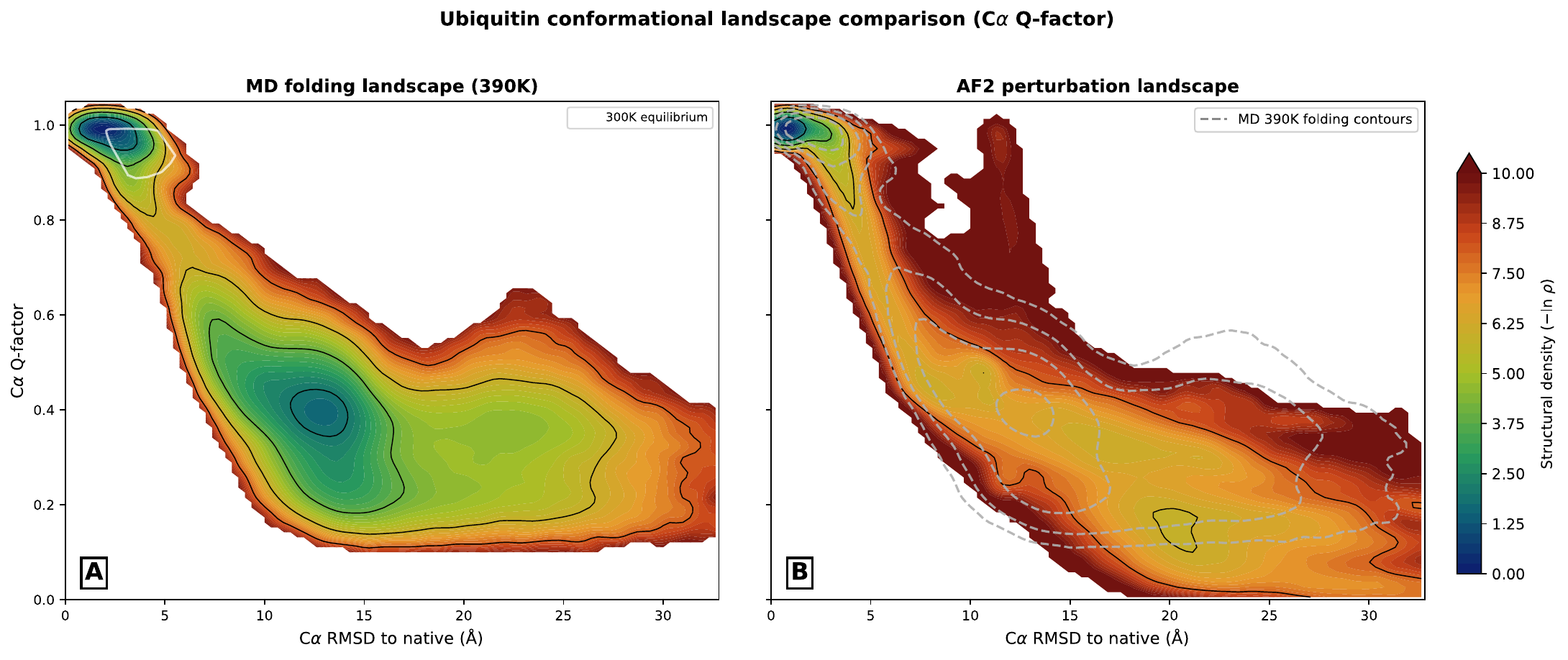}
    \caption{\textbf{Ubiquitin conformational landscape using C$\alpha$-only contacts.}
    Companion to Figure~7 of the main text, with $Q$-factor recomputed from C$\alpha$-only native contacts (12~\AA{} cutoff, sequence separation $\geq 3$) rather than heavy-atom BHE. \textbf{(A)}~Structural density landscape from 673{,}519 MD frames at 390~K (Piana et al.~\cite{piana2013}) plotted as $-\ln\rho$ on C$\alpha$ RMSD $\times$ C$\alpha$ $Q$-factor axes; white outline marks the 300~K equilibrium region (82{,}280 frames, CHARMM22*/TIP3P + CHARMM36m~\cite{robustelli2018}). \textbf{(B)}~AF2 perturbation landscape (912{,}384 frames, model\_1\_ptm, full $\sig$/$\lam$ grid). Dashed magenta contours reproduce the MD 390~K folding contours for direct comparison. Under the C$\alpha$-only definition, both AF2 and MD reach $Q = 1.0$ at the native state (the heavy-atom offset at $Q = 0.806$ reflects imprecise sidechain placement rather than missing backbone contacts). Both landscapes share the same L-shaped topology; kNN territory coverage is 65.0\%, stable across $k = 5$--$80$. This confirms that the topological agreement reported in the main text is robust to the choice of contact definition.}
    \label{fig:supp_landscape_ca}
\end{figure}

\section{KaiB: per-model and cognate-sequence landscapes}

\begin{figure}[H]
    \centering
    \includegraphics[width=\textwidth]{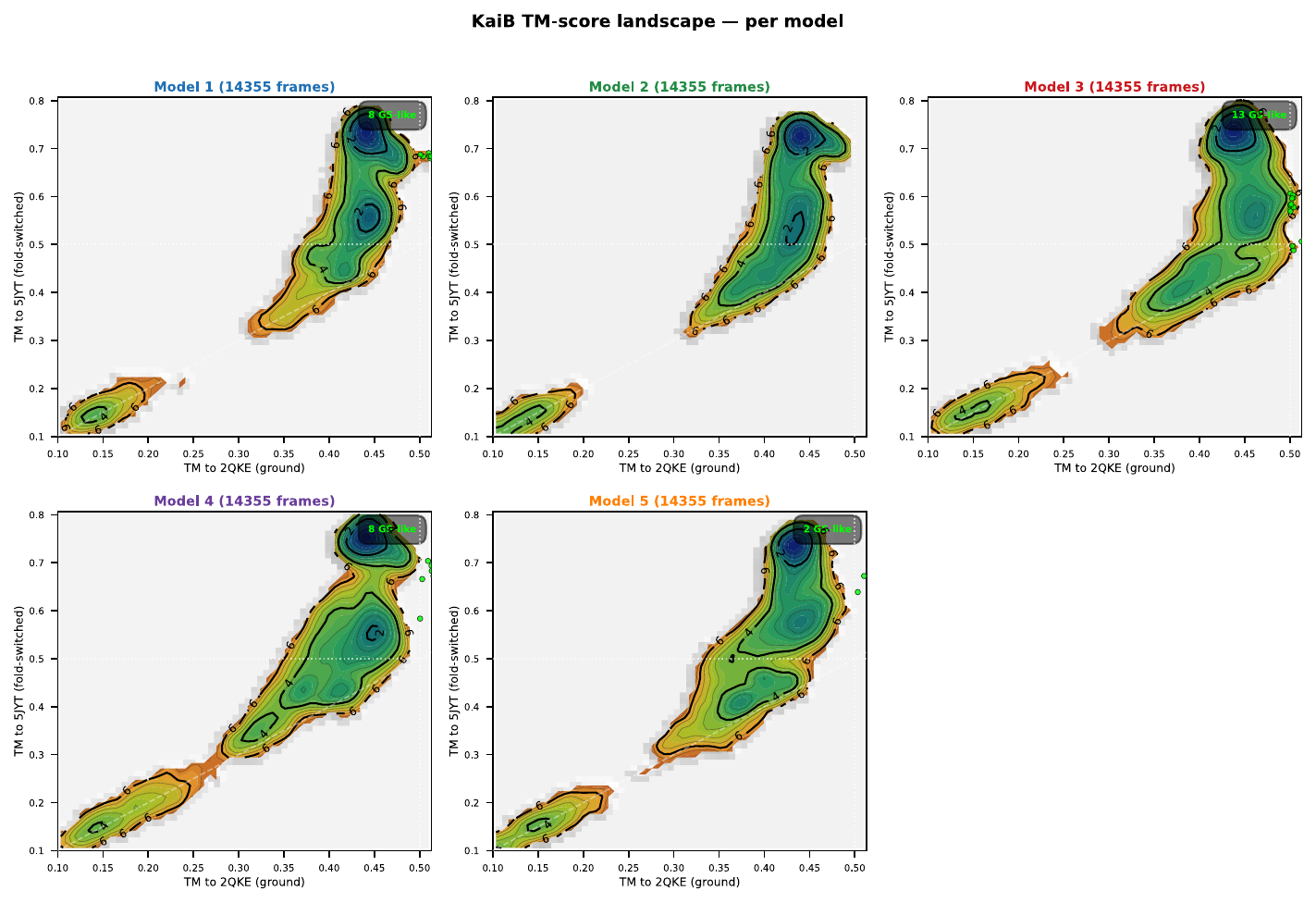}
    \caption{\textbf{KaiB per-model TM-score landscapes.}
    Individual landscapes for AlphaFold2 models 1--5, plotted as TM$_\text{gs}$ (ground state, PDB~2QKE chain~B) vs.\ TM$_\text{fs}$ (fold-switched, PDB~5JYT chain~A). All five models produce the same one-dimensional denaturation axis: points arrange along a diagonal where losing similarity to the fold-switched state also reduces similarity to the ground state, with no orthogonal branch toward TM$_\text{gs} > 0.5$. Only 31 of 71{,}775 frames (0.04\%) cross TM$_\text{gs} > 0.5$; those are shallow-depth threshold fluctuations (models 1, 4, 5) or moderate-depth molten intermediates (model 3) equidistant from both folds rather than ground-state-like structures. Model 2 never crosses the threshold at any depth. Each panel pools three perturbation conditions ($\sig = 0.30, \lam = 0.75$; $\sig = 0.30, \lam = 0.85$; $\sig = 0.20, \lam = 0.75$), all 48 perturbation depths, and 3 seeds. This figure supplements the cross-model pooled landscape shown as Figure~14 in the main text; the convergent absence of fold-switching across five independently trained models is the central finding of Section~\ref{sec:results-kaib}.}
    \label{fig:supp_kaib_per_model}
\end{figure}

\begin{figure}[H]
    \centering
    \includegraphics[width=\textwidth]{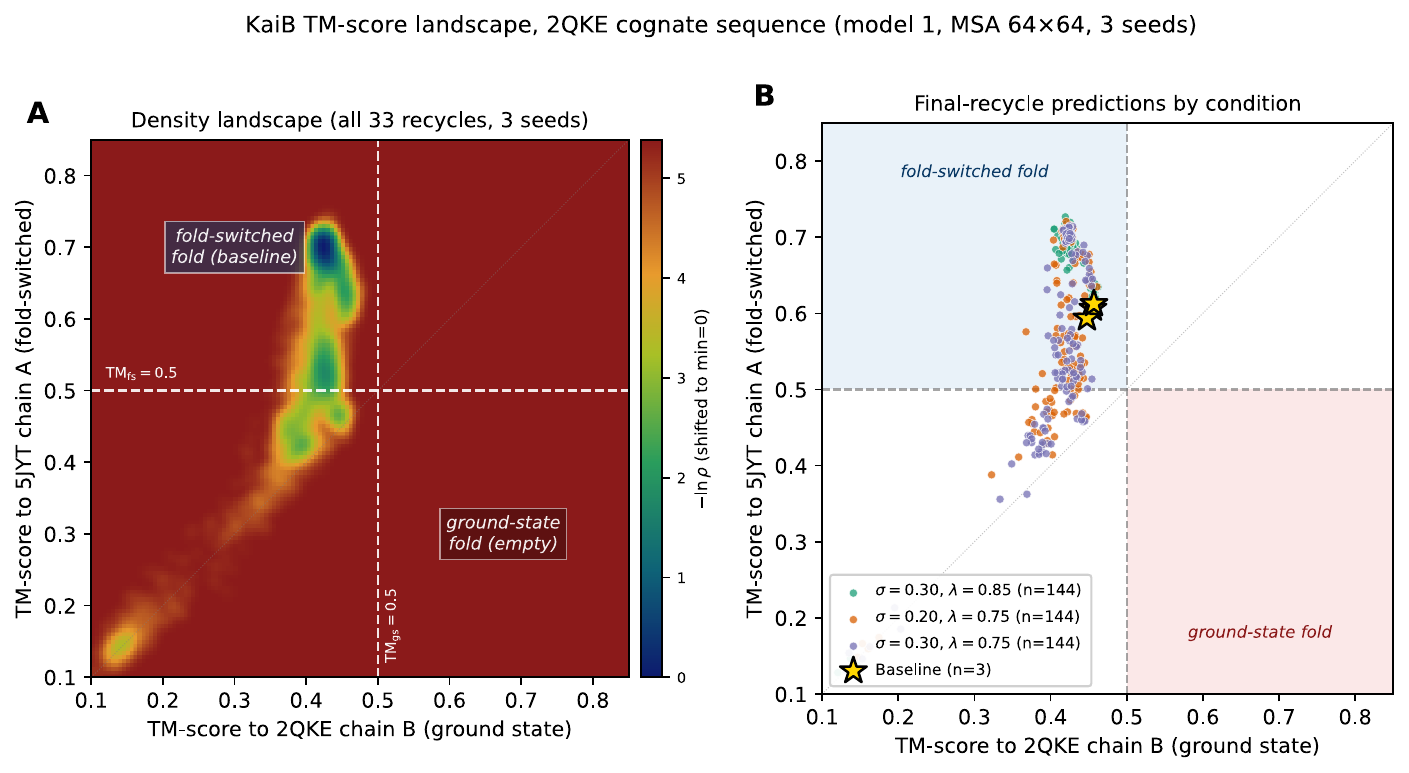}
    \caption{\textbf{KaiB TM-score landscape under the cognate 2QKE sequence.}
    Companion to main-text Section~\ref{sec:results-kaib} (footnote). The default AF2 input used throughout the main KaiB analysis is the \emph{S.\ elongatus} PCC~7942 sequence (UniProt Q79PF5), which is \mbox{$\sim$85\%} identical to both reference structures (2QKE from \emph{S.\ elongatus}, 5JYT from \emph{T.\ elongatus} with fold-switch-stabilizing mutations~\cite{tseng2017,pattanayek2008}). To eliminate the homolog gap between AF2 input and ground-state reference, we repeated the experiment using the 2QKE chain~B sequence directly as AF2 input (108~aa).
    \textbf{(A)}~Density landscape pooling all 33 recycling iterations across 3 seeds and 3 perturbation conditions (14{,}355 frames, model~1 only). Density is concentrated in the fold-switched quadrant (TM$_\text{fs} > 0.5$) and along the denaturation diagonal; the ground-state quadrant (TM$_\text{gs} > 0.5$) is empty at all depths.
    \textbf{(B)}~Final-recycle predictions by perturbation condition (n=435 final-recycle frames). All three conditions --- ranging from mild ($\sig = 0.30$, $\lam = 0.85$) to strong ($\sig = 0.30$, $\lam = 0.75$) --- produce structures that migrate away from the fold-switched baseline along the shared denaturation axis. No frame crosses TM$_\text{gs} > 0.5$; the maximum TM$_\text{gs}$ observed across all 14{,}355 frames is 0.467. The gold star marks the unperturbed baseline prediction (TM$_\text{gs} \approx 0.44$, TM$_\text{fs} \approx 0.72$). The diagonal dotted line shows the reference cross-TM(2QKE, 5JYT) axis; crossing it would indicate motion from one fold basin to the other. None of the perturbed predictions do.
    Taken together, this figure confirms that the convergent absence of fold-switching reported in Section~\ref{sec:results-kaib} is not an artifact of the \mbox{$\sim$15\%} sequence difference between Q79PF5 and the 2QKE reference. This cognate-sequence run uses model~1 only; it serves as a control for the sequence-gap concern rather than a full five-model replication.}
    \label{fig:supp_kaib_2qke}
\end{figure}

\clearpage
\begin{figure}[H]
    \centering
    \includegraphics[width=0.62\textwidth]{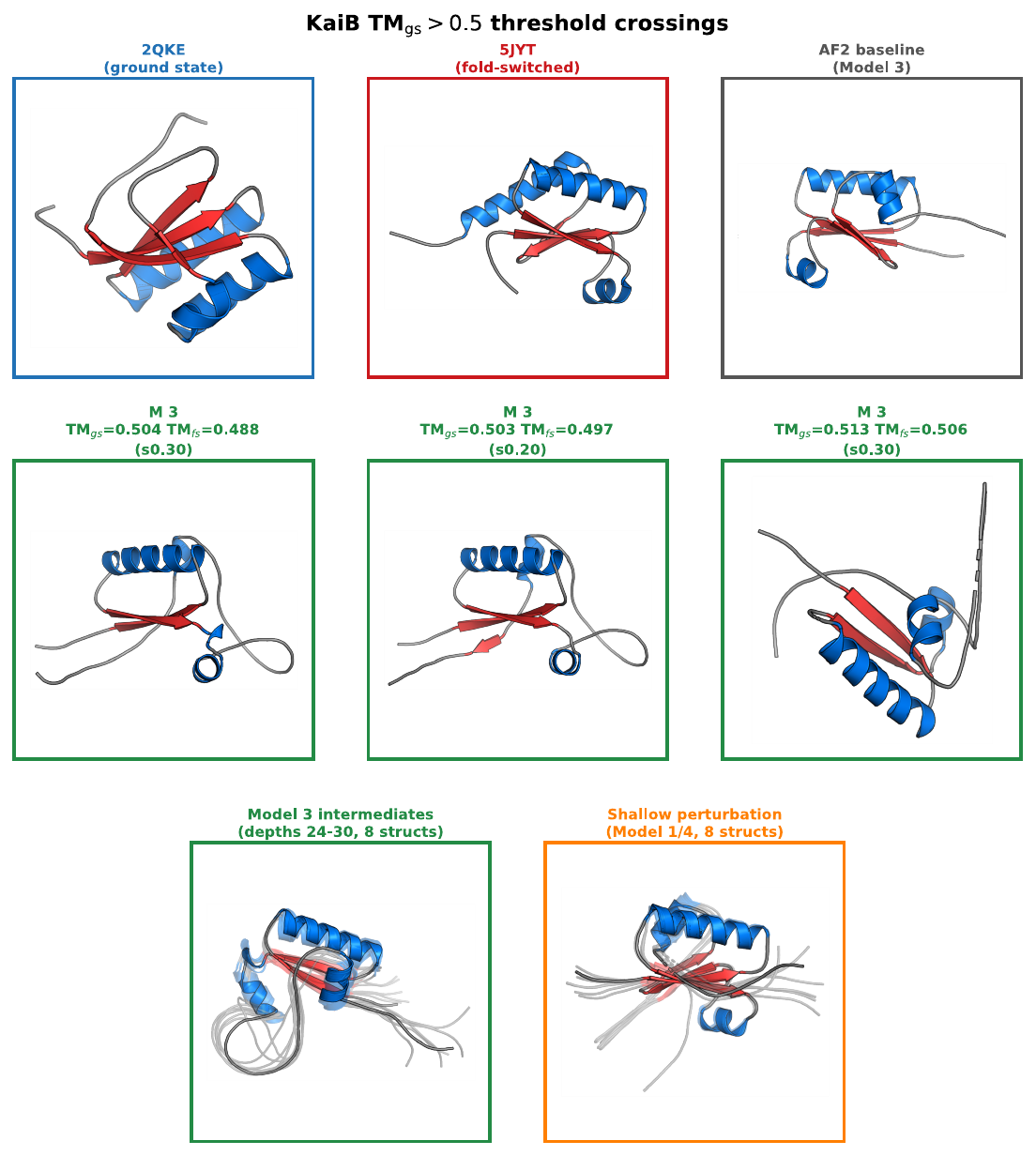}
    \caption{\textbf{Structural gallery of the 31 KaiB frames that cross TM$_\text{gs} > 0.5$ (threshold crossings, not ground-state-like structures).}
    Section~\ref{sec:results-kaib} reports that only 31 of 71{,}280 perturbed frames (and zero of 495 unperturbed-baseline frames) exceed the TM$_\text{gs} > 0.5$ threshold, and argues that these are not ground-state-like structures. They divide into two classes: shallow threshold crossings that remain fold-switched-like, and a small subset of model-3 structurally intermediate crossings that approach TM$_\text{gs} \approx$ TM$_\text{fs}$.
    \textbf{Top row:} the two reference structures (2QKE chain~B, ground state, blue border; 5JYT chain~A, fold-switched, red border; same colour convention as main Fig.~13) alongside the unperturbed AF2 baseline for model~3 (grey border). The baseline reproduces the 5JYT topology, three $\alpha$-helices packed against a three-stranded $\beta$-sheet with the characteristic fold-switched $\beta$-hairpin, consistent with the documented AF2 bias toward the fold-switched fold~\cite{chakravarty2022}.
    \textbf{Middle row:} the three crossings with the smallest $|\mathrm{TM}_\text{gs} - \mathrm{TM}_\text{fs}|$ in the sweep ($\mathrm{TM}_\text{gs} \approx 0.50$, $\mathrm{TM}_\text{fs} \approx 0.49$--$0.51$), all from model~3 (two depth-24 frames, one depth-5 outlier). In each, helical content is preserved but reoriented, and the $\beta$-sheet is partially disassembled relative to the AF2 baseline (Supplementary Table~\ref{tab:supp_ss}).
    \textbf{Bottom row:} superimposed cartoons of illustrative subsets. \emph{Left:} 8 model-3 crossings from depths 24--30 (first opaque, remaining 7 semi-transparent). \emph{Right:} 8 shallow crossings from models 1 and 4 at depths 1--5 (model~5 omitted for space); these cross the threshold not by adopting the ground-state fold but by staying close enough to the 5JYT baseline that threshold fluctuations pull them over the cutoff.
    DSSP analysis of all 31 PDBs (Supplementary Table~\ref{tab:supp_ss}) gives mean helix content $29 \pm 2$\% pooled (vs.\ 38\% in the 5JYT reference) and mean strand content $19 \pm 5$\% pooled, with $15 \pm 3$\% strand in the model-3 near-balanced subset. Pooled strand content does not exceed the AF2 baseline (25\%) and the near-balanced subset does not exceed the 5JYT strand fraction (22\%). DSSP counts secondary structure rather than sheet arrangement, so the complementary visual check is the absence of the straightened $\beta$3 strand and the ordered $\beta$-hairpin in the cartoons. Taken with the TM-score landscapes in main Fig.~13 and the per-model panels in Supplementary Fig.~\ref{fig:supp_kaib_per_model}, these crossings are consistent with a sliding cutoff on the same denaturation axis rather than a distinct fold-switch branch.}
    \label{fig:supp_kaib_gallery}
\end{figure}

\section{Evoformer representational divergence}

\begin{figure}[H]
    \centering
    \includegraphics[width=\textwidth]{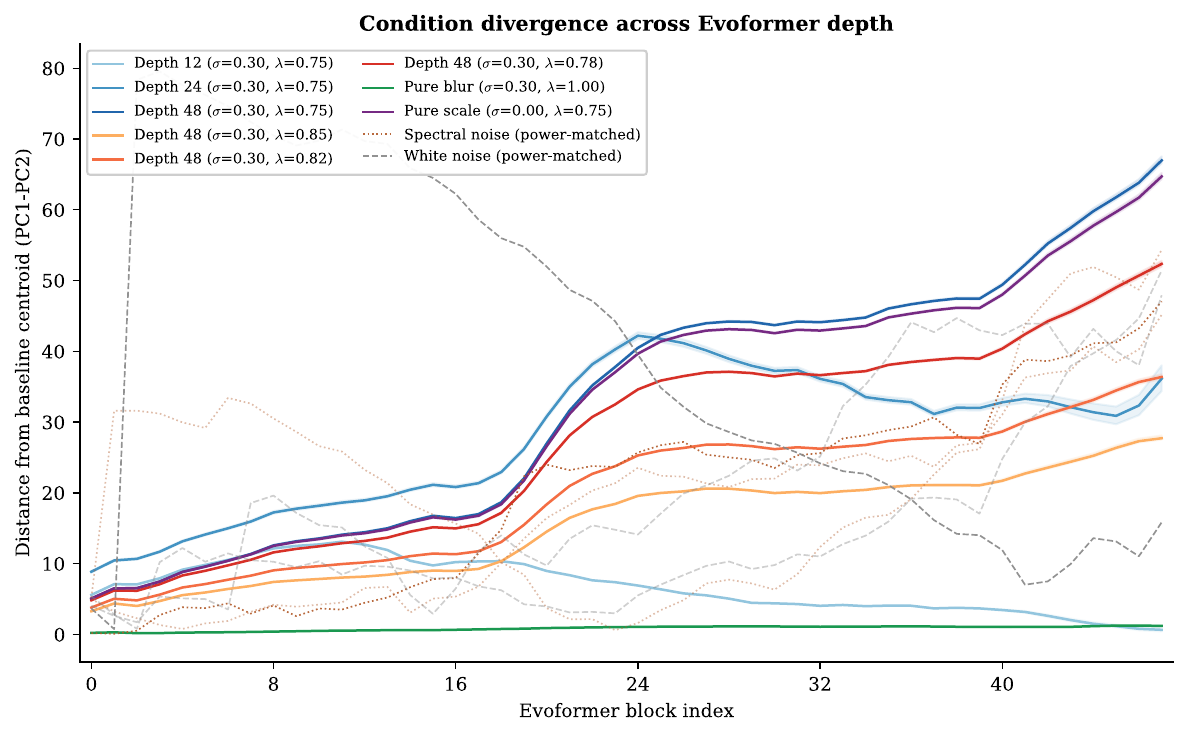}
    \caption{\textbf{Condition-specific representational divergence builds through Evoformer depth.}
    Per-seed Euclidean distance from each condition's centroid to the matched baseline centroid in residualized PC1--PC2 space, as a function of Evoformer block index. Structured conditions (solid lines with seed-range ribbons) show smooth, monotonic divergence that tracks perturbation depth. Noise controls (dashed/dotted: individual seed traces) show erratic, seed-dependent trajectories---different seeds of the same noise type follow different paths, consistent with the stochastic structural behavior reported in Section~\ref{sec:results-noise}. The depth-12 condition (cyan) reconverges after its 12 perturbed blocks, demonstrating partial reversibility of the representational effect.}
    \label{fig:supp_divergence}
\end{figure}

\section{Chain Integrity: Structural Gallery}
\label{sec:supp-chain-integrity}


The following four pages show backbone stick representations (N, CA, C, O atoms) rendered in PyMOL for 48 structures at matched perturbation power. Columns show the recycling progression (recycles 8, 16, 24, 32); rows within each page show three independent MSA seeds. Blue indicates intact backbone (\mbox{CA--CA} $\leq$ 5~\AA{}); red indicates broken or atomized chains (\mbox{CA--CA} $>$ 5~\AA{}). Intact structures are aligned to the native 1UBQ crystal structure; broken structures are shown in their own orientation.

\begin{figure}[H]
    \centering
    \includegraphics[width=\textwidth]{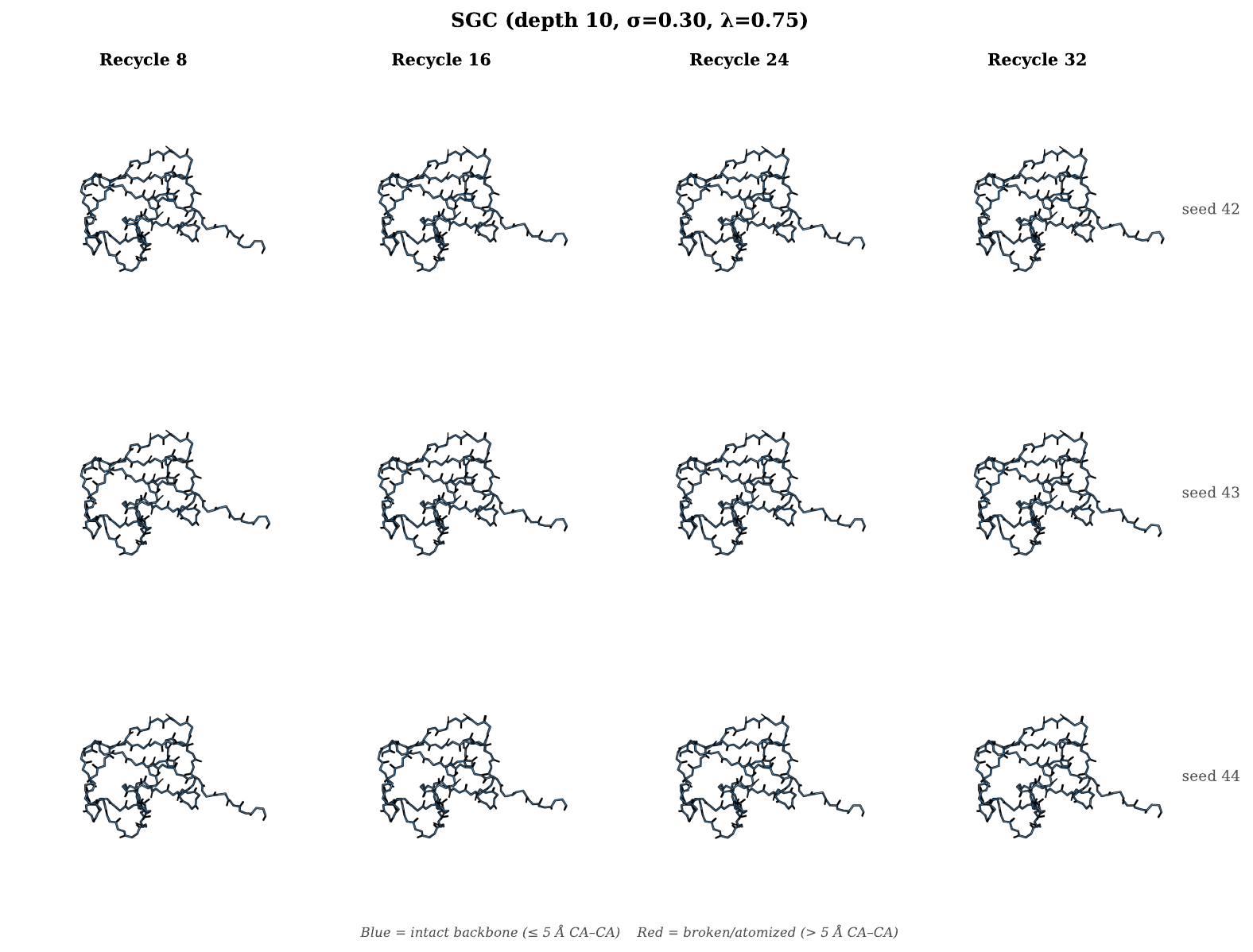}
    \caption{\textbf{\sgc{} at depth 10 ($\sig = 0.30$, $\lam = 0.75$): deterministic chain integrity.}
    All 12 panels show the same compact ubiquitin fold---identical across seeds and recycles. Zero broken CA--CA bonds in any frame (0/75 per structure, RMSD 0.6--0.7~\AA{} to native). The visual monotony is the result: \sgc{} produces deterministic, reproducible, physically connected structures at this perturbation depth.}
    \label{fig:supp_breakage_sgc}
\end{figure}

\begin{figure}[H]
    \centering
    \includegraphics[width=\textwidth]{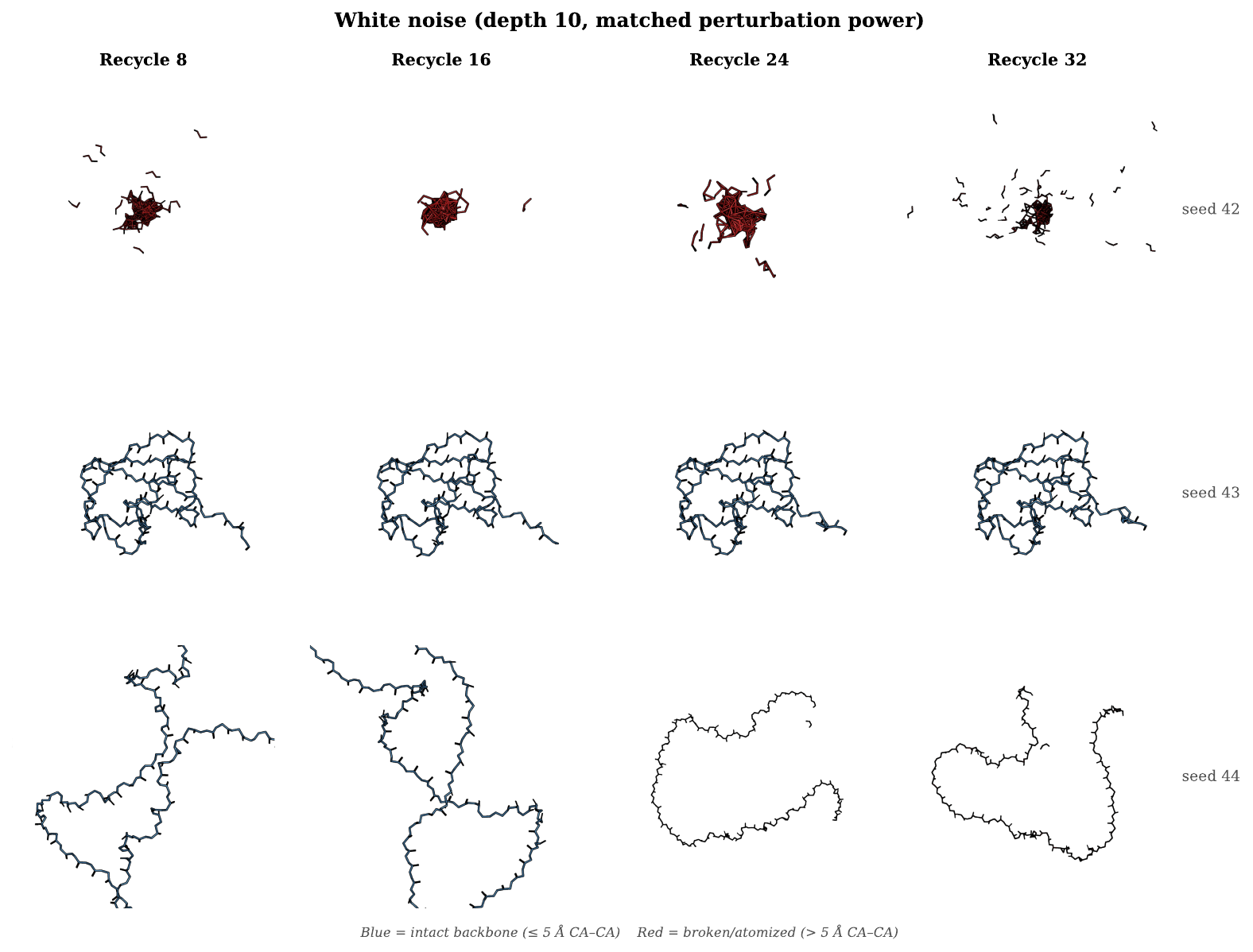}
    \caption{\textbf{White noise at depth 10 (matched perturbation power): stochastic, seed-dependent fates.}
    Three seeds at the same perturbation depth produce three different outcomes. \textbf{Seed~42} (top row): chaotic breakage that fluctuates across recycles (28, 10, 28, 58 broken bonds at recycles 8, 16, 24, 32). The structure never converges. \textbf{Seed~43} (middle row): intact throughout (0/75 broken, RMSD 0.9--1.2~\AA{}). \textbf{Seed~44} (bottom row): unfolded but mostly connected (RMSD ${\sim}23$~\AA{}, 0 broken bonds at recycles 8--16, developing 1 stretched bond by recycle 24). The same perturbation power produces intact, unfolded, or atomized structures depending solely on the random seed.}
    \label{fig:supp_breakage_white}
\end{figure}

\begin{figure}[H]
    \centering
    \includegraphics[width=\textwidth]{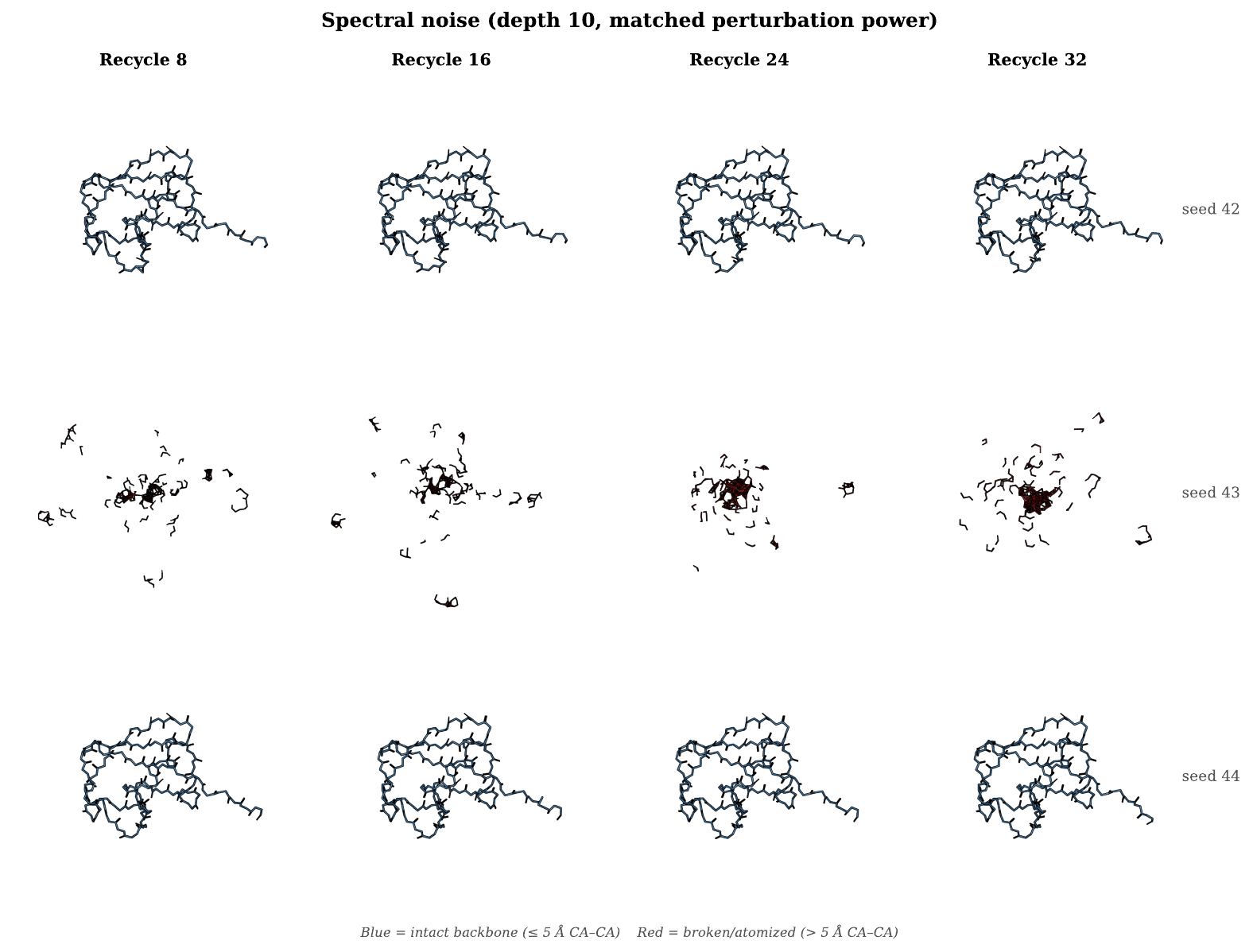}
    \caption{\textbf{Spectral noise at depth 10 (matched perturbation power): bimodal fate.}
    \textbf{Seeds~42 and 44} (rows 1, 3): intact and near-native across all recycles (0/75 broken, RMSD 0.7--0.8~\AA{}). \textbf{Seed~43} (row 2): consistently shattered at every recycle (46--59/75 broken bonds), with no recovery across recycling iterations. Unlike white noise, which shows chaotic oscillation (Figure~\ref{fig:supp_breakage_white}, seed 42), spectral noise fate is binary---seeds either survive or are irreversibly destroyed.}
    \label{fig:supp_breakage_spectral}
\end{figure}

\begin{figure}[H]
    \centering
    \includegraphics[width=\textwidth]{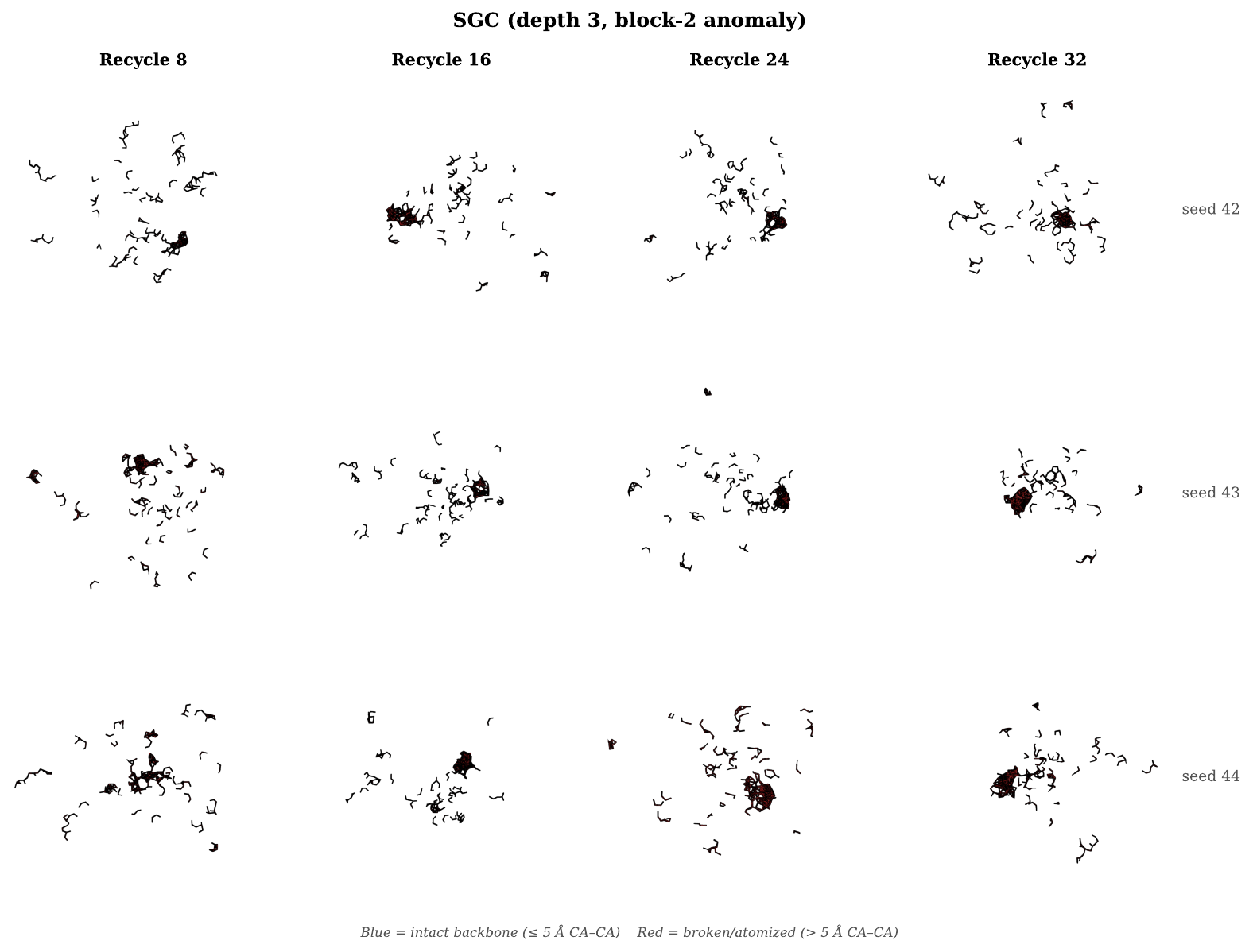}
    \caption{\textbf{\sgc{} at depth 3 (block-2 anomaly): universal chain breakage.}
    When \sgc{} perturbs only blocks 0--2, all three seeds produce extensively broken chains (43--66 out of 75 bonds broken, RMSD 15--19~\AA{}). This is the known block-2 anomaly: Evoformer block~2 is uniquely sensitive to weight perturbation, and perturbing it produces catastrophic structural failure regardless of seed. This depth is excluded from all analyses in the main text. The anomaly is reproducible and deterministic---in contrast to the seed-dependent breakage pattern of noise controls (Figures~\ref{fig:supp_breakage_white}--\ref{fig:supp_breakage_spectral}).}
    \label{fig:supp_breakage_blk2}
\end{figure}

\section{Unfolding pathway: parameter-grid sweeps}
\label{sec:supp-unfolding-multicondition}


Section~\ref{sec:results-unfolding} illustrates the group-resolved $Q$-factor analysis at a single canonical condition ($\sig = 0.30$, $\lam = 0.75$). Here we show that G2 (intermediate) consistently breaks first and G1 (nucleus) never breaks before the other groups, across the full $\sig \times \lam$ grid and three MSA-subsampling seeds. In 67 of 84 qualifying conditions G1 strictly outlasts G3 ($\beta$5); in the remaining 17 (concentrated at $\lam = 0.75$--$0.76$) the two groups lose their contacts at the same depth.

\begin{figure}[H]
    \centering
    \includegraphics[width=\textwidth]{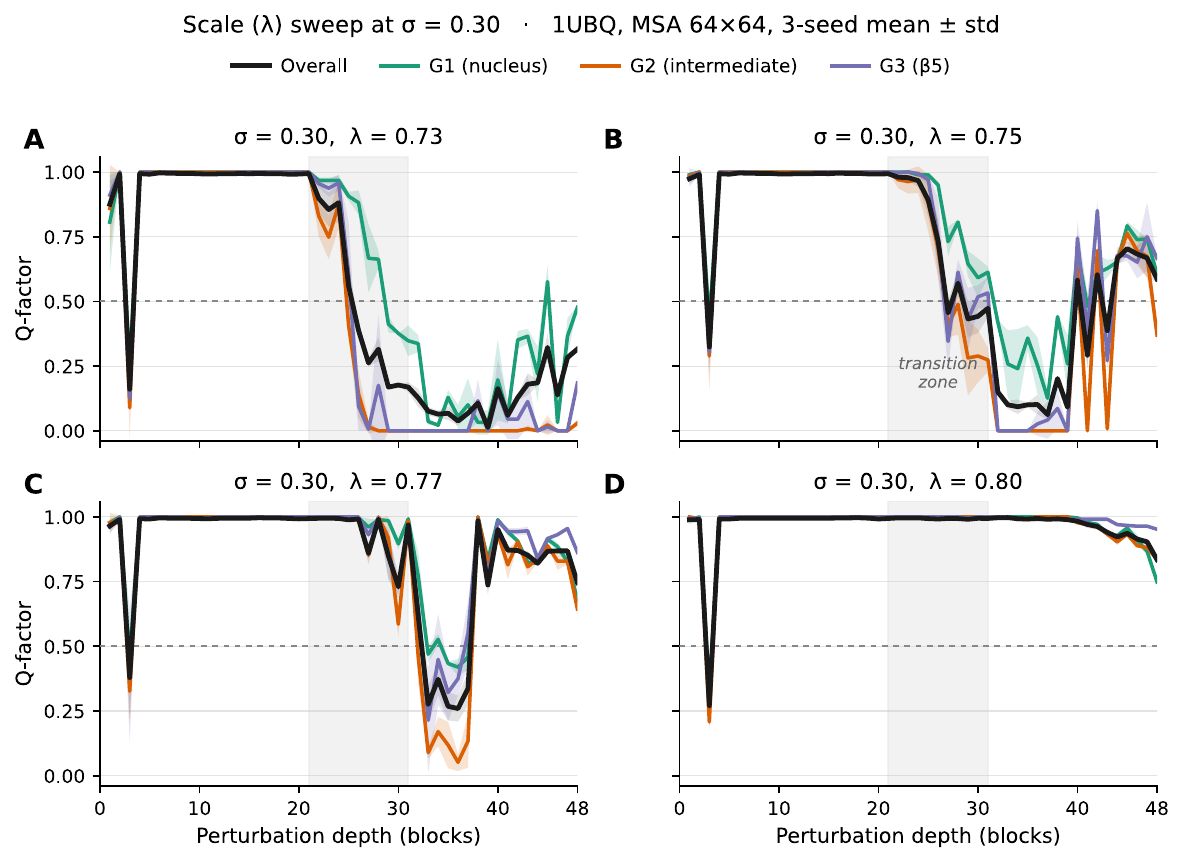}
    \caption{\textbf{Group-resolved $Q$-factor under a scale sweep at $\sig = 0.30$.}
    Four regimes spanning the transition boundary.
    \textbf{(A)}~$\lam = 0.73$: deep perturbation (depth $\gtrsim 25$) fully dismantles all three contact groups; G2 (orange, intermediate) drops first, G1 (teal, nucleus) survives longest, and G3 (purple, $\beta$5) follows G2.
    \textbf{(B)}~$\lam = 0.75$: the canonical reporting condition; the hierarchy is clearest here, with a well-defined gap between groups across depths 25--35.
    \textbf{(C)}~$\lam = 0.77$: partial collapse---structures partially recover at deeper depths, but the ordering at each depth is preserved.
    \textbf{(D)}~$\lam = 0.80$: perturbation too mild to drop any group sustainedly below 0.5; the ordering holds for transient excursions.
    Grey band: approximate transition zone (depths 21--31). Curves: mean $\pm$ std across three MSA-subsampling seeds. Dashed line: $Q = 0.5$ threshold. 1UBQ, model~1, MSA $64 \times 64$.}
    \label{fig:supp_unfolding_lambda}
\end{figure}

\begin{figure}[H]
    \centering
    \includegraphics[width=\textwidth]{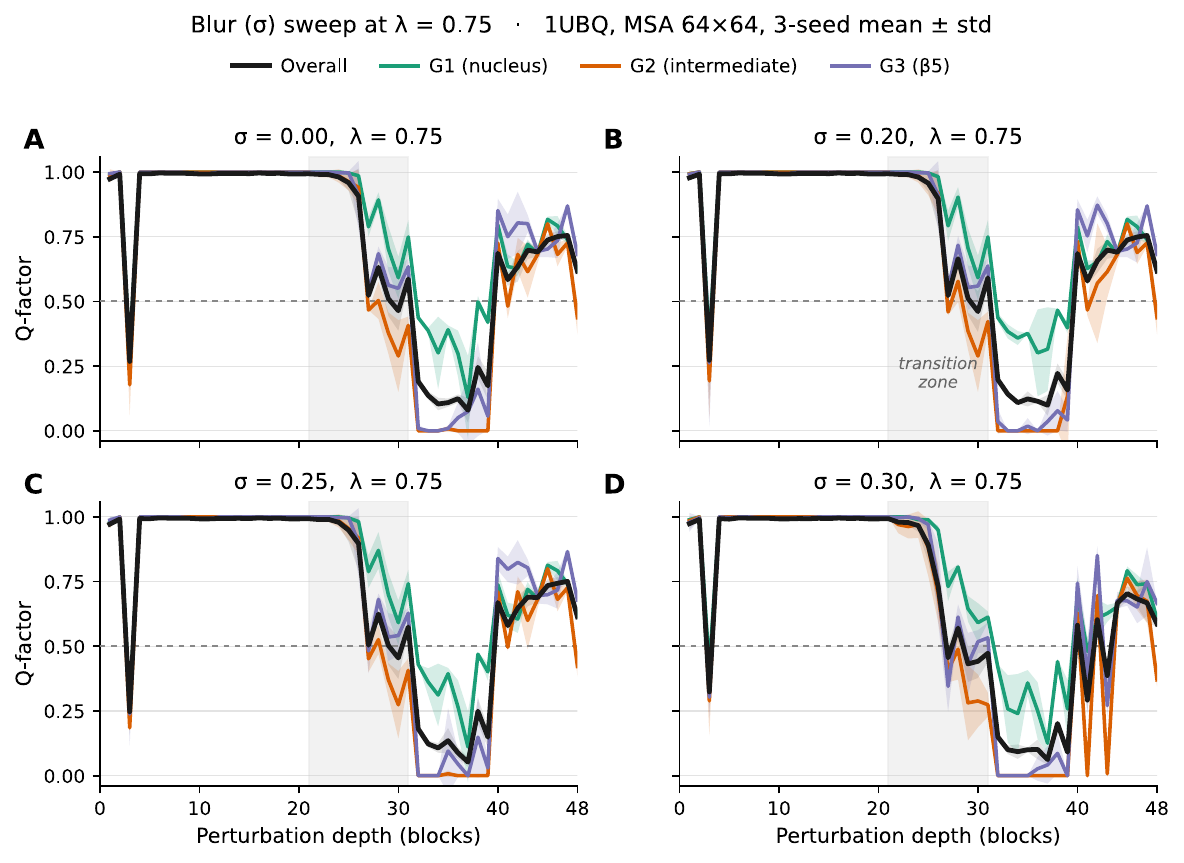}
    \caption{\textbf{Group-resolved $Q$-factor under a blur sweep at $\lam = 0.75$.}
    Four blur widths at the canonical scaling condition.
    \textbf{(A)}~$\sig = 0.00$ (pure scaling): the nucleus (G1) and $\beta$5 (G3) break at nearly identical depths---scaling alone cannot resolve them.
    \textbf{(B--D)}~$\sig = 0.20, 0.25, 0.30$: as $\sig$ increases, G3 exhibits progressively deeper transient dips while G1 remains relatively intact, producing the blur-specific separation discussed in Section~\ref{sec:results-unfolding}. Both groups permanently lose their contacts at the same depth ($\sim$32), but the transient vulnerability of G3 at intermediate depths grows with blur.
    The across-seed spread (shaded bands) also widens with $\sig$ at post-transition depths, consistent with smoothing amplifying the response near the transition boundary (Section~\ref{sec:results-boundary}). Conventions as in Figure~\ref{fig:supp_unfolding_lambda}.}
    \label{fig:supp_unfolding_sigma}
\end{figure}

\begin{figure}[H]
    \centering
    \includegraphics[width=\textwidth]{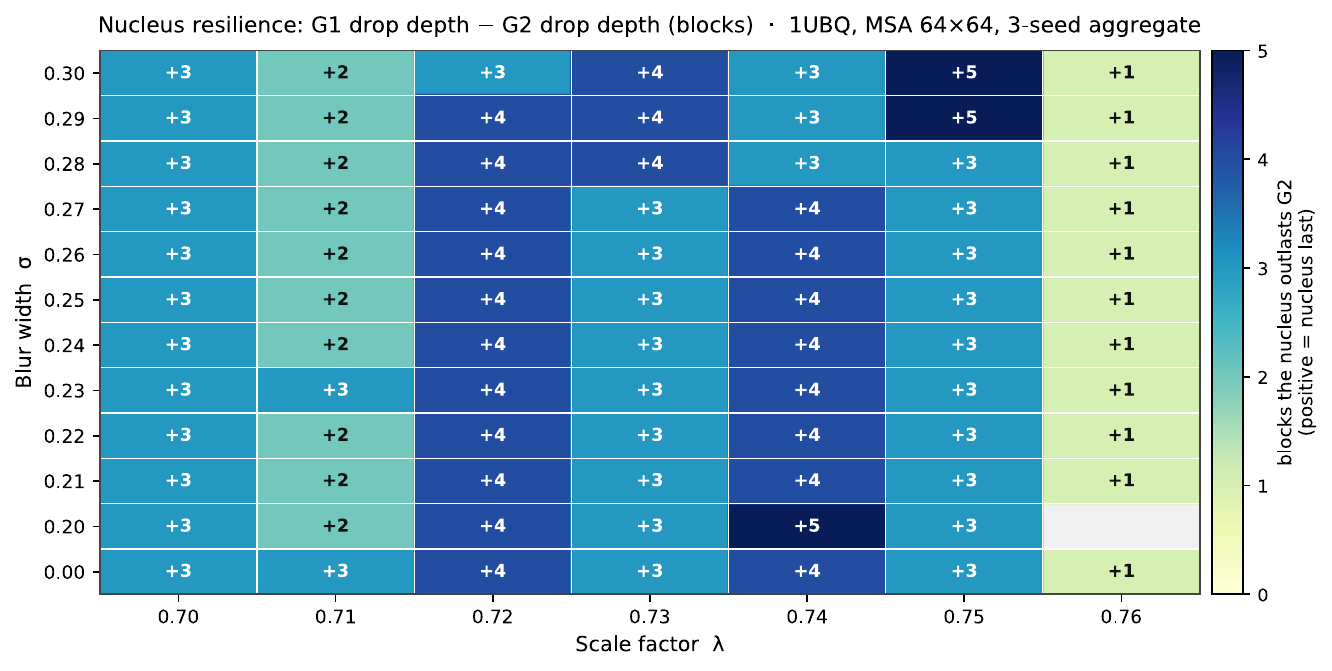}
    \caption{\textbf{The nucleus consistently outlasts the intermediate across the $\sig \times \lam$ grid.}
    Difference between the first depth at which G1 (nucleus) sustains $Q < 0.5$ for three consecutive depths and the equivalent depth for G2 (intermediate). Positive values (darker blue) indicate the nucleus survives longer. Of the 84 ($\sig$, $\lam$) conditions in which both groups cross the 0.5 threshold, all 84 show the nucleus outlasting the intermediate. The tested range ($\sig = 0.00$--$0.30$, $\lam = 0.70$--$0.77$) spans the entire transition boundary and the regime immediately below it. Blank cells correspond to conditions where one or both groups never cross $Q < 0.5$ within depth~48. 1UBQ, model~1, MSA $64 \times 64$, 3-seed aggregate.}
    \label{fig:supp_unfolding_resilience}
\end{figure}

\begin{table}[H]
\centering
\caption{\textbf{Residue-pair contacts used in the group-resolved $Q$-factor analysis of Section~\ref{sec:results-unfolding}.}
\textbf{(A)}~Eight specific native contacts drawn from the ubiquitin folding literature~\cite{went2005, sosnick2004, piana2013}, anchoring the contact-group definitions analysed in main Fig.~\ref{fig:unfolding}. Residue numbering is PDB-based (1-indexed); secondary-structure assignments follow the conventions of the top blur-sensitive pairs table (E = $\beta$-strand, H = $\alpha$-helix, L = loop/turn).
\textbf{(B)}~Region-based contact definitions used to compute the group-resolved $Q$-factors reported in the main text. Every C$\alpha$--C$\alpha$ pair $(i, j)$ with one residue in each listed region contributes to the group count, subject to a 12~\AA{} distance cutoff and minimum sequence separation $|i - j| > 3$. The resulting counts (G1: 93 contacts, G2: 45, G3: 112) correspond to the totals quoted in Section~\ref{sec:results-unfolding}. Regions are reported in PDB (1-indexed) numbering.}
\label{tab:supp_unfolding_contacts}
\small

\textbf{(A) Specific native contacts from the folding literature}\\[2pt]
\begin{tabular}{lllccl}
\toprule
Group & Pair & Residues & $|i - j|$ & SS & Structural role \\
\midrule
G1 (nucleus)      & (5, 18)  & Val--Glu & 13 & E--E & $\beta$1 to $\beta$2 strand --- hairpin docking \\
G1 (nucleus)      & (2, 31)  & Gln--Gln & 29 & L--L & $\beta$1 to $\alpha$1 --- nucleus clasp \\
G1 (nucleus)      & (22, 35) & Thr--Gly & 13 & L--L & helix N-cap to C-cap turn \\
\midrule
G2 (intermediate) & (15, 51) & Leu--Glu & 36 & E--L & $\beta$2 to $\beta$3--$\beta$5 loop \\
G2 (intermediate) & (28, 61) & Ala--Ile & 33 & H--L & $\alpha$1 to 3$_{10}$ loop \\
G2 (intermediate) & (25, 61) & Asn--Ile & 36 & H--L & $\alpha$1 to 3$_{10}$ loop \\
\midrule
G3 ($\beta$5)     & (18, 64) & Glu--Glu & 46 & E--L & $\beta$2 region to $\beta$5 \\
G3 ($\beta$5)     & (16, 62) & Glu--Gln & 46 & E--L & $\beta$2 to pre-$\beta$5 loop \\
\bottomrule
\end{tabular}

\vspace{1em}

\textbf{(B) Region-based contact definitions for the group-resolved $Q$-factors}\\[2pt]
\begin{tabular}{lllll}
\toprule
Group & Region 1 & Region 2 & Structural element & Count \\
\midrule
G1 (nucleus)      & 1--7   & 10--17 & $\beta$1 $\leftrightarrow$ $\beta$2 (hairpin)              & \multirow{2}{*}{93} \\
G1 (nucleus)      & 1--17  & 23--34 & $\beta$1--$\beta$2 hairpin $\leftrightarrow$ $\alpha$1      &                    \\
\midrule
G2 (intermediate) & 23--34 & 48--62 & $\alpha$1 $\leftrightarrow$ $\beta$3--$\beta$5 / 3$_{10}$ loop & \multirow{2}{*}{45} \\
G2 (intermediate) & 10--17 & 48--62 & $\beta$2 $\leftrightarrow$ $\beta$3--$\beta$5 / 3$_{10}$ loop &                    \\
\midrule
G3 ($\beta$5)     & 1--17  & 64--72 & $\beta$1 / $\beta$2 $\leftrightarrow$ $\beta$5             & \multirow{2}{*}{112}\\
G3 ($\beta$5)     & 40--45 & 64--72 & $\beta$3 $\leftrightarrow$ $\beta$5                        &                    \\
\bottomrule
\end{tabular}
\end{table}

\section{Per-Pair Blur Specificity}


\begin{figure}[H]
    \centering
    \includegraphics[width=\textwidth]{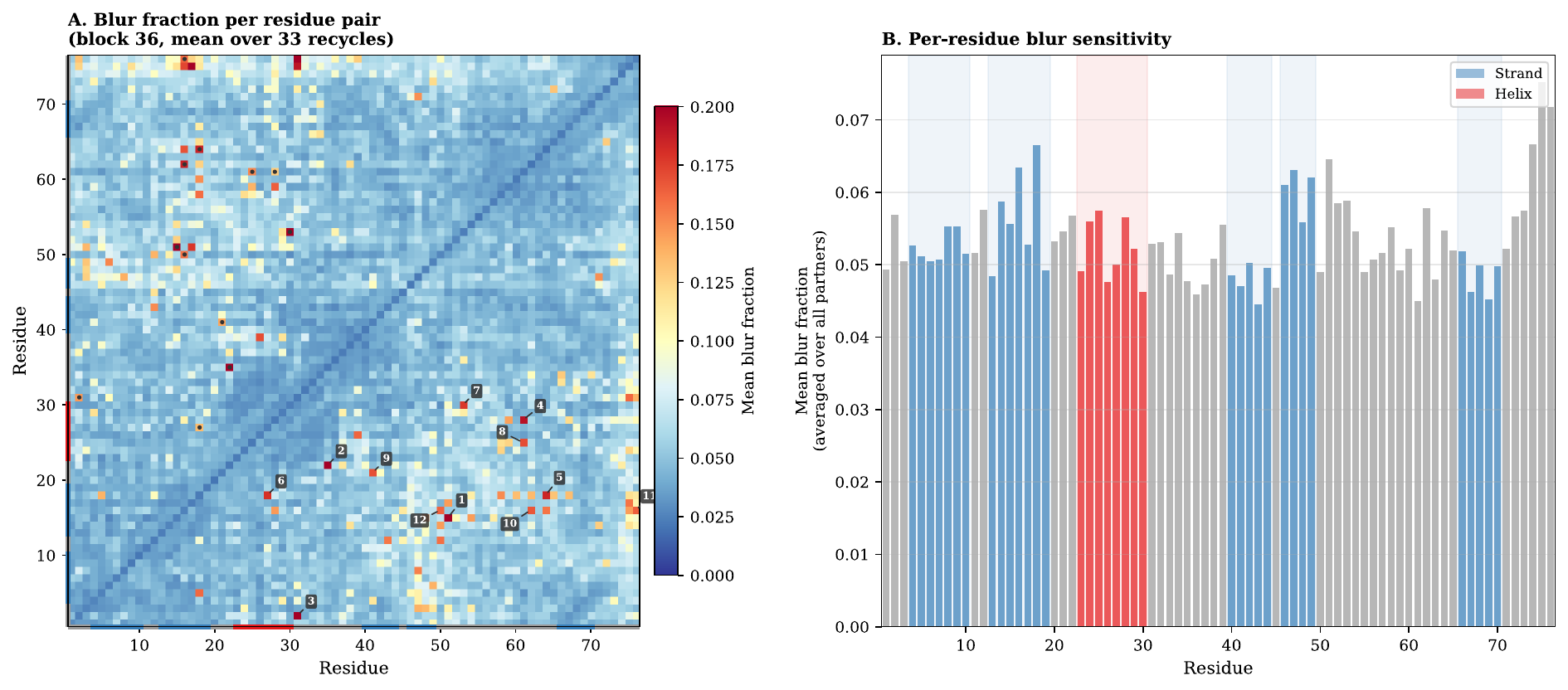}
    \caption{\textbf{Gaussian blur concentrates on residue pairs at kinetic-boundary regions.}
    \textbf{(A)}~Mean blur fraction per residue pair at Evoformer block~36, averaged over 33 recycling iterations. Blur fraction is defined as $\|\mathbf{z}^{\text{cumul}} - \mathbf{z}^{\text{scale}}\|_2 \,/\, \|\mathbf{z}^{\text{scale}} - \mathbf{z}^{\text{baseline}}\|_2$ for each pair $(i,j)$, where cumul = $\sig + \lam$, scale = $\lam$ only, and baseline = unperturbed. Most pairs show blur fractions below 5\%, but specific off-diagonal hotspots reach 15--21\%. Numbered labels (1--12) mark the top pairs (Table~\ref{tab:supp_blur_pairs}); corresponding positions in the lower triangle are marked with dots. Colored bars along axes indicate secondary structure (blue = strand, red = helix, grey = loop).
    \textbf{(B)}~Per-residue blur sensitivity (mean blur fraction averaged over all partners). Residues in the $\alpha$-helix (23--30) and $\beta$2 strand (13--19) show elevated sensitivity, consistent with their role at the kinetic boundary between folding nucleus and late-forming contacts.
    Condition: $\sig = 0.30$, $\lam = 0.75$, MSA $64 \times 64$, seed~42.}
    \label{fig:supp_blur_perpair}
\end{figure}

\begin{table}[H]
\centering
\caption{\textbf{Top 12 blur-sensitive residue pairs.}
Ranked by mean blur fraction at Evoformer block~36, averaged over 33 recycling iterations. SS = secondary structure (E = strand, H = helix, L = loop). Sep = sequence separation $|i - j|$. CV = coefficient of variation across recycles (lower = more consistent). Folding groups: G1 = folding nucleus contacts (rate-limiting), G2 = native/intermediate boundary (last-folding contacts), G3 = post-TSE $\beta$5 sheet contacts. Annotations based on ubiquitin folding literature~\cite{went2005,sosnick2004,piana2013}. Pair (16, 76) is marked anomalous: no known folding contact at this position; possibly an AF2 encoding artifact or polyubiquitin-chain geometry.}
\label{tab:supp_blur_pairs}
\small
\begin{tabular}{rlccccl}
\toprule
Rank & Pair & SS & $|i-j|$ & Blur frac. & CV & Folding role \\
\midrule
1 & (15, 51) & E--L & 36 & 0.213 & 0.36 & G2: Nucleus to last-folding loop \\
2 & (22, 35) & L--L & 13 & 0.212 & 0.33 & G1: Helix N-cap to C-cap boundary \\
3 & (2, 31)  & L--L & 29 & 0.209 & 0.30 & G1: $\beta$1 to $\alpha$-helix --- nucleus \\
4 & (28, 61) & H--L & 33 & 0.192 & 0.40 & G2: $\alpha$-helix to 3$_{10}$ loop --- last to fold \\
5 & (18, 64) & E--L & 46 & 0.184 & 0.39 & G3: $\beta$2 to $\beta$5 --- post-TSE sheet \\
6 & (18, 27) & E--H &  9 & 0.180 & 0.51 & G1: $\beta$2 to $\alpha$-helix --- nucleus packing \\
7 & (30, 53) & H--L & 23 & 0.177 & 0.64 & G2: Helix C-term to loop \\
8 & (25, 61) & H--L & 36 & 0.170 & 0.37 & G2: $\alpha$-helix to 3$_{10}$ loop --- late intermediate \\
9 & (21, 41) & L--E & 20 & 0.169 & 0.42 & G2: $\beta$2 turn to $\beta$3 --- intermediate \\
10 & (16, 62) & E--L & 46 & 0.167 & 0.36 & G3: $\beta$2 to $\beta$5 region --- post-TSE \\
11 & (16, 76) & E--L & 60 & 0.165 & 0.59 & Anomalous: $\beta$2 to C-terminal tail \\
12 & (16, 50) & E--L & 34 & 0.163 & 0.39 & G2: $\beta$2 to loop \\
\bottomrule
\end{tabular}
\end{table}

\section{Power Decomposition and Kernel Independence of \sgc{}}
\label{sec:supp-power-decomp}


This note provides the quantitative basis for claims in Sections~\ref{sec:methods-sgc} and~\ref{sec:results-sigma-lambda} of the main text regarding the relative contributions of scaling and smoothing within the \sgc{} perturbation.

\subsection{Power decomposition}

The \sgc{} perturbation $\Delta W = \lam \cdot G_\sig(W) - W$ decomposes exactly into a scaling component $\Delta_{\text{scale}} = (\lam - 1)W$ and a smoothing residual $\Delta_{\text{blur}} = \lam(G_\sig(W) - W)$:
\begin{equation}
\Delta W = \underbrace{(\lam - 1)W}_{\Delta_{\text{scale}}} + \underbrace{\lam\bigl(G_\sig(W) - W\bigr)}_{\Delta_{\text{blur}}}
\end{equation}

We computed $\|\Delta_{\text{scale}}\|^2$ and $\|\Delta_{\text{blur}}\|^2$ for every weight tensor in the AlphaFold2 Evoformer at our primary operating point ($\sig = 0.30$, $\lam = 0.75$), summed across all 4{,}464 tensors (93 unique parameter tensors per block $\times$ 48 blocks, totaling 87{,}837{,}696 parameters). The results are shown in Table~\ref{tab:supp-power}.

\begin{table}[H]
\centering
\caption{\textbf{Power decomposition of \sgc{} at $\sig = 0.30$, $\lam = 0.75$.} L2 norms and power fractions computed across all Evoformer weight tensors. The smoothing residual $\Delta_{\text{blur}}$ is further decomposed into a component parallel to $\Delta_{\text{scale}}$ (i.e., parallel to $W$) and a component orthogonal to it.}
\label{tab:supp-power}
\small
\begin{tabular}{lrr}
\toprule
Component & $\|{\cdot}\|_2$ & Power fraction \\
\midrule
$\Delta_{\text{scale}} = (\lam - 1)W$ & 526.44 & 99.57\% \\
$\Delta_{\text{blur}} = \lam(G_\sig(W) - W)$ & 34.74 & 0.43\% \\
\quad Parallel to $W$ (redundant scaling) & 30.39 & 0.33\% \\
\quad Orthogonal to $W$ (unique direction) & 16.81 & 0.10\% \\
\midrule
Cosine similarity $(\Delta_{\text{blur}}, W)$ & \multicolumn{2}{c}{0.875} \\
\bottomrule
\end{tabular}
\end{table}

The per-tensor power fraction of scaling ranges from 99.51\% to 100\% (median 99.84\%) across all 4{,}464 tensors. At $\sig = 0.30$, the Gaussian kernel is $[0.004, 0.992, 0.004]$---nearly an identity operation---so the smoothing residual is dominated by its projection onto the weight direction (76.6\% parallel, 23.4\% orthogonal). The geometrically unique component---the 0.10\% of total power that moves orthogonally to $W$---is nonetheless the component responsible for the structural effects described in Section~\ref{sec:results-sigma-lambda}, because uniform scaling along $W$ cannot alter the relative structure within each weight matrix, while the orthogonal component does.

\subsection{Frequency-band composition}
\label{sec:supp-band-decomp}

\begin{figure}[H]
\centering
\includegraphics[width=\textwidth]{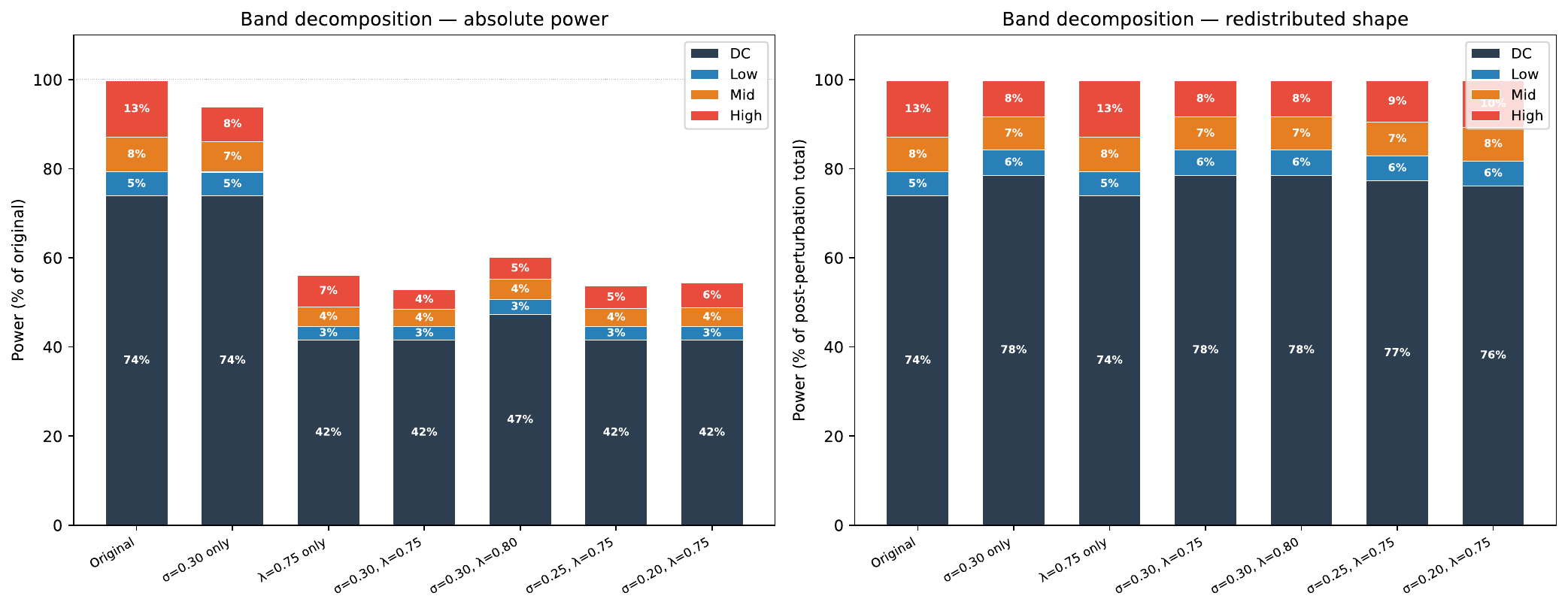}
\caption{\textbf{Frequency-band composition of Evoformer parameters under \sgc{}.} Power in four frequency bands (DC: $[0, 0.01)$; Low: $[0.01, 0.10)$; Mid: $[0.10, 0.25)$; High: $[0.25, 0.50)$) for seven perturbation conditions, computed by applying the analytical power-transfer function $|H(f)|^2 = \exp(-4\pi^2 \sig^2 f^2)$ to the averaged Evoformer parameter spectrum (93 parameter arrays, 160 (tensor, axis) spectra; see Algorithm~\ref{alg:band-decomp}). \textbf{Left:} absolute band power as a percentage of the unperturbed reference. Pure scaling ($\sig = 0$, $\lam = 0.75$) attenuates every band uniformly to $\lam^2 \approx 56\%$; adding Gaussian smoothing at $\sig = 0.30$ preferentially damps the High band while leaving DC and Low nearly untouched. \textbf{Right:} the same band powers renormalized to the post-perturbation total. Pure scaling preserves the spectral shape exactly, while Gaussian smoothing shifts it modestly toward DC (DC share: $73.8\% \to 78.5\%$; High share: $12.7\% \to 8.2\%$). The surviving power is therefore a rescaled, gently low-passed version of the original Evoformer spectrum, not a surgical excision of any single band.}
\label{fig:supp_band_decomp}
\end{figure}

The norm decomposition above quantifies \emph{how much} power \sgc{} removes. A companion question is \emph{where in frequency space} that removal lands: does the perturbation act uniformly across the spectrum of Evoformer parameter values, or does it selectively damp particular bands? We answer this by taking a one-dimensional real FFT along each non-block axis of every Evoformer parameter tensor carrying a 48-block axis (93 such tensors, including weight matrices, biases, and LayerNorm scale/offset terms), averaging the resulting power spectra over the remaining axes, interpolating onto a common 512-point grid in $[0, 0.5]$ cycles/index, and taking the arithmetic mean across the resulting 160 (tensor, axis) spectra. \sgc{} acts as a multiplicative filter in the frequency domain: its effect on any spectrum $\bar{S}(f)$ is
\begin{equation}
S_{\sgc{}}(f) = \lam^2 \, |H(f)|^2 \, \bar{S}(f), \qquad |H(f)|^2 = \exp\!\bigl(-4\pi^2 \sig^2 f^2\bigr),
\end{equation}
where $|H(f)|^2$ is the squared magnitude of the Fourier transform of the Gaussian smoother.\footnote{The amplitude-transfer function $H(f) = \exp(-2\pi^2 \sig^2 f^2)$ is the continuous Fourier transform of a unit-area Gaussian with standard deviation $\sig$; the power-transfer is its square.} Integrating this expression over four frequency windows (DC, Low, Mid, High) at each of the seven $(\sig, \lam)$ conditions produces Figure~\ref{fig:supp_band_decomp}. Algorithm~\ref{alg:band-decomp} specifies the computation exactly as implemented in \texttt{1ubq\_analysis/evoformer\_spectral\_distribution.py}.

\begin{spacing}{1.2}
\begin{algorithm}[H]
\SetInd{0.6em}{1.0em}
\DontPrintSemicolon
\caption{\textsc{BandPowerSpectrum}---frequency-band decomposition of Evoformer weights under \sgc{}}
\label{alg:band-decomp}
\KwIn{parameter archive $P$; band edges $\mathcal{B} = \{[0, 0.01), [0.01, 0.10), [0.10, 0.25), [0.25, 0.50)\}$; conditions $\mathcal{C} = \{(\sig_c, \lam_c)\}$; grid size $M = 512$}
\KwOut{band-power matrix $B[c, b]$}
\tcp{(1) Select Evoformer tensors with a 48-block axis}
$\mathcal{E} \leftarrow \{k \in P.\text{keys} : \text{``evoformer''} \in k \text{ and } 48 \in \text{shape}(P[k])\}$\;
$g \leftarrow \text{linspace}(0, 0.5, M)$ \tcp*{common grid, cycles/index}
$\mathcal{S} \leftarrow [\,]$\;
\tcp{(2) Per-tensor 1D real FFTs, averaged over all non-transform axes}
\ForEach{tensor key $k \in \mathcal{E}$}{
    $W \leftarrow P[k]$;\ \ $a_\text{blk} \leftarrow$ first axis of shape$(W)$ with size $48$\;
    \ForEach{$a \in \{0, \ldots, \text{ndim}(W)-1\} \setminus \{a_\text{blk}\}$ with shape$(W)[a] \geq 4$}{
        $F \leftarrow \texttt{rfft}(W,\ \text{axis}=a)$ \tcp*{real-input FFT}
        $S \leftarrow \text{mean}(|F|^2,\ \text{over every axis} \neq a)$\;
        $f_\text{bin} \leftarrow \texttt{rfftfreq}(\text{shape}(W)[a])$\;
        $\tilde{S} \leftarrow \texttt{interp}(g,\ f_\text{bin},\ S,\ \text{left}=S[0],\ \text{right}=0)$\;
        append $\tilde{S}$ to $\mathcal{S}$\;
    }
}
$\bar{S} \leftarrow \text{mean}(\mathcal{S},\ \text{axis}=0)$ \tcp*{averaged reference spectrum}
\tcp{(3) Apply \sgc{} transfer function and integrate each band}
\ForEach{$(\sig, \lam) \in \mathcal{C}$}{
    $H^2 \leftarrow \exp(-4\pi^2 \sig^2\, g^2)$ \tcp*{Gaussian power-transfer}
    $S_c \leftarrow \lam^2 \cdot H^2 \cdot \bar{S}$\;
    \ForEach{band $[f_\text{lo}, f_\text{hi}) \in \mathcal{B}$}{
        mask $\leftarrow (g \geq f_\text{lo}) \wedge (g < f_\text{hi})$\;
        $B[c, b] \leftarrow \texttt{trapezoid}(S_c[\text{mask}],\ g[\text{mask}])$\;
    }
}
\Return{$B$}\;
\end{algorithm}
\end{spacing}

Two observations follow. First, total surviving power tracks $\lam^2$ closely: blur alone ($\sig = 0.30$, $\lam = 1$) removes only ${\sim}6\%$ of total power, while $\lam = 0.75$ brings survival into the $53$--$56\%$ range across the $\sig$ values tested (against the scaling-only expectation $\lam^2 \approx 56\%$). Second, at $\sig = 0.30$---the upper end of the operating range used throughout the main text---the power-transfer function attenuates the Nyquist frequency by $1 - \exp(-\pi^2 \sig^2) \approx 59\%$ relative to DC, yet the redistribution of relative band shares is only about $4.5$--$5$~percentage points (DC rises from $73.8\%$ to $78.5\%$; High drops from $12.7\%$ to $8.2\%$). \sgc{} therefore acts as a near-uniform attenuator in the frequency domain, not as a selective spectral filter. This is consistent with the kernel-independence observations in \S\ref{sec:supp-kernel-indep}: at $\sig = 0.30$ any near-identity smoother (Gaussian, median, box) produces the same structural output because none of them meaningfully reshapes the parameter spectrum.

Magnitude is not the whole picture, however. The subsections that follow (\S\ref{sec:supp-directional}, \S\ref{sec:supp-permutation}) show that although the smoothing residual is sub-percent by power and redistributes the spectrum by only a few percentage points, the \emph{direction} of that residual in parameter space is geometrically specific: the orthogonal components of local kernels (Gaussian, median, box) share moderate-to-high cosine alignment ($0.63$--$0.83$), while an index-permuted Gaussian---carrying identical smoothing power but scrambled spatial structure---is nearly orthogonal to all three ($\sim 0.15$) and produces a measurably weaker structural effect in the regime where blur dominates ($\sig = 0.50$, $\lam = 1.0$: RMSD $1.84$~\AA{} vs.\ $2.45$~\AA{} for unpermuted). \sgc{}'s structural footprint is therefore shaped by \emph{where} in parameter space this small removal points, not by \emph{how much} power it carries.

\subsection{Kernel independence}
\label{sec:supp-kernel-indep}

To test whether the Gaussian kernel plays a privileged role, we replaced it with two alternative local smoothing kernels---median filtering (size~3) and box filtering (size~3)---applied to the same weight tensors under otherwise identical conditions ($\sig = 0.30$, $\lam = 0.75$, all 48 Evoformer blocks, MSA $64 \times 64$, seed~42). Because the raw alternative kernels differ in power from the Gaussian, we power-matched each kernel by scaling its smoothing residual $\Delta_{\text{blur}}^{(\text{alt})}$ by $\alpha = \sqrt{\|\Delta_{\text{blur}}^{(\text{gauss})}\|^2 \,/\, \|\Delta_{\text{blur}}^{(\text{alt})}\|^2}$ per tensor. Table~\ref{tab:supp-kernels} summarizes the structural outputs.

\begin{table}[H]
\centering
\caption{\textbf{Kernel independence at $\sig = 0.30$, $\lam = 0.75$.} All three local kernels produce indistinguishable structures when power-matched. Helix and strand fractions are from DSSP. Clashes counted at 2.0~\AA{} heavy-atom threshold. All runs: cumulative depth~48, MSA $64 \times 64$, seed~42.}
\label{tab:supp-kernels}
\small
\begin{tabular}{lccccc}
\toprule
Kernel & RMSD (\AA) & \plddt{} & pTM & Helix\% & Strand\% \\
\midrule
Gaussian ($\sig = 0.30$) & 2.55 & 74.5 & 0.368 & 19.7 & 25.0 \\
Median (size~3, matched) & 2.54 & 74.7 & 0.368 & 19.7 & 25.0 \\
Box (size~3, matched) & 2.57 & 74.3 & 0.366 & 19.7 & 25.0 \\
\bottomrule
\end{tabular}
\end{table}

The explanation is straightforward: at $\sig = 0.30$, all three kernels are size-3 operations with center weights $\geq 0.99$, making them near-identity transformations. The choice of kernel is immaterial at this operating point. We retain the Gaussian for its analytical tractability (closed-form Fourier transform, natural connection to diffusion processes, smooth interpolation to larger $\sig$).

\begin{figure}[H]
\centering
\includegraphics[width=\textwidth]{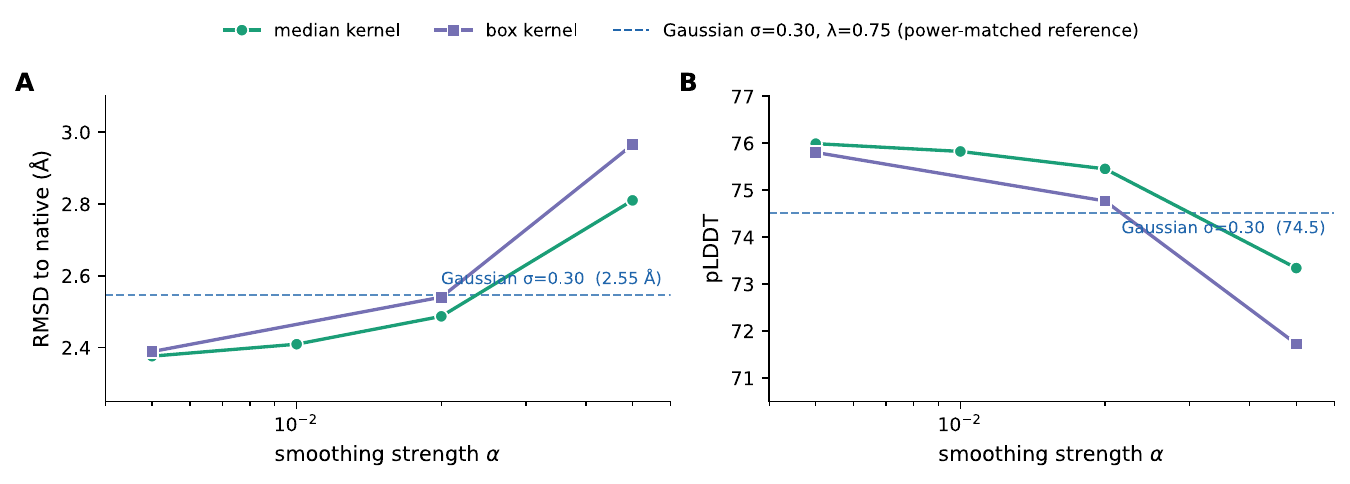}
\caption{\textbf{Independent-$\alpha$ dose-response for median and box kernels.} Structural outputs at cumulative depth~48, $\lam = 0.75$, MSA $64 \times 64$, single seed (42), ubiquitin \texttt{model\_1\_ptm}. \textbf{A:} RMSD to native. \textbf{B:} \plddt{}. The smoothing strength $\alpha$ is varied independently (not power-matched to the Gaussian). Both kernels produce graded, monotonic responses; at every shared $\alpha$ the two sweeps coincide to within $0.16$~\AA{} RMSD and $1.6$ \plddt{} units, with the largest gap falling at $\alpha = 0.05$ in both metrics. The dashed blue line marks the power-matched Gaussian reference ($\sig = 0.30$, RMSD $2.55$~\AA, \plddt{} $74.5$) from Table~\ref{tab:supp-kernels}; both sweeps cross this reference between $\alpha = 0.02$ and $\alpha = 0.05$. The box series omits $\alpha = 0.010$.}
\label{fig:supp_kernel_alpha}
\end{figure}

We also tested unmatched (independent) dose--response for both alternative kernels across a single-seed pilot sweep (seed~42), varying the smoothing strength $\alpha$ from 0.005 to 0.05 while holding $\lam = 0.75$ (Figure~\ref{fig:supp_kernel_alpha}). Both kernels produce a graded, monotonic structural response---median RMSD rises from $2.38$ to $2.81$~\AA{} and box RMSD from $2.39$ to $2.97$~\AA---and the two sweeps cross the power-matched Gaussian reference ($\sig = 0.30$, RMSD $2.55$~\AA) between $\alpha = 0.02$ and $\alpha = 0.05$. pLDDT mirrors this, declining monotonically from ${\sim}76$ at $\alpha = 0.005$ to $73.3$ (median) and $71.7$ (box) at $\alpha = 0.05$ and crossing the Gaussian reference in the same interval. The kernel-independence conclusion of Table~\ref{tab:supp-kernels} therefore holds across the full tested range of smoothing strengths, not just at the single power-matched point.

\subsection{Directional analysis}
\label{sec:supp-directional}

Although the three kernels produce identical structural outputs at $\sig = 0.30$, their orthogonal components point in distinguishable directions. We decomposed each kernel's smoothing residual into components parallel and orthogonal to $W$, then computed cosine similarity between the orthogonal subvectors (averaged across all 4{,}464 parameter tensors). Table~\ref{tab:supp-cosines} summarizes the results.

\begin{table}[H]
\centering
\caption{\textbf{Cosine similarity between orthogonal components of different kernels.} ``Permuted'' denotes a Gaussian kernel whose indices were randomly shuffled before application, destroying local weight structure while preserving global statistics.}
\label{tab:supp-cosines}
\small
\begin{tabular}{lcccc}
\toprule
 & Gaussian & Median & Box & Permuted \\
\midrule
Gaussian & 1.00 & 0.63 & 0.76 & 0.15 \\
Median & 0.63 & 1.00 & 0.83 & 0.14 \\
Box & 0.76 & 0.83 & 1.00 & 0.16 \\
Permuted & 0.15 & 0.14 & 0.16 & 1.00 \\
\bottomrule
\end{tabular}
\end{table}

The three local kernels share moderate-to-high directional alignment (cosine 0.63--0.83), while a permuted Gaussian---which applies the same magnitude of perturbation but in random directions---is nearly orthogonal to all three (cosine 0.14--0.16, consistent with random vectors in high-dimensional space). This confirms that local kernels probe a common subspace of weight space defined by local weight correlations.

\subsection{Permutation test}
\label{sec:supp-permutation}

To directly test whether local weight structure matters, we compared Gaussian smoothing against a permuted Gaussian: for each weight tensor, we randomly shuffled the indices along each axis before applying the Gaussian filter, then inverted the shuffling to produce a perturbation of identical power but scrambled spatial structure. At $\sig = 0.30$, both conditions produce identical output (as expected, since both are near-identity operations). At $\sig = 0.50$, $\lam = 1.0$ (pure smoothing, no scaling), the difference becomes significant:

\begin{itemize}
\item \textbf{Gaussian blur} ($\sig = 0.50$, $\lam = 1.0$, depth 48): RMSD = 2.45~\AA{}, \plddt{} = 70.1
\item \textbf{Permuted Gaussian} ($\sig = 0.50$, $\lam = 1.0$, depth 48): RMSD = 1.84~\AA{}, \plddt{} = 82.7
\end{itemize}

Permutation weakens the structural effect by $+$12.6 \plddt{} points and $-$0.61~\AA{} RMSD. The Gaussian blur exploits local correlations in the weight matrices; the permuted version destroys these correlations while preserving the aggregate smoothing power, producing a weaker and less structured perturbation. At $\sig \geq 1.0$, both conditions collapse the model entirely (radius of gyration $<$ 1~\AA{}, $>$27{,}000 clashes), making the comparison uninformative. The interpretable operating window for the permutation test is $\sig \approx 0.50$, $\lam = 1.0$.

This result has two implications. First, the local arrangement of weights within Evoformer tensors carries information that smoothing selectively accesses---it is not merely the aggregate attenuation that matters, but which features are averaged together. Second, the effect operates within a narrow $\sig$ window: too small and all kernels are identity; too large and the model collapses regardless. Our primary operating point ($\sig = 0.30$) lies at the lower end of this window, where smoothing is mild enough that its structural effects emerge only in combination with scaling near the transition boundary (Section~\ref{sec:results-sigma-lambda}).


\section{MSA depth sensitivity}
\label{sec:supp-msa-depth}

\begin{figure}[h]
\centering
\includegraphics[width=\textwidth]{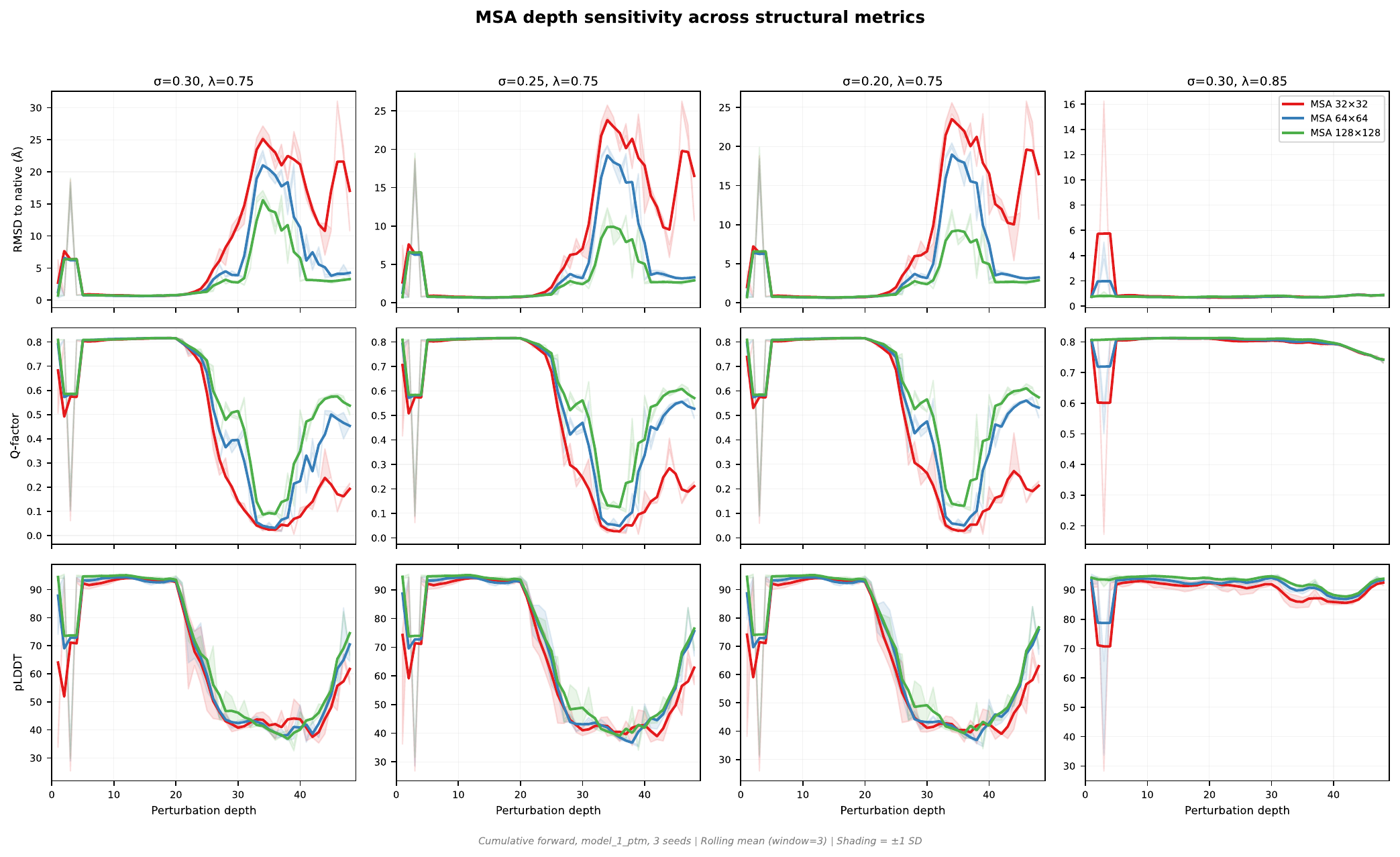}
\caption{\textbf{MSA depth sensitivity across structural metrics.} RMSD to native (top), Q-factor (middle), and \plddt{} (bottom) vs.\ perturbation depth for MSA depths 32 (red), 64 (blue), and 128 (green) at four representative $(\sig, \lam)$ conditions. Shaded regions show $\pm$1~SD across 3 seeds; lines are rolling means (window~3). MSA~32 departs from native at shallower depths and reaches higher RMSD; MSA~64 and MSA~128 track closely until deep perturbation ($d > 30$), where MSA~128 retains modestly better quality. The block-2 anomaly (depth~3) appears as a transient spike across all metrics and depths. \plddt{} (bottom row) is the model's self-assessed confidence and may not track structural quality for strongly perturbed conditions; the RMSD and Q-factor rows provide direct structural validation.}
\label{sfig:msa-sensitivity}
\end{figure}

To justify the choice of MSA depth $64 \times 64$ and assess its influence on the reported phenomena, we repeated a subset of the ubiquitin perturbation grid at MSA depths $32 \times 32$ and $128 \times 128$. Seven matched $(\sig, \lam)$ conditions were run at all 48 perturbation depths with 3 seeds each (4{,}032 runs per depth; 2{,}016 cumulative forward runs per MSA depth for model\_1\_ptm). Table~\ref{tab:msa-baselines} summarizes the unperturbed baselines, and Fig.~\ref{sfig:msa-sensitivity} shows the key comparisons.

\begin{table}[h]
\caption{\textbf{Baseline prediction quality across MSA depths.} Unperturbed model\_1\_ptm predictions for ubiquitin, averaged over 3 seeds. All three depths produce native-like structures; the differences are small and monotonic.}
\label{tab:msa-baselines}
\centering\small
\begin{tabular}{rcccc}
\toprule
MSA depth & \plddt{} & pTM & RMSD (\AA) & $Q$ \\
\midrule
$32 \times 32$  & $92.5 \pm 0.8$ & $0.784 \pm 0.015$ & $0.71 \pm 0.01$ & 0.805 \\
$64 \times 64$  & $93.9 \pm 0.9$ & $0.803 \pm 0.016$ & $0.71 \pm 0.01$ & 0.805 \\
$128 \times 128$ & $94.1 \pm 0.6$ & $0.804 \pm 0.013$ & $0.73 \pm 0.01$ & 0.806 \\
\bottomrule
\end{tabular}
\end{table}

MSA depth controls the model's resilience to perturbation in a monotonic and predictable way: deeper MSAs suppress the effect of weight perturbation, while shallower MSAs amplify it. The key metrics are:

\textbf{Dynamics zone width.} At the primary condition ($\sig = 0.30$, $\lam = 0.75$), the number of perturbation depths producing structures in the dynamics-relevant RMSD range (1.5--5.0~\AA) is 4 at MSA~32, 12 at MSA~64, and 15 at MSA~128. The MSA~32 window is too narrow for systematic exploration; the MSA~64 window spans nearly one-third of the available depth axis.

\textbf{Atomization rate.} At $\lam = 0.75$, the fraction of runs producing atomized structures (RMSD $> 20$~\AA{} or \plddt{} $< 30$) is $17.9\% \pm 2.4\%$ at MSA~32, $3.0\% \pm 2.8\%$ at MSA~64, and $0.2\% \pm 0.3\%$ at MSA~128 (mean $\pm$ SD across 4 matched $\sig$ values). The six-fold reduction from MSA~32 to MSA~64 is the dominant transition; the further reduction from MSA~64 to MSA~128 is modest.

\textbf{Single-layer sensitivity.} Perturbing only the first Evoformer block ($\sig = 0.30$, $\lam = 0.75$, depth~1) produces RMSD $4.7 \pm 1.6$~\AA{} at MSA~32, $1.8 \pm 1.5$~\AA{} at MSA~64, and $0.75 \pm 0.01$~\AA{} at MSA~128. The MSA~128 model is essentially immune to single-layer perturbation, while MSA~32 is already partially unfolded.

\textbf{Deep perturbation.} At depth~30 ($\sig = 0.30$, $\lam = 0.75$), mean RMSD is $11.9 \pm 1.1$~\AA{} (MSA~32), $3.9 \pm 0.1$~\AA{} (MSA~64), and $2.9 \pm 0.1$~\AA{} (MSA~128). MSA~32 has already crossed into the atomized regime, while MSA~64 and MSA~128 remain in the dynamics zone.

\textbf{Consistency of phenomena.} Excluding the known block-2 anomaly at depth~3 (which produces $\sim$17~\AA{} RMSD at all three MSA depths), the qualitative features of the RMSD-vs-depth curve---initial enhancement zone, gradual transition region, and eventual atomization---are present at all three depths. The transition onset shifts to deeper layers as MSA depth increases (reflecting the greater resilience), but the ordering of contact loss groups and the forward/reverse asymmetry are preserved. The block-2 anomaly itself is MSA-independent: it produces comparable RMSD excursions ($17.2$--$18.2$~\AA) at all three depths, consistent with a structural bottleneck in the early Evoformer that is intrinsic to the weight architecture rather than the input representation.

The current comparison is limited to three MSA depths on a single protein (ubiquitin) at 7 of the 192 $(\sig, \lam)$ conditions available at MSA~64. This is sufficient to justify the choice of MSA~64 for the present work: MSA~64 captures most of the stability gain over MSA~32 while remaining sensitive enough for the perturbation to produce conformational diversity. It is not, however, a complete characterization. The behaviour at default MSA depth ($512 \times 5{,}120$), intermediate depths (e.g., $256 \times 256$), and on additional proteins is left to future work.

\section{$\alpha$-synuclein: depth profiles and force-field dependence of the MD reference}
\label{sec:supp_asyn}

\begin{figure}[H]
    \centering
    \includegraphics[width=0.88\textwidth]{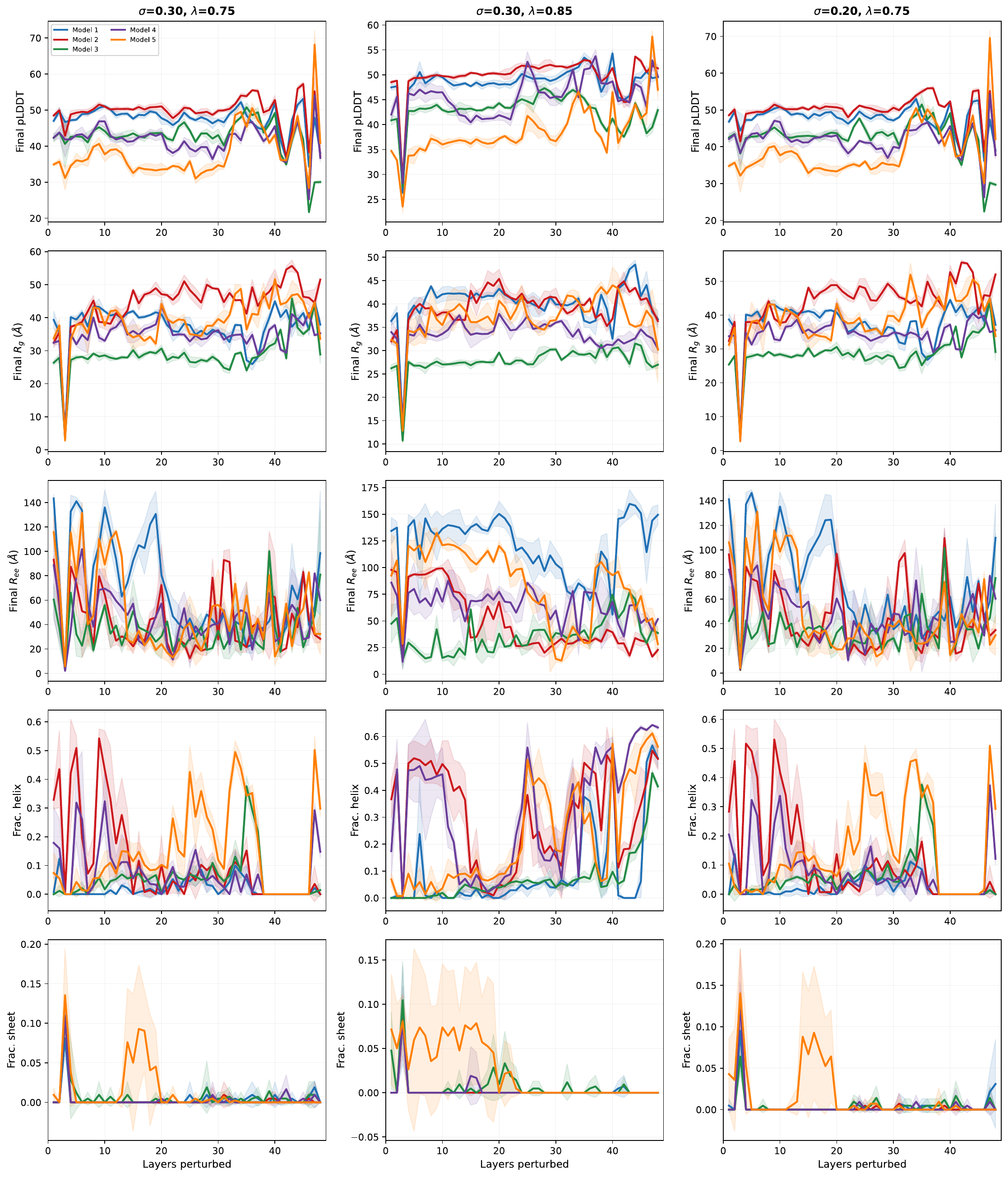}
    \caption{\textbf{Five-model divergence in $\alpha$-synuclein is present at every perturbation depth, not just at the pooled operating point.}
    Final-recycle pLDDT (row~1), radius of gyration $R_g$ (row~2), end-to-end distance $R_{ee}$ (row~3), helix fraction (row~4), and sheet fraction (row~5) plotted against perturbation depth $d$ (number of perturbed Evoformer blocks; cumulative forward, MSA $64\times 64$), for the three shared $(\sig, \lam)$ conditions: $\sig=0.30, \lam=0.75$ (left column), $\sig=0.30, \lam=0.85$ (middle), $\sig=0.20, \lam=0.75$ (right). Each curve is the per-depth mean across 3 MSA-subsampling seeds (42, 43, 44) for one AF2 model (Model~1 blue, Model~2 red, Model~3 green, Model~4 purple, Model~5 orange); shaded bands are $\pm 1$ SD across seeds. The models are already distinguishable at the baseline ($d = 0$, leftmost point) and remain qualitatively distinct across the sweep. Model~3 (green) stays compact throughout ($R_g \approx 26{-}30$~\AA{}); Model~2 (red) sits at the largest $R_g$ across most of the sweep; Models~1, 4, and 5 populate intermediate $R_g$ with distinct $R_{ee}$ and helix-fraction trajectories. Consistent with Section~\ref{sec:results-asyn}, inter-model variation in mean $R_g$ is much larger than inter-seed variation; the pooled operating-point statistic quoted in the main text gives CVs of 14.6\% (inter-model) and 0.1--0.9\% (inter-seed), and a depth-resolved calculation (mean across the 48 depths at $\sig=0.30, \lam=0.75$) gives the same qualitative ordering (18\% inter-model, 3.5--4.8\% per-model inter-seed). That the divergence is present at the baseline and persists through the full perturbation sweep is evidence that it reflects different per-model weight encodings of the same sequence, not a depth-specific perturbation artefact. Sheet fraction remains low throughout (generally $<$10\% across models), consistent with AF2 assigning little $\beta$-sheet content to this sequence under these conditions.}
    \label{fig:supp_asyn_depth}
\end{figure}
\begin{figure}[H]
    \centering
    \includegraphics[width=\textwidth]{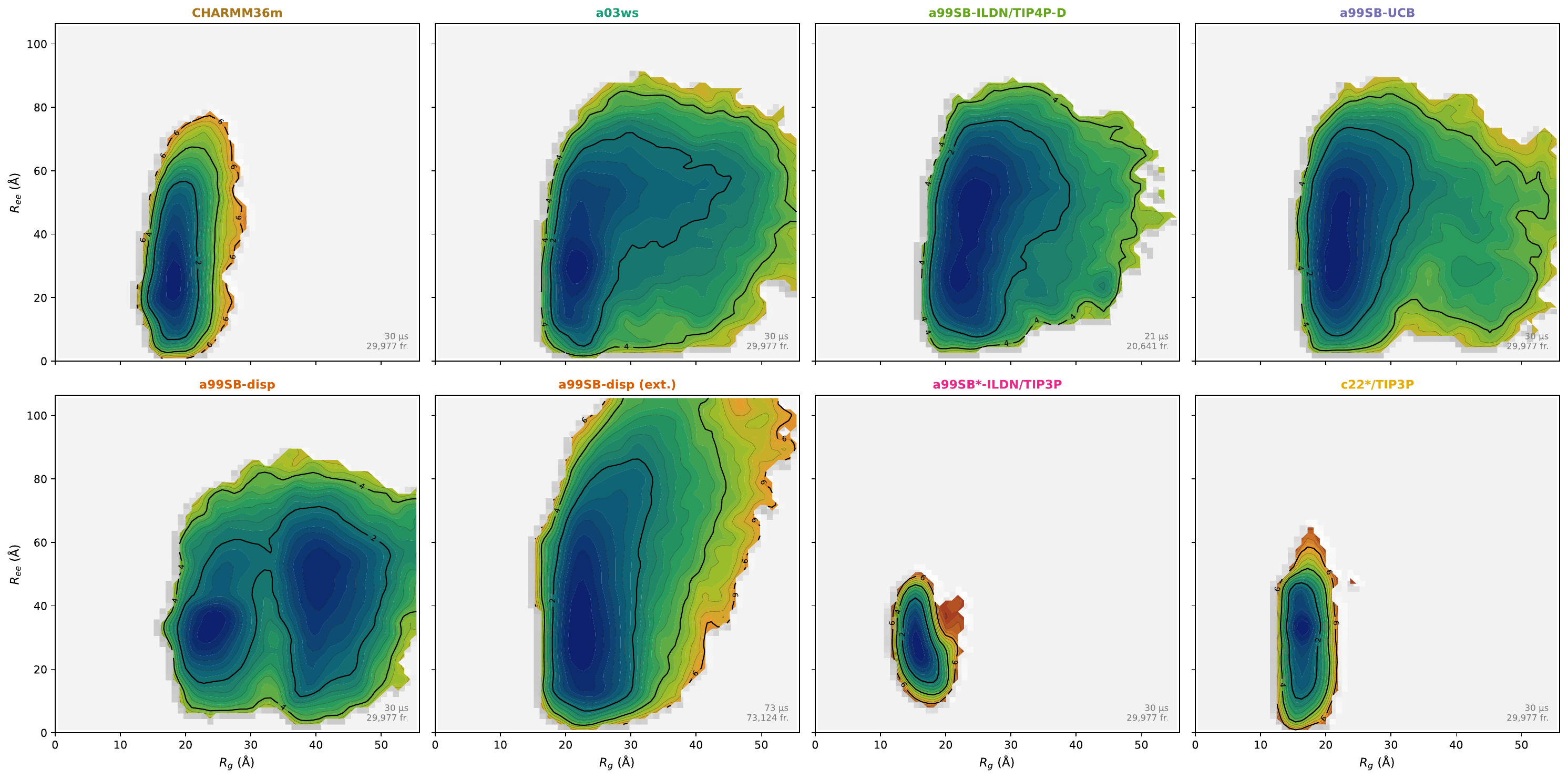}
    \caption{\textbf{The pooled MD reference ensemble is itself a patchwork of force-field-dependent sub-ensembles.}
    Individual $R_g$-versus-$R_{ee}$ density landscapes ($-\ln\rho$; same axes and smoothing as main Fig.~16) for each of the eight DE Shaw explicit-solvent MD trajectories pooled into the main-text reference ensemble (Robustelli, Piana \& Shaw, 2018; total $\sim$273 $\mu$s, 273{,}627 analysed frames). Panel titles give the force-field label; the trajectory duration and frame count are annotated in the lower-right of each panel. For descriptive purposes, the force fields span a continuum of extended-state occupancy: three show no occupancy of the extended-helical box (CHARMM36m, c22*/TIP3P, a99SB*-ILDN/TIP3P---compact-biased, concentrated in the compact corner with 67--85\% of frames at $R_g < 25$~\AA{} and $R_{ee} < 35$~\AA{}); two populate it sparsely (a99SB-UCB, a99SB-ILDN/TIP4P-D; 6--9\% of frames in the extended-helical box); and three populate it more substantially (a03ws, a99SBdisp, a99SBdisp-extended; 11--19\% of frames in that box). Force fields favouring compact, globule-like ensembles and force fields favouring broader, more extended ensembles both exist; no single trajectory recapitulates the union. The pooled MD reference in main Fig.~16B is therefore a union of these disagreeing sub-ensembles, not a single internally consistent reference ensemble. Using the axis-aligned rectangle defined by the pooled-MD marginal 2.5--97.5 percentiles of $R_g$ and $R_{ee}$, 66.6\% of AF2 frames fall inside that rectangle; the remainder lie outside it in model-specific directions (Section~\ref{sec:results-asyn}).}
    \label{fig:supp_asyn_md_per_ff}
\end{figure}

Section~\ref{sec:results-asyn} presents $\alpha$-synuclein as the ``underdetermined'' pole of the three-protein spectrum: five models produce five internally-coherent but mutually-disagreeing landscapes, and the pooled AF2 ensemble is compared against pooled explicit-solvent MD as an external reference. Two subsidiary diagnostics support those claims. The first asks whether the inter-model divergence is specific to the pooled-operating-point ensemble shown in main Fig.~15, or whether it persists across the full perturbation sweep (Fig.~\ref{fig:supp_asyn_depth}). The second asks whether the eight-force-field MD pool used as the reference in main Fig.~16B is internally self-consistent, so that the reader can judge how heterogeneous that reference is (Fig.~\ref{fig:supp_asyn_md_per_ff}).

Taken together, these two diagnostics refine the main-text claim. The per-model divergence (Fig.~\ref{fig:supp_asyn_depth}) is an intrinsic property of each AF2 model's weights, not an emergent feature of any particular perturbation depth; it is already present in the unperturbed baseline and persists throughout the sweep. And the MD ensemble (Fig.~\ref{fig:supp_asyn_md_per_ff}) is a force-field-scale mixture whose sub-ensembles disagree substantially, so the appropriate comparison is ``which MD sub-ensembles does each AF2 model agree with,'' not ``does AF2 match an MD consensus.'' We leave that per-model / per-force-field mapping to future work; the falsifiable claim of Section~\ref{sec:results-asyn} is the shared-core region where all five AF2 models and the pooled MD overlap.

\section{Secondary-structure content: KaiB TM$_\text{gs}$-threshold crossings and $\alpha$-synuclein baselines}
\label{sec:supp_ss_table}


Section~\ref{sec:results-kaib} argues that the 31 KaiB frames crossing the TM$_\text{gs} > 0.5$ threshold are not ground-state-like structures; Section~\ref{sec:results-asyn} argues that the five AF2 models produce five structurally distinct $\alpha$-synuclein landscapes. Table~\ref{tab:supp_ss} reports DSSP helix and strand fractions for both sets (mdtraj \texttt{compute\_dssp}, 8-category scheme collapsed to H+G+I / E+B). Strand fractions are low (19 $\pm$ 5\% pooled; 15 $\pm$ 3\% in the model-3 near-balanced subset) and do not exceed the AF2 baseline (25\%), but DSSP counts alone do not distinguish 5JYT-like from 2QKE-like $\beta$-sheet topology. Whether any 2QKE-specific sheet arrangement appears is assessed visually in Fig.~\ref{fig:supp_kaib_gallery}.

\begin{table}[H]
\centering
\caption{\textbf{Secondary-structure content of the 31 KaiB TM$_\text{gs}$-threshold crossings and the five per-model $\alpha$-synuclein baselines.}
\textbf{Panel A (KaiB).} Helix (H+G+I) and strand (E+B) fractions from DSSP for the 31 perturbation-sweep frames that cross TM$_\text{gs} > 0.5$ (Section~\ref{sec:results-kaib}; Fig.~\ref{fig:supp_kaib_gallery}), grouped into the two categories identified in the main text (shallow threshold fluctuations in models~1, 4, 5 at depths 1--6; model-3 near-balanced crossings at $d = 5, 24\text{--}30$). Reference rows report the two reference structures (2QKE, ground state; 5JYT, fold-switched) and the unperturbed AF2 model-3 baseline (recycle~32) for comparison. Strand counts are low relative to the AF2 baseline (25\%) and similar to or below the 5JYT reference (22\%) across all groups. DSSP reports strand counts, not sheet arrangement; whether any 2QKE-topology $\beta$-sheet arrangement appears is therefore not decidable from the numbers here, and is addressed by direct structural inspection in Fig.~\ref{fig:supp_kaib_gallery}. The model-3 near-balanced subset shows partial loss of strand content ($\sim$15\% vs.\ 25\% in the AF2 baseline) while helix content is essentially unchanged ($\sim$29\%).
\textbf{Panel B ($\alpha$-synuclein).} Per-model statistics of the unperturbed baselines (final recycle, mean $\pm$ SD across 3 seeds; $n_\text{seeds}=3$ per row). The models differ markedly at baseline: Model~3 is compact and helix-free in all three seeds; Model~2 is consistently extended and helical; Model~4 remains helical but shows substantial seed-to-seed variation in both compactness and helix content (seed~44 gives $R_g = 40.1$~\AA{} and 63.6\% helix, versus $29.9$--$30.5$~\AA{} and $41$--$44$\% for seeds 42 and 43); Models~1 and~5 lie between these extremes, with Model~1 helix content itself varying appreciably across seeds (18\%, 21\%, 45\%). Helix fraction at baseline spans $\sim$0\% (Model~3) to $\sim$50\% (Model~4 mean) and $R_g$ from $25.6$ to $33.5$~\AA{}, supporting the main-text observation that the five-model divergence is a difference in encoded secondary-structure content, not only in overall chain dimensions.}
\label{tab:supp_ss}
\small

\vspace{0.5em}
\textbf{Panel A. KaiB: 31 TM$_\text{gs}$-threshold crossings and reference structures} \\[0.2em]
\begin{tabular}{lccccc}
\toprule
Category & $n$ & $\langle\mathrm{TM}_\text{gs}\rangle$ & $\langle\mathrm{TM}_\text{fs}\rangle$ & Helix\% & Strand\% \\
\midrule
2QKE reference (ground state)              & ---  & ---   & ---   & 28 & 23 \\
5JYT reference (fold-switched)             & ---  & ---   & ---   & 38 & 22 \\
AF2 baseline model~3 (recycle 32)          & 1    & ---   & ---   & 29 & 25 \\
\midrule
Shallow fluctuations (M1/M4/M5, $d \leq 6$)  & 18 & 0.511 & 0.675 & 29 $\pm$ 2 & 21 $\pm$ 5 \\
Model-3 near-balanced (1 at $d=5$; 12 at $d=24\text{--}30$) & 13 & 0.503 & 0.567 & 29 $\pm$ 1 & 15 $\pm$ 3 \\
\midrule
All 31 crossings (pooled)                  & 31 & 0.508 & 0.630 & 29 $\pm$ 2 & 19 $\pm$ 5 \\
\bottomrule
\end{tabular}

\vspace{1em}
\textbf{Panel B. $\alpha$-synuclein: per-model unperturbed baselines (final recycle)} \\[0.2em]
\begin{tabular}{lcccc}
\toprule
Model & $R_g$ (\AA) & $R_{ee}$ (\AA) & Helix\% & Strand\% \\
\midrule
model\_1\_ptm & 32.2 $\pm$ 0.6 &  99 $\pm$ 10 & 28 $\pm$ 15 & 0 $\pm$ 0 \\
model\_2\_ptm & 31.9 $\pm$ 0.5 & 104 $\pm$  5 & 45 $\pm$  3 & 0 $\pm$ 0 \\
model\_3\_ptm & 25.6 $\pm$ 0.8 &  61 $\pm$ 16 &  0 $\pm$  0 & 8 $\pm$ 3 \\
model\_4\_ptm & 33.5 $\pm$ 5.7 &  60 $\pm$  6 & 50 $\pm$ 12 & 0 $\pm$ 0 \\
model\_5\_ptm & 31.3 $\pm$ 0.5 &  98 $\pm$ 10 &  6 $\pm$  4 & 4 $\pm$ 4 \\
\bottomrule
\end{tabular}
\end{table}

Two points follow. First, across the 31 KaiB threshold crossings mean helix content is $29 \pm 2$\%---lower than the 5JYT reference (38\%) and similar to the AF2 baseline (29\%); helix fraction alone does not distinguish 5JYT-like from 2QKE-like topology, and strand count (19 $\pm$ 5\% pooled) does not exceed the AF2 baseline (25\%). Whether the $\beta$-sheet arrangement is 5JYT-like or 2QKE-like is a topology question beyond DSSP's resolution and is assessed directly in Fig.~\ref{fig:supp_kaib_gallery}. Second, Table~\ref{tab:supp_ss} shows strong model dependence of the $\alpha$-synuclein baseline ($R_g$ 25.6--33.5~\AA, helix 0--50\%), but the final-recycle baselines are not uniformly seed-tight: Model~4 compactness and helix content both vary appreciably across the three seeds (Model~4 seed~44 differs from seeds 42 and 43 by $\sim$10~\AA{} in $R_g$ and $\sim$20 percentage points in helix fraction), and Model~1 helix fraction spans 18--45\%. This baseline-level seed sensitivity is not reflected in the pooled-across-depths CV (0.1--0.9\%) reported in Section~\ref{sec:results-asyn}, which summarizes the full depth-pooled landscape rather than the final-recycle baseline alone.

\section{Noise controls: supplementary diagnostics}
\label{sec:supp-noise-diagnostics}

\begin{figure}[H]
    \centering
    \includegraphics[width=\textwidth]{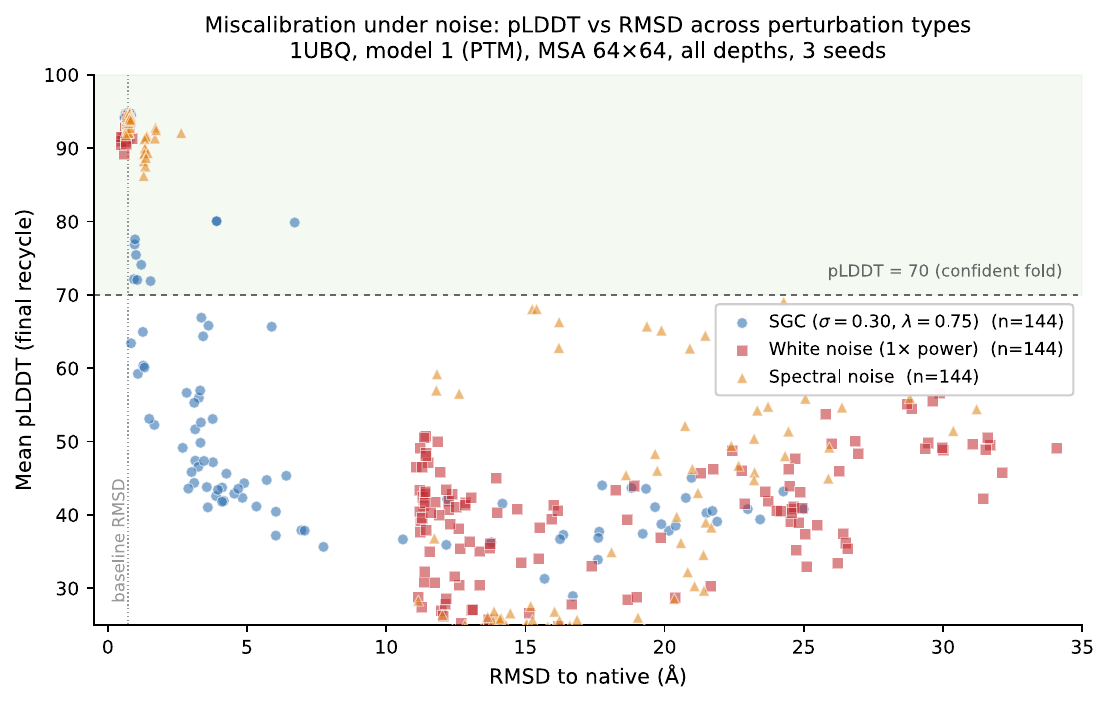}
    \caption{\textbf{Predicted confidence versus structural deviation across perturbation types.}
    Each point is the final-recycle structure of one run ($n = 144$ per condition: 48 perturbation depths $\times$ 3 MSA-subsampling seeds). \sgc{} runs (blue circles) occupy the upper-left of the plane, tracing a calibrated curve from the near-native baseline down through the transition boundary---high \plddt{} tracks low RMSD. White-noise (red squares) and spectral-noise (orange triangles) controls, power-matched to \sgc{} in weight-space, scatter broadly across high RMSD with variable \plddt{}, including a substantial fraction of runs in the ``confident fold'' band (\plddt{} $\geq$ 70, green shading) at RMSD 10--25~\AA{}. The dissociation of confidence from native-likeness under random perturbation is the miscalibration signature that the \sgc{} instrument is not. Single-condition reference: \sgc{} at $\sig = 0.30$, $\lam = 0.75$; noise controls at power-matched amplitudes.}
    \label{fig:supp_noise_plddt}
\end{figure}
\begin{figure}[H]
    \centering
    \includegraphics[width=\textwidth]{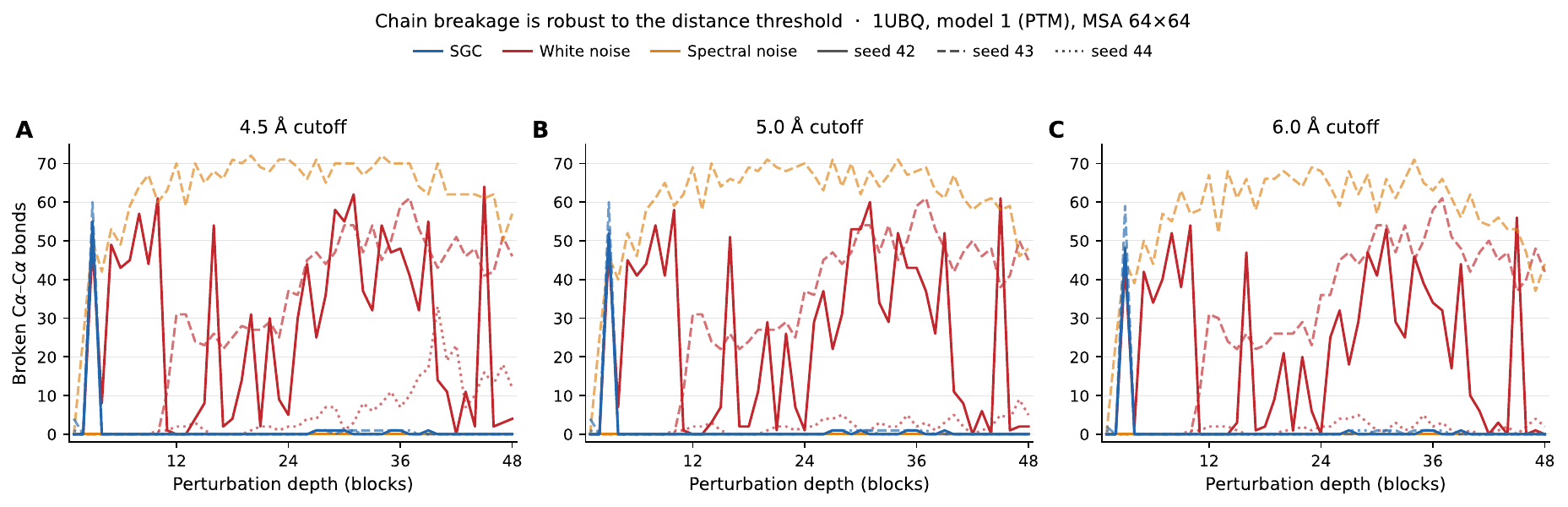}
    \caption{\textbf{Chain-breakage robustness to the C$\alpha$--C$\alpha$ distance threshold.}
    Number of sequential C$\alpha$--C$\alpha$ bonds exceeding the indicated cutoff, plotted against perturbation depth, for three thresholds: \textbf{(A)}~4.5~\AA{}, \textbf{(B)}~5.0~\AA{} (the main-text choice), \textbf{(C)}~6.0~\AA{}. Under \sgc{} (blue), no threshold yields an appreciable number of broken bonds at any depth; under white-noise (red) and spectral-noise (orange), the characteristic $\sim$50--70 broken-bond plateau is preserved across all three thresholds with only quantitative shifts. The conclusion that noise controls break the polypeptide chain while \sgc{} does not is therefore not an artefact of the cutoff choice.}
    \label{fig:supp_noise_breakage}
\end{figure}
\begin{figure}[H]
    \centering
    \includegraphics[width=\textwidth]{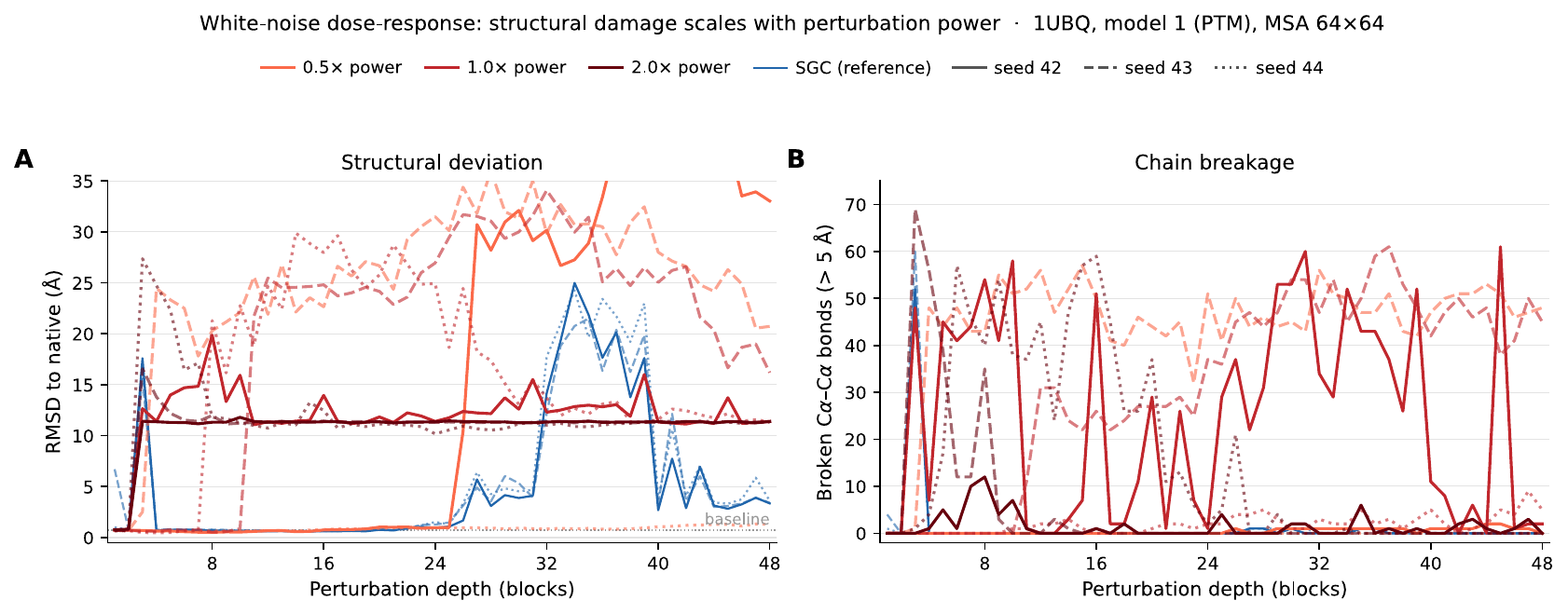}
    \caption{\textbf{White-noise dose-response across a factor-of-four range in perturbation power.}
    White-noise perturbations applied at 0.5$\times$, 1.0$\times$, and 2.0$\times$ the power-matched baseline of \sgc{} at $\sig = 0.30$. \textbf{(A)}~RMSD to native: at 0.5$\times$ power, structural damage is deferred to deeper perturbation (depth $\gtrsim 20$); at 1.0$\times$ and 2.0$\times$, damage appears at shallower depths and with large seed-to-seed variation. \textbf{(B)}~Chain breakage ($>$~5.0~\AA{} C$\alpha$--C$\alpha$ cutoff): a similar pattern. In every regime, the across-seed spread (solid / dashed / dotted) is far larger than the across-dose spread, underscoring the stochastic character of the noise controls. The \sgc{} reference (blue) is shown for context: same per-tensor perturbation power, tight seed-to-seed agreement, no chain breakage.}
    \label{fig:supp_noise_dosage}
\end{figure}

Section~\ref{sec:results-noise} contrasts \sgc{} with two power-matched noise controls (white and spectral). Three additional diagnostics support that comparison: the joint distribution of predicted confidence and structural deviation (Fig.~\ref{fig:supp_noise_plddt}), the robustness of the chain-breakage conclusion to the C$\alpha$--C$\alpha$ distance threshold (Fig.~\ref{fig:supp_noise_breakage}), and the response of the white-noise control to a factor-of-four variation in perturbation power (Fig.~\ref{fig:supp_noise_dosage}).

\section{Territory-coverage estimator robustness}
\label{sec:supp-territory}

\begin{figure}[H]
    \centering
    \includegraphics[width=\textwidth]{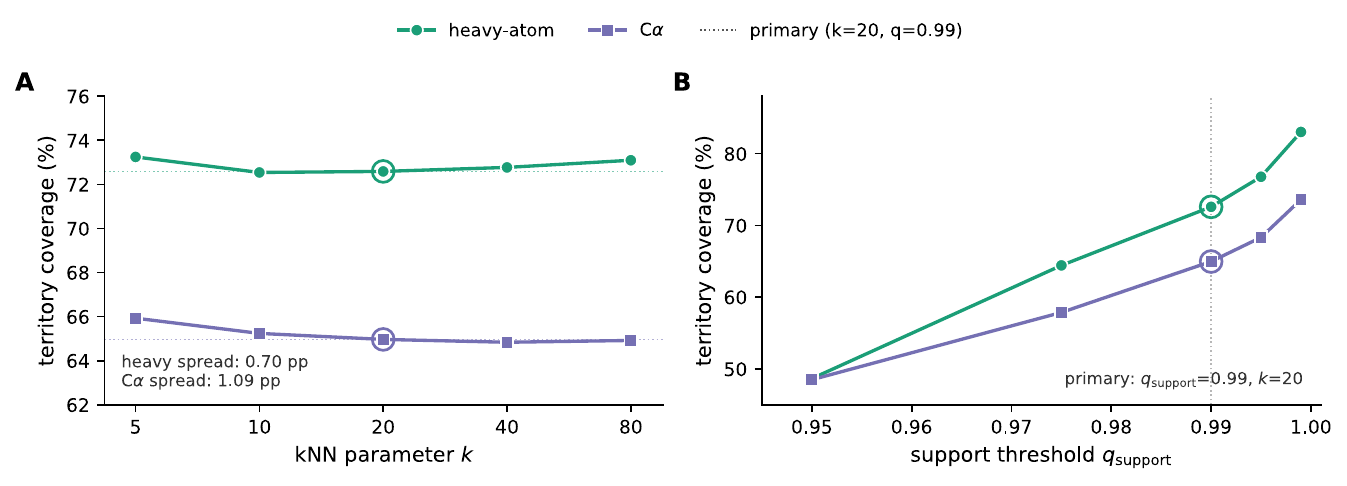}
    \caption{\textbf{Sensitivity of kNN territory-coverage to estimator hyperparameters.}
    \textbf{(A)}~Varying $k$ at fixed $q_{\text{support}} = 0.99$: heavy-atom coverage stays in the narrow range $72.54$--$73.25\%$ ($0.70$~pp spread) across $k \in \{5, 10, 20, 40, 80\}$; C$\alpha$ coverage stays in $64.84$--$65.93\%$ ($1.09$~pp spread). The cited $k = 20$ point (circled) is not a cherry-pick: any $k$ in this range yields the same coverage to better than a percentage point.
    \textbf{(B)}~Varying $q_{\text{support}}$ at fixed $k = 20$: coverage depends monotonically on how strict the support-threshold is, as expected from its definition ($q_{\text{support}} \to 1$ admits the tails and raises coverage; $q_{\text{support}} \to 0$ restricts to the dense core and lowers it). The chosen $q_{\text{support}} = 0.99$ (dotted line) is the conventional 99\% level. The $k$-stability in panel~A therefore holds conditional on this threshold choice.
    Dotted horizontal lines in panel~A mark the cited heavy-atom and C$\alpha$ primary values.}
    \label{fig:supp_territory_sensitivity}
\end{figure}

The territorial overlap between the AF2 perturbation landscape and the 390~K MD folding ensemble reported in Section~\ref{sec:results-topology} is quantified with a bin-free kNN support estimator (Methods). The estimator has two tunable hyperparameters: the neighbourhood size $k$ used to define each point's local support radius, and the percentile threshold $q_{\text{support}}$ that sets how permissively a point is counted as \emph{inside} the support. The primary result in the main text ($72.6\%$ heavy-atom; $65.0\%$ C$\alpha$) uses $k = 20$ and $q_{\text{support}} = 0.99$. Figure~\ref{fig:supp_territory_sensitivity} documents sensitivity to both knobs.

\section{Forward-reverse perturbation asymmetry}
\label{sec:supp-forward-reverse}

The Results (Section~\ref{sec:results-transition}) report a forward-reverse perturbation asymmetry, which the Discussion cites as supporting evidence that the \sgc{} signal is not generic capacity loss. This section expands that summary with the underlying depth-resolved numbers; the analysis is restricted to ubiquitin, model\_1\_ptm, at the primary operating point ($\sig = 0.30$, $\lam = 0.75$, MSA $64 \times 64$).

Both schemes apply the same per-block Gaussian-and-scale perturbation; they differ only in \emph{which} contiguous block range is corrupted. At perturbation depth $d$ (the number of perturbed Evoformer blocks), the forward scheme perturbs blocks $\{0, \ldots, d{-}1\}$ and leaves the later blocks $\{d, \ldots, 47\}$ clean; the reverse scheme perturbs blocks $\{48{-}d, \ldots, 47\}$ and leaves the early blocks $\{0, \ldots, 47{-}d\}$ clean. We plot the unperturbed baseline at $d = 0$ as a notational anchor (the sweep itself runs over $d = 1$--$48$); at $d = 48$ both schemes perturb all 48 blocks and are identical by definition (the 3-seed means coincide to within 0.01~\AA{}); at intermediate $d$ they apply the same per-block perturbation budget ($d$ blocks with the same $\sig$ and $\lam$) to different positions in the Evoformer stack. Both forward and reverse runs use three seeds (42, 43, 44) across all 48 depths (Table~1, row ``Reverse''). Seed-to-seed spread is small in both schemes: forward s.d.\ is 0.12~\AA{} at $d = 30$ and 0.23~\AA{} at $d = 35$; reverse s.d.\ is comparably small throughout the sweep.

\begin{figure}[H]
    \centering
    \includegraphics[width=\textwidth]{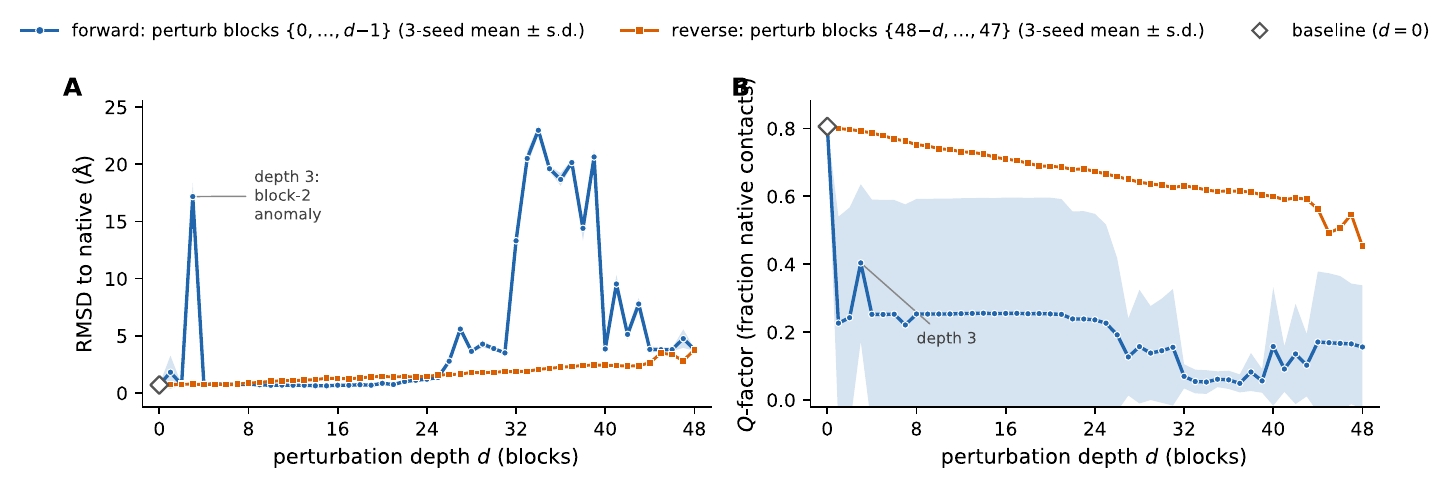}
    \caption{\textbf{Forward-reverse perturbation asymmetry at the primary operating point.}
    RMSD to native \textbf{(A)} and $Q$-factor \textbf{(B)} versus perturbation depth $d$ (number of perturbed Evoformer blocks) for ubiquitin, model\_1\_ptm, $\sig = 0.30$, $\lam = 0.75$, MSA $64 \times 64$. Forward (blue) and reverse (orange) both show 3-seed mean $\pm$ s.d.\ (seeds 42, 43, 44). Forward perturbs blocks $\{0, \ldots, d{-}1\}$; reverse perturbs blocks $\{48{-}d, \ldots, 47\}$; $d = 0$ (grey diamond) marks the unperturbed baseline shared by both schemes; at $d = 48$ both schemes perturb the full Evoformer and are identical by definition (3-seed means overlap to within 0.01~\AA{}). The depth-3 block-2 anomaly (Section~\ref{sec:results-transition}) is annotated. Under forward perturbation the prediction enters a structural transition between depths 25 and 35 with RMSD peaking at 23.0~\AA{} and $Q$ collapsing below 0.1; under reverse perturbation both metrics drift gradually and neither reaches the forward-transition regime at any depth. If the response were position-independent at matched $d$, the two curves should be similar; the visible separation between depths $\sim$25--44 is the asymmetry cited in the Discussion as supporting evidence (Section~\ref{sec:results-noise}).}
    \label{fig:supp_forward_reverse}
\end{figure}

Figure~\ref{fig:supp_forward_reverse} shows both curves overlaid on common axes. Under forward perturbation, excluding the depth-3 block-2 anomaly (a separate reproducible phenomenon, mean RMSD 17.2~\AA{}, discussed in Section~\ref{sec:results-transition}) and a depth-1 single-seed outlier (seed~43: 3.49~\AA{}; seeds~42 and~44 both near 0.7~\AA{}), ubiquitin's RMSD to native remains near baseline through depth 24 (0.63--1.21~\AA{} mean over 3 seeds at depths 2 and 4--24). Between depths 25 and 35 the prediction enters a structural transition: RMSD rises from 1.36~\AA{} at $d = 25$ through a peak of 22.96~\AA{} at $d = 34$ and remains at 19.60~\AA{} at $d = 35$; over the same range $Q$ collapses from 0.705 to 0.051 and $R_g$ expands from 11.71 to 22.78~\AA{}. At very deep perturbation the forward tail becomes non-monotonic, with several depths re-entering the 3.8--9.5~\AA{} range (e.g., $d = 40, 42, 44$--$48$). Under reverse perturbation (3-seed mean), RMSD increases gradually overall, with small local fluctuations: 0.73~\AA{} at $d = 1$, 1.03 at $d = 10$, 1.44 at $d = 20$, 1.81 at $d = 30$, 2.15 at $d = 35$, 2.82 at $d = 47$, and 3.79~\AA{} at $d = 48$. $Q$ and $R_g$ show the same asymmetry: reverse $Q$ declines from 0.799 at $d = 1$ to 0.453 at $d = 48$ without collapsing below 0.49 anywhere in between, while reverse $R_g$ peaks at 13.1~\AA{} over the full sweep. Forward $R_g$, by contrast, peaks at 26.4~\AA{} at $d = 34$ before contracting back to 13.1~\AA{} at $d = 48$---so the two schemes coincide at their shared $d = 48$ endpoint but diverge by more than 10~\AA{} in $R_g$ at matched intermediate depths.

Under a position-independent generic-damage null, where the response depends only on $d$, $\sig$, and $\lam$ regardless of \emph{which} block window is corrupted, the forward and reverse curves should be similar at matched $d$. They are not. Corrupting the early $d$ blocks drives the prediction into a strongly unfolded regime (RMSD $\gtrsim 20$~\AA{}, $Q < 0.1$) that corrupting the last $d$ blocks never reaches at this operating point. This indicates that, for ubiquitin in model\_1\_ptm at $\sig = 0.30$, $\lam = 0.75$, early-block perturbation is much more effective than late-block perturbation at driving the prediction out of the native basin---consistent with, but not by itself proving, a disproportionate early-block role in coarse fold establishment. The perturbed stack is re-applied across 32 recycling iterations, so the asymmetry is also consistent with recycling-mediated compensation, in which clean early blocks partially wash out late-block corruption on subsequent passes through the Evoformer stack. Cross-model reverse comparisons (data collected for models~2--5, Table~1) are not analyzed here; this asymmetry is cited as supporting evidence in Section~\ref{sec:results-noise} and is one of the observations motivating the per-block profiling future work called out in Section~\ref{sec:discussion-future}.

\section{Raw vs energy-minimized structural agreement}
\label{sec:supp_raw_vs_min}

Methods (Section~\ref{sec:methods-structural}) states that all structural analyses use raw, unminimized AlphaFold2 predictions, with the caveat that force-field-dependent hydrogen placement and vacuum minimization introduce biases that would complicate attribution to the learned weights. That choice motivates a complementary question: are the raw predictions geometrically reasonable to begin with, or do they carry the kind of local strain---steric clashes, bad angles, residual distortions---that would invalidate downstream structural observables? This section addresses a narrower question than full thermodynamic stability: under a restrained all-atom vacuum minimization, do the raw predictions require only modest C$\alpha$ adjustment to reach a nearby local minimum? This readout is intended to detect gross local strain, not to claim full stereochemical validation or unrestrained thermodynamic stability.

\textbf{Scope.} We analyze a single validation slice of the survey: ubiquitin, model\_1\_ptm, seed~42, $\sig = 0.30$, $\lam = 0.75$, MSA $64 \times 64$, forward cumulative depths $d = 0$--$48$. All stored recycle outputs (recycles 0--32; 33 structures from 32 recycling iterations) were minimized independently with two all-atom force fields (CHARMM36 and AMBER14SB), yielding 3{,}234 raw/minimized PDB pairs (49 runs $\times$ 33 stored recycle outputs $\times$ 2 force fields). Minimization used harmonic restraints of $k = 100$~kJ/mol/nm$^2$ on heavy atoms, NoCutoff nonbonded interactions, HBonds constraints, and L-BFGS to a 1.0~kJ/mol tolerance, implemented via OpenMM~8.4.0 and PDBFixer~1.12.0 on the CUDA platform (Supplementary Methods, \ref{sec:supp_methods}). C$\alpha$ RMSD is computed over the 76 shared C$\alpha$ atoms only, in the shared coordinate frame without Kabsch superposition---the restraints preserve the global coordinate frame so no rigid-body motion needs to be removed. Energy summaries use the subset of rows with recorded, finite minimized energies: AMBER14SB lacks energy tables at $d = 1$--$5$, and a small number of additional per-recycle energy rows are missing or filtered, so Panel~B has depth-dependent $n$; the RMSD panel is unaffected.

\begin{figure}[H]
    \centering
    \includegraphics[width=\textwidth]{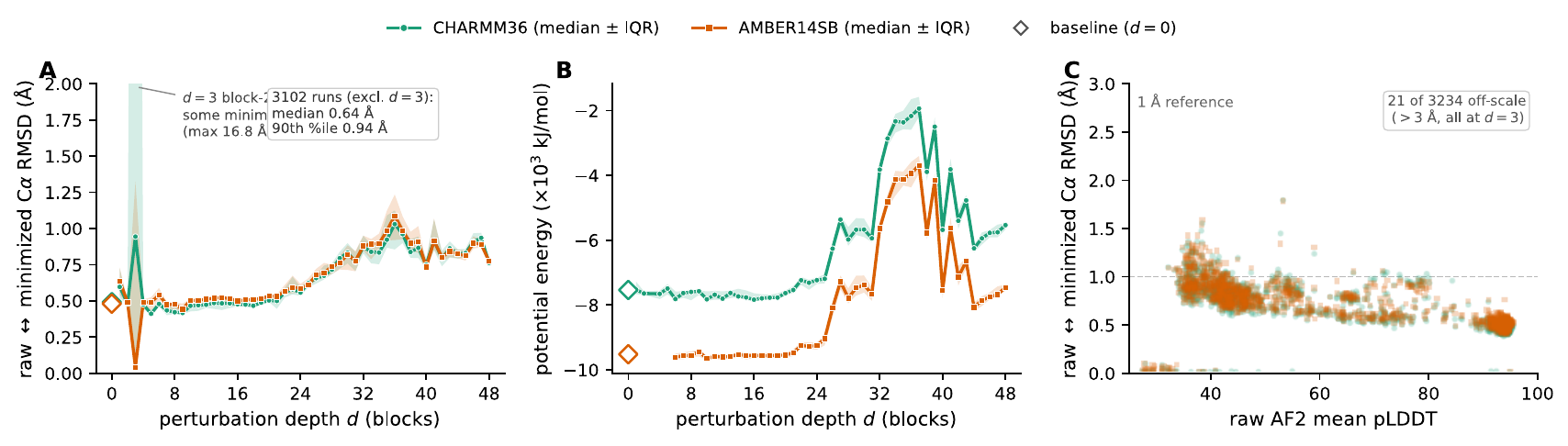}
    \caption{\textbf{Raw AF2 predictions require only modest C$\alpha$ adjustment to reach a nearby local minimum across the operating-point sweep.}
    \textbf{(A)}~C$\alpha$ RMSD between raw and restrained-minimized structures versus perturbation depth $d$ (76 shared C$\alpha$ atoms, no Kabsch superposition). Median and IQR over the 33 stored recycle outputs per depth; CHARMM36 (teal circles) and AMBER14SB (vermillion squares). Baseline ($d = 0$) marked with open diamonds. The $y$-axis is clipped at 2~\AA{} to resolve the sub-{\AA}ngstr\"om signal; the $d = 3$ block-2 anomaly (Section~\ref{sec:results-transition}) produces individual-recycle displacements up to 16.8~\AA{} that fall off scale and are annotated. The inset reports the aggregate over $d = 1$--$48$ excluding $d = 3$ (3{,}102 raw/minimized pairs, both force fields).
    \textbf{(B)}~Potential energy after minimization versus depth. The AMBER14SB curve begins at $d = 6$ because per-recycle energy CSVs were not written for $d = 1$--$5$, and a few additional energy rows are missing or filtered (minimized PDBs still contribute to Panel~A). Both force fields show native-like energies at $d \leq 24$, a rise coinciding with the high-RMSD transition at $d = 25$--$40$, and partial recovery in the deeply perturbed tail. The offset between curves reflects force-field parametrization. Energies are a within-protocol sanity check, not evidence of absolute thermodynamic stability.
    \textbf{(C)}~Per-recycle C$\alpha$ RMSD (raw $\leftrightarrow$ minimized) versus raw AlphaFold2 mean pLDDT of the same recycle, pooled across depths. The dashed 1~\AA{} line is a visual benchmark, not a pass/fail threshold; 21 of 3{,}234 points exceed 3~\AA{}, all at $d = 3$. Lower-pLDDT predictions shift modestly more than high-pLDDT ones (pLDDT $35$--$60$: median 0.83~\AA{}, 90th percentile 1.03~\AA{}; pLDDT $85$--$95$: 0.49~\AA{}, 0.54~\AA{}), but outside the $d = 3$ anomaly displacements remain small in both bins. Pipeline: restrained vacuum minimization with $k = 100$~kJ/mol/nm$^2$ on heavy atoms, NoCutoff nonbonded interactions, HBonds constraints, L-BFGS tolerance 1.0~kJ/mol; OpenMM 8.4.0 with PDBFixer 1.12.0 on the CUDA platform (\texttt{framework/af\_perturb/minimize.py}).}
    \label{fig:supp_raw_vs_min}
\end{figure}

\textbf{Raw structures require only modest C$\alpha$ adjustment across the entire sweep.} Figure~\ref{fig:supp_raw_vs_min}A shows the per-depth median and interquartile range of the raw--minimized C$\alpha$ RMSD. Pooled across the perturbed depths $d = 1$--$48$ excluding the $d = 3$ anomaly (3{,}102 raw/minimized pairs across both force fields), the median displacement is 0.64~\AA{} and the 90th percentile is 0.94~\AA{}. Per-depth medians remain close to 1~\AA{} throughout, peaking at $d = 36$ (1.03~\AA{} for CHARMM36; 1.09~\AA{} for AMBER14SB). At the high-RMSD depths $d = 33$, $34$, and $39$, where the raw prediction is far from native (RMSD-to-native $> 20$~\AA{}, $Q < 0.1$; Section~\ref{sec:results-transition}), the median raw--minimized displacement remains 0.83--0.91~\AA{}. CHARMM36 and AMBER14SB track each other closely (aggregate medians 0.62 vs 0.65~\AA{}; largest per-depth medians 1.03 vs 1.09~\AA{}), making a force-field-specific explanation unlikely within this protocol.

\textbf{Energies track the depth-dependent transition within the restrained protocol.} Figure~\ref{fig:supp_raw_vs_min}B shows the potential energy after minimization. At depths $d \leq 24$, minimized energies remain in the native-like range (CHARMM36: $-7.5$ to $-7.8 \times 10^3$ kJ/mol; AMBER14SB: $-9.5$ to $-9.8 \times 10^3$ kJ/mol). Between $d = 25$ and $d = 40$ they become less negative in parallel with the high-RMSD transition reported in Section~\ref{sec:results-transition}, reaching the shallowest values ($-2 \times 10^3$ kJ/mol for CHARMM36) at depths 33--39 where the raw prediction departs furthest from native. Within this restrained vacuum protocol, no depth outside the $d = 3$ block-2 anomaly requires large additional C$\alpha$ distortion to reach a nearby local minimum, and both force fields show the same depth-dependent trend with a constant offset reflecting parametrization differences. We interpret these energies only as a within-protocol sanity check, not as evidence of absolute thermodynamic stability.

\textbf{C$\alpha$ displacement is modestly pLDDT-dependent but remains small.} Figure~\ref{fig:supp_raw_vs_min}C scatters per-recycle raw--minimized C$\alpha$ RMSD against the raw AlphaFold2 mean pLDDT of the same recycle. Lower-pLDDT predictions shift modestly more under minimization than high-pLDDT predictions, but outside the $d = 3$ anomaly the displacements remain small: the pLDDT $35$--$60$ bin has a median of 0.83~\AA{} and 90th percentile 1.03~\AA{}, versus 0.49~\AA{} and 0.54~\AA{} for pLDDT $85$--$95$. Low pLDDT therefore does not selectively mark gross local geometric failure in this dataset; the 21 outliers above 3~\AA{} are all located at $d = 3$.

\textbf{The depth-3 block-2 anomaly is sensitively detected by the minimization diagnostic.} At $d = 3$ the raw prediction carries a reproducible $\sim$17~\AA{} excursion documented in Section~\ref{sec:results-transition} (and discussed further in Supplementary Section~\ref{sec:supp-chain-integrity} and~16). Within this validation slice, a subset of $d = 3$ recycles show large raw--minimized excursions (C$\alpha$ displacement up to 16.8~\AA{} under CHARMM36), consistent with local strain that requires unusually large within-frame adjustment under the restrained protocol. These recycles are excluded from depth-pooled summaries (consistent with the treatment in Section~\ref{sec:results-transition} and Supplementary Section~\ref{sec:supp-unfolding-multicondition}). That the diagnostic is sensitive to the documented $d = 3$ block-2 anomaly while returning small displacements everywhere else is independent evidence that the sub-{\AA}ngstr\"om displacements at other depths are not a consequence of diagnostic insensitivity.

\textbf{Summary.} Across this validation slice, median raw--minimized C$\alpha$ displacement is sub-{\AA}ngstr\"om at most perturbation depths; the largest non-anomalous per-depth medians are 1.03~\AA{} (CHARMM36) and 1.09~\AA{} (AMBER14SB) at $d = 36$, while the high-RMSD depths $d = 33$, $34$, and $39$ remain 0.83--0.91~\AA{}. Both force fields give the same depth-dependent picture. This supports the Methods decision to analyze raw structures: the RMSD, $Q$-factor, $R_g$, and contact-based observables cited throughout Section~\ref{sec:results} are not dominated by local geometric strain that a restrained minimizer would remove. The test is narrow---one seed, one operating point, one protein, one model---but within its scope the perturbed predictions require only modest C$\alpha$ adjustment to reach a nearby local minimum in either force field, well below the multi-\AA{}ngstr\"om excursions that define the structural transition.

\end{document}